\documentclass[11pt]{article}

\usepackage[margin=1in]{geometry}
\usepackage[T1]{fontenc}
\usepackage[utf8]{inputenc}
\usepackage{lmodern}
\usepackage{textcomp}
\usepackage{amsmath}
\usepackage{amssymb}
\usepackage[numbers,sort&compress]{natbib}
\usepackage{booktabs}
\usepackage{array}
\usepackage{longtable}
\usepackage{lscape}
\usepackage{graphicx}
\usepackage{float}
\usepackage[section]{placeins}
\usepackage{xcolor}
\usepackage{hyperref}

\hypersetup{
  colorlinks=true,
  linkcolor=black,
  citecolor=black,
  urlcolor=blue
}
\newcolumntype{L}[1]{>{\raggedright\arraybackslash}p{#1}}
\newcommand{\ttbreak}[2]{\shortstack[l]{\texttt{#1}\\\texttt{#2}}}
\newcommand{\ttthree}[3]{\shortstack[l]{\texttt{#1}\\\texttt{#2}\\\texttt{#3}}}
\newcommand{\plainbreak}[2]{\shortstack[l]{#1\\#2}}

\newcommand{\AfsRand}{\ttbreak{A5\_schema}{random\_fields}}
\newcommand{\AfsPosthoc}{\ttbreak{A5\_schema}{posthoc\_fields}}
\newcommand{\AfsScrambled}{\ttbreak{A5\_schema}{scrambled\_fields}}
\newcommand{\Aflong}{\ttbreak{A5\_long}{scratchpad}}
\newcommand{\AfourPosthoc}{\ttbreak{A4\_posthoc}{fields}}
\newcommand{\AfourScrambled}{\ttbreak{A4\_scrambled}{fields}}

\newcommand{\reviewarchiveurl}{\href{https://zenodo.org/records/20161579}{Zenodo record 20161579}}
\title{\textbf{Causal state binding predicts action control in language agents}}
\author{Xiao Jia\\School of Artificial Intelligence, The Chinese University of Hong Kong, Shenzhen\\\texttt{xiaojia@link.cuhk.edu.cn}}
\date{}

\begin{document}
\sloppy

\maketitle

\begin{abstract}
Autonomous language agents increasingly expose traces, memories, plans and constraints, but existing evaluations rarely test whether these state variables are causally bound to final actions. We introduce causal state binding: an intervention-coupled evaluation framework that measures whether actions change with the event-specific decisive state while remaining invariant to irrelevant cues. The primary readout is a hidden-target finite-action benchmark in which scorer-side intervention targets are assigned before generation and withheld from the model-visible prompt. Across 57,816 scored records in seven corpus-level units, with primary inference at the dataset-unit level, the structured-agent condition exceeded high-randomness controls and targeted component removals on reason, memory, veto and self-continuity responsiveness. Open-weight validation across Qwen2.5 7B, 14B and 32B plus Mistral-7B showed that action priors, no-field prompts and scrambled decisive context did not recover the structured-control signature; a strict target-lesion slice identified the decisive intervention-carrying field as an information boundary. Within finite-action causal-field probes, the minimal decisive-field readout recovered the prescribed action pattern whereas surface-only, action-prior-only and scrambled-field controls did not. Across 300 SWE-bench Lite issue records and six API models, 18,000 condition/repeat rows were aggregated to 1,800 model-issue records for the primary AUC analysis; adding an oracle-free causal state-binding composite to a full non-CSB baseline increased constraint-clean issue-to-file hit@3 AUC from 0.873 to 0.935 (\(\Delta\)AUC 0.062; issue-cluster lower 95\% bound 0.025). The SWE-bench analysis evaluates the upstream issue-to-file localization endpoint rather than patch application or test-suite resolution. Auxiliary audits define reporting-format, entropy and calibration boundaries. These results support a positive measurement principle for agent evaluation: action control is predicted by event-specific state--action binding, not by output entropy, action-prior matching or rationale format alone.
\end{abstract}

Language agents increasingly expose internal state variables that look action-relevant: reasoning traces, episodic or reflective memory, planning records, self-state summaries and constraint or veto records. Existing agent architectures already show that such components can improve task behavior \citep{wei2022cot,yao2023react,shinn2023reflexion,yao2023tot,park2023generative,wang2024voyager}. The open evaluation problem is therefore not whether modules can affect behavior in general. The harder gap is measurement: how can an evaluator tell whether a final action is bound to the event-specific state that should control it, rather than to a prior, a plausible rationale format or an irrelevant surface cue?

We call this property causal state binding. For an event \(e\), intervention variable \(z\), and finite action \(a\), causal state binding asks whether the final action follows the post-intervention target \(y^{+}\) when the decisive state is present, and whether it avoids unnecessary change when only irrelevant cues vary. This is different from sampling many actions, matching an empirical action prior or producing a plausible rationale. Prior work on cognitive control motivates the distinction between variability and context-sensitive action selection \citep{miller2001integrative,logan1984inhibit,aron2014inhibition,shenhav2013expected}; here these ideas are used only as evaluation design motivation, not as evidence of neural homology or human-like cognition.

\paragraph{Interpretation of causal.}
The word causal refers to experimenter-controlled prompt and state interventions paired with scorer-side counterfactual targets. It does not assert that the emitted trace identifies a latent internal mechanism, a psychological state or the causal status of model reasoning. The framework tests whether action records vary with the intervention-relevant state available to the action query under matched controls.

The framework uses three complementary readouts. Hidden-target finite-action probes test whether final choices from a six-action response set follow scorer-side intervention targets that are assigned before generation and hidden from the model-visible prompt. State-lesion controls test whether removing, scrambling or replacing the decisive event-specific state breaks that behavior while preserving surface task structure, action priors or output format. Matched-interface trace analyses test whether reported reason, memory, self-state and veto fields are coupled to final action under a common schema. Entropy analyses are retained as calibration and boundary checks rather than as the object being optimized.

This design turns a broad qualitative question into a measurement problem. The central quantity is not whether a model can generate diverse outputs, but whether event-specific decisive state predicts the final action after controlling for action priors, reporting format, surface context and irrelevant cues. The main empirical tests therefore focus on no-field, scrambled-context, distribution-matched no-context and strict target-lesion controls. Open-weight local validation is emphasized because it makes model, tokenizer, generation and hardware records more inspectable than provider-only API runs; API validation is used as external robustness evidence.

The contribution is a benchmark-style protocol and evidence chain for measuring intervention-coupled action control in language agents. The paper does not claim model agency, human-like cognition, a causal status of reasoning or broad strict entropy/token matching across all provider and model settings. It instead asks a narrower positive question: can we identify when action selection depends on the decisive state field rather than on output entropy, action priors or rationale format alone?

The evidence is organized accordingly. The finite-action benchmark defines the measurement paradigm. Open-weight no-context, scrambled-context, distribution-matched and strict-lesion controls test the necessity of event-specific decisive state across scaffolds and models, and SWE-bench Lite issue-to-file localization tests whether the resulting state-binding score predicts a real issue-record reliability endpoint. A minimal causal-field experiment then asks whether the decisive field alone can recover the finite-action binding readout, and a local control-binding wrapper tests an auditable mitigation for mismatched decisive fields. Matched-interface AFCI analyses serve as auxiliary trace-faithfulness evidence. Formal entropy/token audits and a bounded local forced-budget matching audit define what is and is not established by entropy and compute matching.

\section*{Results}

We first define causal state binding with hidden-target finite-action probes. We then bring forward the two strongest validation tests: open-weight state-boundary controls and a full 300-record SWE-bench Lite issue-to-file localization predictor analysis. Finite-action readouts, matched-interface trace analyses and entropy/token checks are then used to define the scope of the claim.

\subsection*{Causal state binding is measured with hidden action targets}

The primary readout is deliberately constrained. Each trial asks the model to emit one member of a six-action response set: choose option A, choose option B, veto, defer, recall a prior commitment or mark the output as invalid/unmapped. The expected action is a scorer-side label determined before generation from the intervention condition and hidden from the model-visible prompt. The key measurement is whether the final action changes in the component-specific direction when the decisive state changes, while remaining stable under irrelevant cues. This hidden-target finite-action probe is the format-independent behavioral test; trace-field coupling and entropy are secondary readouts.

\subsection*{Bound state, not action entropy, predicts hidden-target behavior}

The primary behavioral test removes free-form rationale wording from the score. The model emits one of six allowed action codes, and scoring compares the canonicalized final action with a hidden scorer-side target determined before generation. Expected labels were deterministic consequences of the predefined probe manipulation, not labels chosen after observing model outputs. This design separates the semantic content of the prompt from the scorer-side target used to evaluate whether the final action changed in the component-specific direction; the full action ontology is given in the Supplementary Information.

\begin{figure}[p]
\centering
\includegraphics[width=\textwidth]{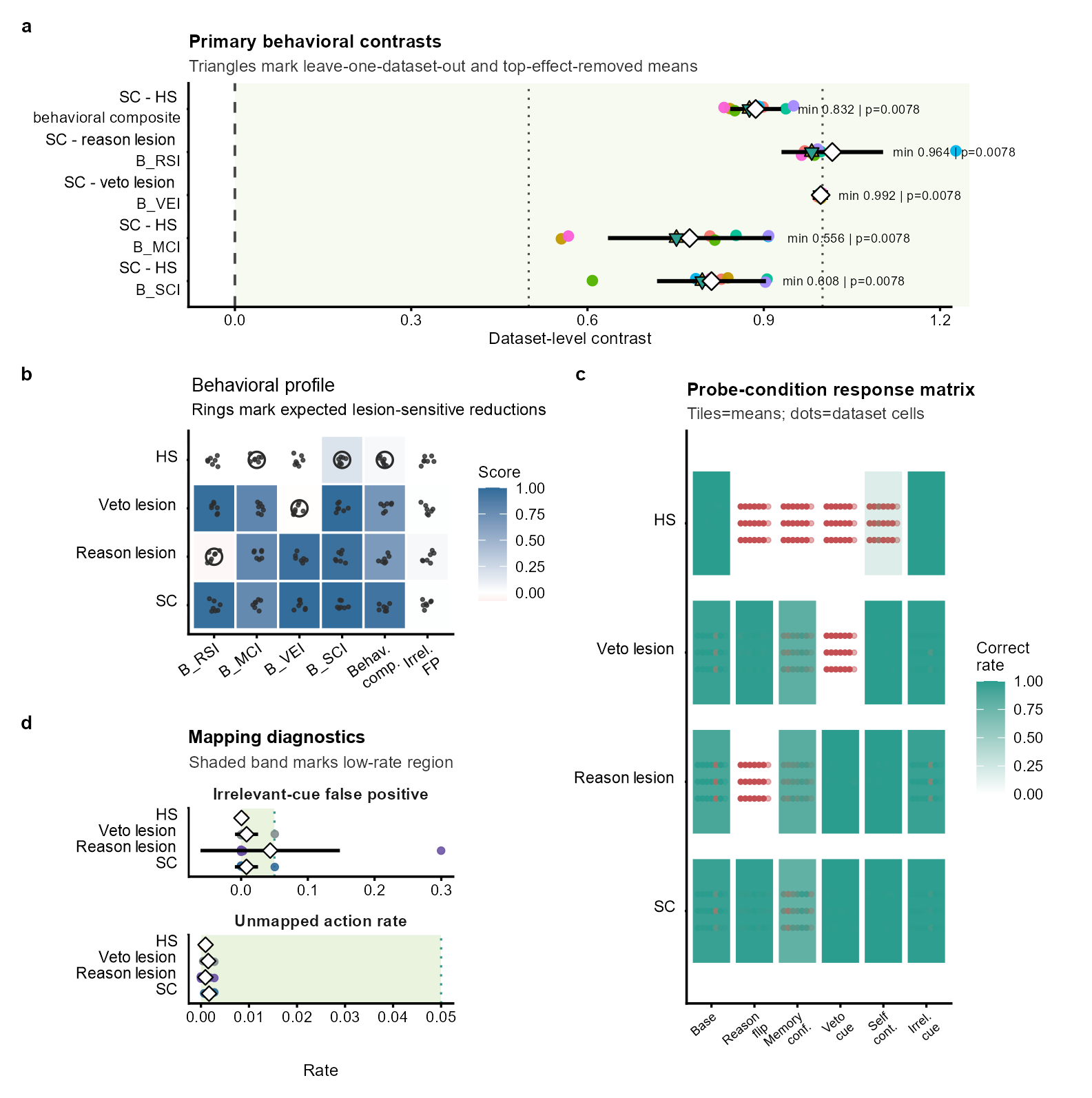}
\caption{Finite-action probes test whether action choices follow hidden intervention targets rather than free-form rationales. \textbf{a}, Dataset-level primary behavioral contrasts. Points are dataset units; diamonds and horizontal segments show descriptive means and 95\% intervals; triangles mark leave-one-dataset-out and top-effect-removed summaries, and labels report minimum dataset contrasts and exact one-sided sign-test \(p\)-values. \textbf{b}, Behavioral profile matrix. \(B_{\mathrm{RSI}}\), \(B_{\mathrm{MCI}}\), \(B_{\mathrm{VEI}}\) and \(B_{\mathrm{SCI}}\) denote behavior-level reason, memory, veto and self-continuity component scores; rings mark the expected lesion-sensitive reductions. \textbf{c}, Probe-condition response matrix under the finite experimental action alphabet. Tiles show weighted condition means; embedded dots show the underlying dataset/replicate condition cells. \textbf{d}, Mapping and specificity error-rate summary for unmapped actions and irrelevant-cue false positives; the shaded band marks the low-rate region. Expected labels were scorer-side annotations hidden from the model-visible prompt.}
\label{fig:finite_action_behavior}
\end{figure}

The benchmark used all seven corpus-level units, 120 sampled events per namespace where available, six probe conditions, four variants and three replicate indices. After predefined deduplication and inclusion rules, it retained 57,816 scored records, with mapping errors and irrelevant-cue false positives below predefined thresholds. The predefined behavioral-composite criterion was evaluated at the dataset-unit level and was met in 7/7 datasets for the structured agent condition exceeding the high-randomness control. Reason, memory, veto and self-continuity scores were also diagnostic. Component-specific criteria were met in 7/7 datasets: the structured agent condition exceeded the reason-removal control on reason responsiveness, the veto-removal control on veto responsiveness, and the high-randomness control on memory and self-continuity responsiveness. The corresponding formal scores were \(B_{\mathrm{RSI}}\), \(B_{\mathrm{VEI}}\), \(B_{\mathrm{MCI}}\) and \(B_{\mathrm{SCI}}\). The primary uncertainty summaries used dataset namespaces as clusters; the dataset-cluster bootstrap lower 95\% bound was 0.8557, and the smallest dataset-level contrasts were 0.832, 0.964 and 0.992 for the behavioral-composite, reason-ablation and veto-ablation contrasts. Figure~\ref{fig:finite_action_behavior} summarizes these scores, contrasts, condition responses and action-mapping quality.

Formal definitions of these finite-action scores are given below for reproducibility; the main behavioral criterion depends only on whether the canonicalized final action matches a hidden scorer-side target.
Let \(\mathcal{Y}=\{\texttt{ACTION\_A},\texttt{ACTION\_B},\texttt{VETO},\texttt{DEFER},\texttt{RECALL\_PRIOR},\texttt{INVALID\_OR\_UNMAPPED}\}\) be the finite action-code ontology. The canonicalizer \(g\) maps the raw model output \(o_i\) to \(\hat{y}_i=g(o_i)\in\mathcal{Y}\). For each probe record \(i\), \(y_i^{-}\) denotes the scorer-side expected action before the manipulation and \(y_i^{+}\) denotes the scorer-side expected action after the manipulation. Correct behavior and unnecessary change were scored as
\begin{align}
c_i &= \mathbf{1}\{\hat{y}_i=y_i^{+}\},\\
u_i &= \mathbf{1}\{z_i=\mathrm{irrelevant}\}\mathbf{1}\{\hat{y}_i\neq y_i^{-}\},
\end{align}
where \(z_i\) indexes the probe condition. For component \(k\in\{\mathrm{RSI},\mathrm{MCI},\mathrm{VEI},\mathrm{SCI}\}\), the behavioral component score is the target-minus-irrelevant contrast
\begin{equation}
B_k(v)=\widehat{\Pr}(c_i=1\mid z_i\in\mathcal{T}_k,v)-\widehat{\Pr}(u_i=1\mid z_i\in\mathcal{N}_k,v),
\end{equation}
where \(\mathcal{T}_k\) is the predefined target-probe set for component \(k\) and \(\mathcal{N}_k\) is the predefined irrelevant or placebo-probe set. This formulation makes the behavioral criterion independent of free-form trace wording: only the post-generation canonical action code is compared with hidden scorer-side labels.

The behavior figure reports the result at the dataset-unit level and shows condition-specific responses and mapping quality. Extended Data Figure 3 shows the complete condition-by-dataset response matrix for the finite-action-code benchmark. In this framing, \(B_{\mathrm{RSI}}\), \(B_{\mathrm{MCI}}\), \(B_{\mathrm{VEI}}\) and \(B_{\mathrm{SCI}}\) are component-level causal-state-binding scores: they estimate whether actions follow decisive reason, memory, veto and self-continuity state while resisting irrelevant-cue changes.
\FloatBarrier

\subsection*{State-boundary controls predict issue-record localization}

The strongest test of the framework is not whether a random comparator fails, but whether controls that preserve action format, action priors or surface context fail when the event-specific decisive state is removed or broken. Open-weight validation used Qwen2.5-7B-Instruct, Qwen2.5-14B-Instruct, Qwen2.5-32B-Instruct and Mistral-7B-Instruct-v0.3. Each model contributed 126,000 scored generations over three scaffolds, 20 task families, seven variants and three seeds. The control set separated different alternatives to causal state binding: no-fields controls removed structured state fields, scrambled-context controls paired the action query with decisive context from a different event, and distribution-matched offline/no-context controls sampled from the structured variant's empirical action prior without observing the decisive intervention context.

\begin{figure}[p]
\centering
\includegraphics[width=\textwidth]{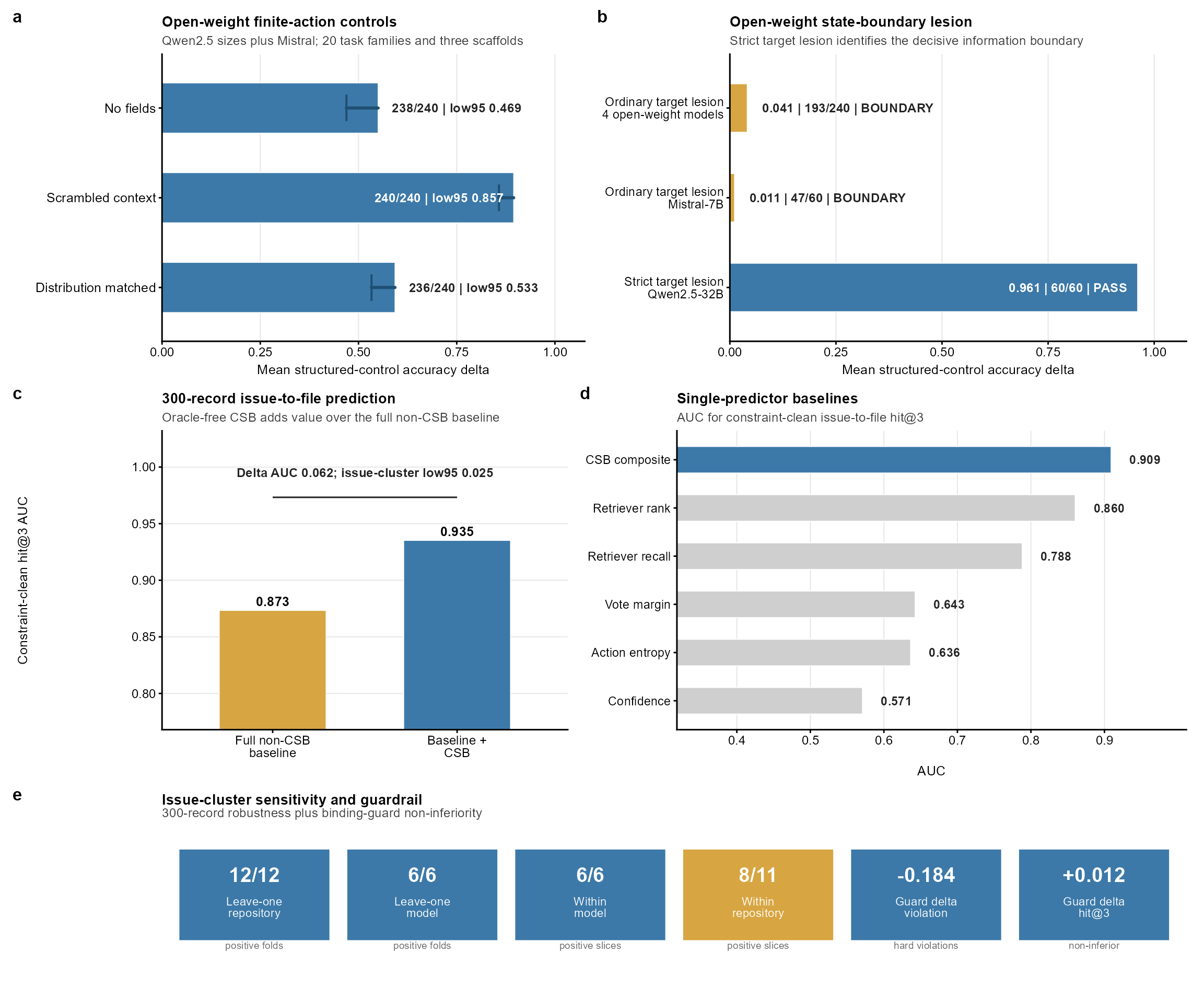}
\caption{Primary validation evidence for state-boundary control and issue-record localization. \textbf{a}, Four-model open-weight core controls. Bars show mean structured-control accuracy deltas for Qwen2.5 7B, 14B and 32B plus Mistral-7B across 20 task families and three scaffolds; labels report positive model--scaffold--task cells and the lowest model-, scaffold- or task-family-cluster lower 95\% bound. \textbf{b}, Open-weight state-boundary lesion analysis. Ordinary target lesions are retained as boundary conditions, whereas the Qwen2.5-32B strict lesion satisfied the predefined cell-level criterion and identified the decisive intervention-carrying field as an information boundary. \textbf{c}, Full 300-record SWE-bench Lite issue-to-file predictor test. Adding oracle-free causal state binding to the full non-CSB baseline increased constraint-clean hit@3 AUC from 0.873 to 0.935. \textbf{d}, Single-predictor AUC comparison for the same issue-to-file outcome; the primary claim uses the incremental baseline-plus-CSB test in panel c. \textbf{e}, Issue-cluster sensitivity and binding-guard summary, including leave-one-repository/model checks and the hard-constraint violation reduction under the guard. Panels c--e evaluate localization of implementation files from SWE-bench Lite issue records; patch generation and test-suite execution are outside this analysis.}
\label{fig:robustness_validation}
\end{figure}

Across this open-weight matrix, structured accuracy exceeded distribution-matched offline/no-context controls in 236/240 model--scaffold--task-family cells, scrambled-context controls in 240/240 cells and no-fields controls in 238/240 cells. The mean deltas were 0.593, 0.895 and 0.550, respectively. A stricter Qwen2.5-32B target-lesion slice then removed the decisive intervention-carrying field while preserving non-decisive context and the action ontology; structured accuracy exceeded the strict lesion in 60/60 cells, with structured mean 0.998, strict-lesion mean 0.037 and mean delta 0.961. These controls identify the decisive state field as an information boundary for the action-control signature in the tested protocols (Figure~\ref{fig:robustness_validation}a,b).

We then ran a SWE-bench Lite issue-to-file localization validation using all 300 retained test issue records and six provider-served model identifiers. This validation was oracle-free on the predictor side: decisive state was constructed from issue-visible and repository-visible information, including the problem statement, repository file tree, retrieval candidates and model-visible localization state, while benchmark reference implementation files were used only for scoring. The primary outcome was constraint-clean implementation-file hit@3. The run produced 18,000 condition/repeat rows, which were aggregated to 1,800 model-issue records for the primary AUC analysis. On these model-issue records, a full non-CSB baseline using model identity, repository, retrieval score, issue statistics, action entropy, self-consistency vote margin, rationale length and confidence achieved AUC 0.873, whereas adding the causal state-binding composite increased AUC to 0.935 (\(\Delta\)AUC 0.062; issue-cluster bootstrap lower 95\% bound 0.025). The gain was positive in 12/12 leave-one-repository and 6/6 leave-one-model analyses, with positive within-model sensitivity in 6/6 models and within-repository sensitivity in 8/11 repositories. A binding-guard arm reduced hard-constraint violations by 0.184 while changing hit@3 by +0.012, above the prespecified non-inferiority margin. These results support oracle-free localization reliability for SWE-bench Lite issue records; patch generation and test-suite resolution are evaluated separately as exploratory scope checks (Figure~\ref{fig:robustness_validation}c--e).

\FloatBarrier

\begin{table}[!htbp]
\centering
\caption{Claim ledger: supported claims and excluded claims. Auxiliary counterfactual-transform, formal API and blinded-audit controls are reported in the Supplementary Information.}
\label{tab:validation_interpretation}
\scriptsize
\resizebox{\textwidth}{!}{%
\begin{tabular}{p{0.20\textwidth}p{0.34\textwidth}p{0.22\textwidth}p{0.18\textwidth}}
\toprule
Evidence block & Primary result & Supported claim & Excluded inference \\
\midrule
Open-weight state-boundary controls & Across Qwen2.5 7B/14B/32B plus Mistral-7B, structured accuracy exceeded no-fields, scrambled-context and distribution-matched controls in 238/240, 240/240 and 236/240 cells; the Qwen2.5-32B strict target-lesion slice passed in 60/60 cells with mean delta 0.961. & Primary model-family and information-boundary validation evidence for the tested finite-action protocols. & Scope: measured finite-action protocols, not model agency, latent reasoning mechanisms or universal generalization to all agent architectures. \\
SWE-bench Lite issue-to-file validation & Across 300 issue records and six API models, 18,000 condition/repeat rows were aggregated to 1,800 model-issue records; adding CSB to the full non-CSB baseline increased constraint-clean hit@3 AUC from 0.873 to 0.935. & Oracle-free issue-to-file localization reliability predictor evidence. & Scope: upstream localization; excludes patch generation, issue resolution, broad software repair and deployed-agent task execution. \\
Minimal decisive-field experiment and wrapper & In 1,440 API generations, full-state and only-decisive-field accuracy were both 1.000, while the best non-decisive control was 0.208; the local wrapper reduced scrambled-field following to 0.033 in both models. & Finite-action binding readout and guardrail evidence. & Scope: finite-action probes; excludes unconstrained reasoning and real-environment task execution. \\
Local forced-budget entropy/token audit & In a bounded Qwen2.5-1.5B CPU run, prompt and completion tokens matched exactly by event, total-token ratio was 1.000, latency ratio was 0.997 and structured accuracy was 0.900 versus 0.325 for the action-prior/no-state control. & Bounded local strict-matching pass under exact local prompt and generation budgets. & Does not convert the larger formal API/open-weight entropy or token/compute attempts into strict matches. \\
\bottomrule
\end{tabular}%
}
\end{table}

Conservative derived effect-size summaries are provided in the Extended Data source figures and in the Figure~\ref{fig:robustness_validation} source-data files. These summaries aggregate the existing evidence blocks at the most conservative dataset-level, model--scaffold--task or reference-model scale available for each comparison. Extended Data Figures 5--8 provide the auxiliary cross-model, open-weight, audit, conservative effect-size, proxy compute and interface-diagnostic displays. The validation source-data map links claim-supporting figures and tables to source-data files, analysis entry points and SHA-256 prefixes, so the claim ledger can be audited from the local outputs and the Zenodo archive.
\FloatBarrier

\subsection*{Minimal decisive fields recover the finite-action binding readout}

Necessity controls show that the finite-action binding readout fails when decisive state is removed or broken. We next tested a narrower diagnostic question with a five-condition causal-state experiment. For each event, the model received either the full state, only the decisive causal field, surface context only, an action-prior-only prompt or a scrambled decisive field from another event; an additional irrelevant-cue condition tested whether an irrelevant surface suggestion changed the action. The experiment used \texttt{gpt-5.4-mini} and \texttt{qwen3-max}, 20 task families, six held-out events per family, six conditions and 1,440 fallback-free API generations with zero parse or transport errors.

Within these intentionally diagnostic finite-action probes, the minimal decisive-field readout recovered the prescribed action pattern. For both models, full-state accuracy was 1.000 and only-decisive-field accuracy was also 1.000. The only-decisive condition explicitly exposed the current event's minimal action-relevant field while omitting surface context; it therefore tests whether the action selector follows the current decisive field rather than a surface cue, action prior or scrambled field. The best non-decisive control was the action-prior-only condition at 0.208; surface-context-only accuracy was 0.033, and scrambled-decisive-field accuracy was 0.008 for \texttt{gpt-5.4-mini} and 0.017 for \texttt{qwen3-max}. The sufficiency recovery fraction, computed as the only-decisive improvement over the best non-decisive control divided by the full-state improvement over that same control, was 1.000 for both models (Table~\ref{tab:causal_sufficiency}). The Supplementary Information includes a leakage and scope audit clarifying that \texttt{expected\_action} is a scorer-side column, while the only-decisive condition intentionally contains an action-code-bearing decisive field. Provider token and latency metadata were retained for the generated rows; mean prompt lengths for full-state and only-decisive prompts were similar within each model, but this block is interpreted as finite-action causal-field sufficiency rather than as strict token/compute matching.

We also tested a local control-binding wrapper on the same records. Before accepting a final action, the wrapper checked whether the visible decisive field belonged to the current event and whether the action matched that field; if a decisive field came from a different event, the wrapper deferred rather than binding to the mismatched field. This wrapper added no second model call. It preserved irrelevant-cue accuracy at 1.000 for both models and reduced acceptance of scrambled decisive fields: raw outputs followed the scrambled field in 0.958 of \texttt{gpt-5.4-mini} rows and 0.533 of \texttt{qwen3-max} rows, whereas wrapped outputs followed the scrambled field in 0.033 of rows for both models. The result establishes a finite-action guardrail over the intervention records; real-environment localization and task-execution effects require separate evaluation.

\begin{table}[!htbp]
\centering
\caption{Minimal causal-state sufficiency and local wrapper results.}
\label{tab:causal_sufficiency}
\scriptsize
\resizebox{\textwidth}{!}{%
\begin{tabular}{lrrrrrrr}
\toprule
Model & Full state & Only decisive field & Surface only & Action prior only & Scrambled field & Scrambled-field following, raw & Scrambled-field following, wrapped \\
\midrule
\texttt{gpt-5.4-mini} & 1.000 & 1.000 & 0.033 & 0.208 & 0.008 & 0.958 & 0.033 \\
\texttt{qwen3-max} & 1.000 & 1.000 & 0.033 & 0.208 & 0.017 & 0.533 & 0.033 \\
\bottomrule
\end{tabular}%
}
\end{table}
\FloatBarrier

\subsection*{Entropy analyses serve as calibration and boundary checks}

After establishing the state-binding readout, entropy analyses were used only to test an alternative explanation: whether output dispersion or stochastic opportunity could recover the same control profile. In the baseline lesion matrix, higher randomness increased action entropy but did not recover reason-, memory-, self-state- or veto-coupled behavior.

Across seven corpora spanning narrative, self-continuity, choice, uncertainty and recall settings, increasing stochasticity raised action entropy but did not reproduce structured-control readouts. In the 74,352-call baseline lesion matrix, the high-randomness control was more unpredictable than the structured agent condition in 7/7 datasets, whereas the structured condition's self-continuity index exceeded the comparator's in 5/7 datasets. Removing the reason graph reduced reason-sensitive structure in 7/7 datasets, and removing veto reduced veto-sensitive structure in 7/7 datasets. This pattern separates action dispersion from control coupling within the measured protocol.

\begin{figure}[p]
\centering
\includegraphics[width=\textwidth]{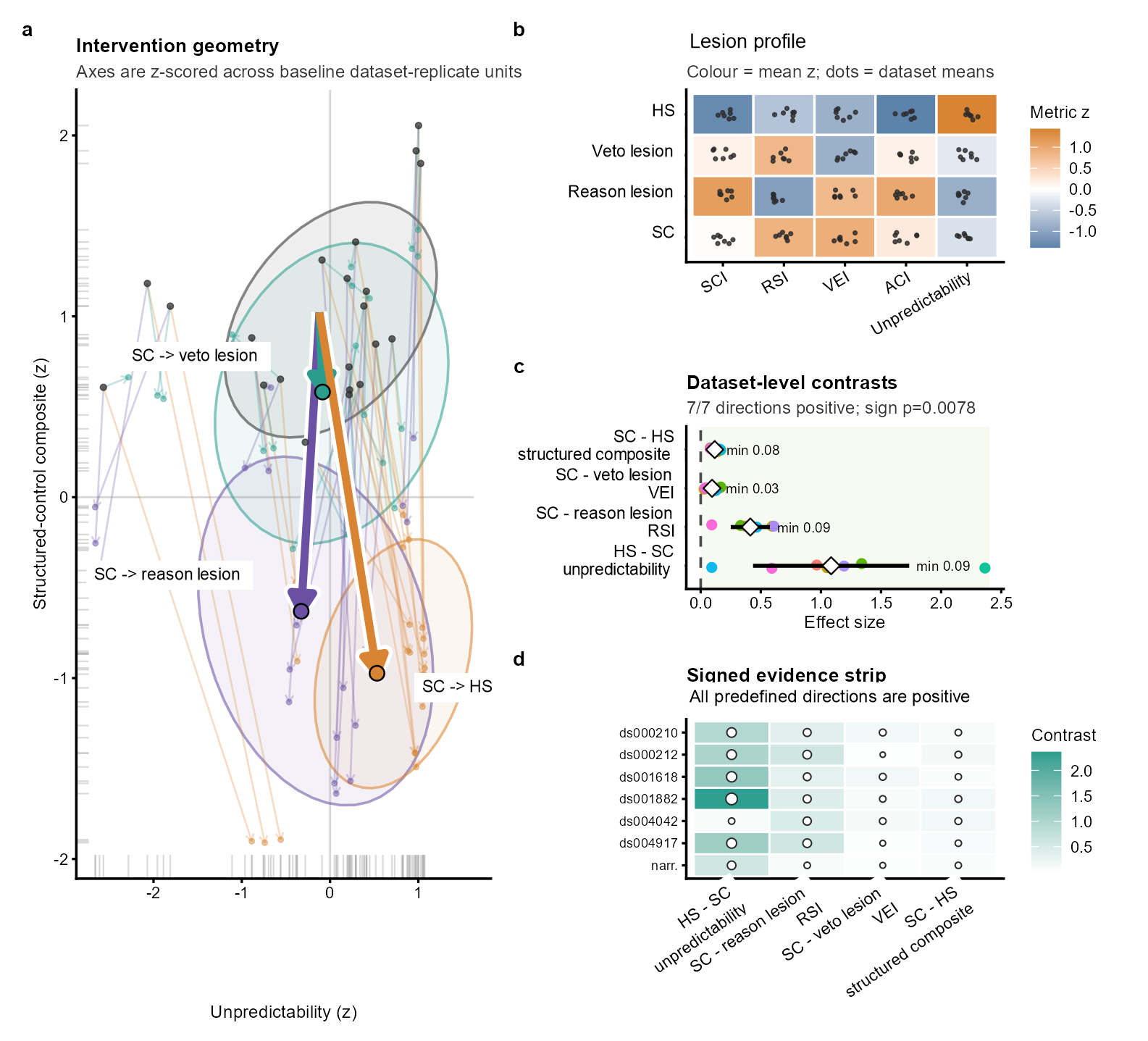}
\caption{High-randomness sampling increases output entropy without reproducing the structured-control profile. \textbf{a}, Each vector shows how replacing the structured agent condition with a control condition changes output entropy and structured-control readouts. The standardized-space axes are raw action unpredictability and the mean of SCI, RSI, VEI and ACI; thin grey vectors show dataset-replicate units; dotted ellipses and marginal rugs show the dataset-replicate distribution; thick labelled vectors show mean displacement. \textbf{b}, Component readout matrix for SCI, RSI, VEI, ACI and raw unpredictability. Colour reports metric-wise \(z\)-score; overlaid dots show dataset-level means. \textbf{c}, Dataset-level contrast forest plot for the predefined baseline directions. Points are dataset units; diamonds and horizontal segments show descriptive means and 95\% intervals across dataset units; the shaded region marks the expected positive direction. \textbf{d}, Signed evidence strip showing direction and magnitude for each dataset-level contrast.}
\label{fig:baseline_dissociation}
\end{figure}

Figure~\ref{fig:baseline_dissociation} visualizes this alternative-explanation check. Entropy was computed over final actions, whereas the structured-control composite summarizes self-continuity, reason sensitivity, veto effect and action continuity; Methods and the Supplementary Information provide the formal definitions. Separate strict entropy and token/compute attempts did not satisfy their registered closeness criteria. No retained calibration candidate matched the structured-control variant across raw-string, rule-canonical, semantic-cluster and action-family entropy representations; the smallest observed maximum gap was 1.171 against the 0.15 criterion. These diagnostics are therefore interpreted as calibration boundaries rather than as entropy-matched controls, with the detailed threshold-normalized analysis reported in Extended Data Figure 4 and the source-data tables.

The baseline visualization therefore summarizes the calibration boundary: output variability rises in the high-randomness control, whereas state-binding readouts remain concentrated in the structured-control variant and its targeted lesion contrasts. Extended Data Figure 1 provides the full dataset- and replicate-resolved baseline metric map supporting this summary display.
\FloatBarrier

\subsection*{Matched-interface traces provide auxiliary faithfulness evidence}

All matched-interface variants used the same reporting schema. The test therefore asks whether the availability of fields such as reason, memory, self-state and veto is enough, or whether those fields must be coupled to final action. The common schema also included the initial impulse, candidate actions, final action, rationale and provenance.

\begin{figure}[p]
\centering
\includegraphics[width=\textwidth]{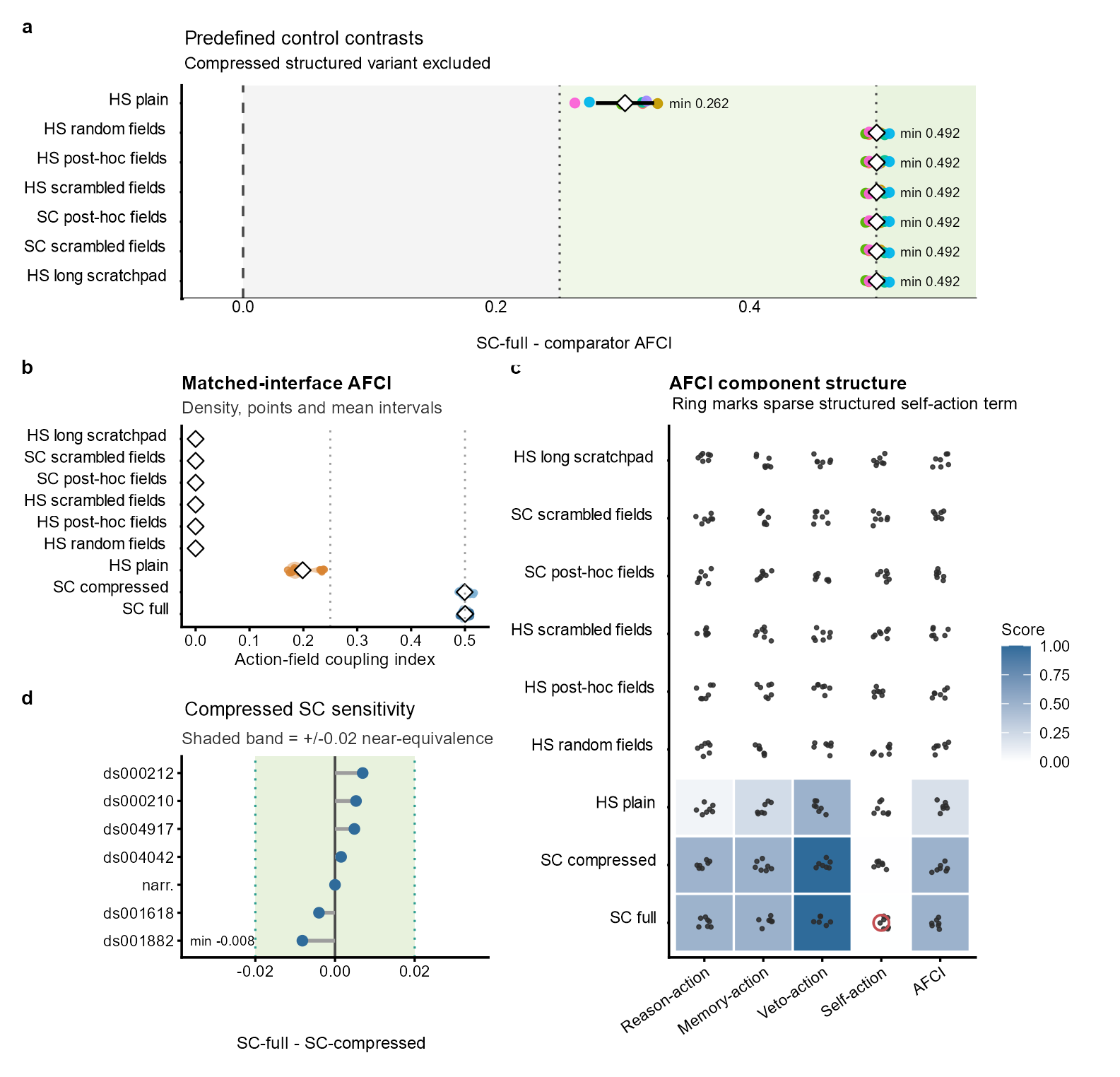}
\caption{Matched reporting formats do not by themselves produce trace-action coupling. The action-field coupling index (AFCI) tests whether emitted reason, memory, self-state and veto information is coupled to final action under a common schema. \textbf{a}, Predefined control-comparator contrasts with dataset points, mean intervals, minimum contrasts and reference bands. The seven control comparators are enumerated by exact protocol identifier in Methods and the Supplementary registry. \textbf{b}, AFCI distributions across dataset-replicate units, with density envelopes, raw points and mean intervals. The compressed structured variant is shown as a positive sensitivity comparison excluded from the predefined control-comparator set. \textbf{c}, AFCI component decomposition into reason-action, memory-action, veto-action, self-action and composite terms; the ring marks the sparse self-action term for the structured matched-interface variant. \textbf{d}, Dataset-wise structured--compressed differences; the shaded band marks a \(\pm 0.02\) near-equivalence region. The AFCI panels summarize action-field coupling under the common schema.}
\label{fig:matched_interface_afci}
\end{figure}

We call this trace-action coupling, measured here with the action-field coupling index (AFCI). The structured matched-interface variant exceeded seven predefined control comparators on AFCI in all seven datasets: HS plain, HS random fields, HS post-hoc fields, HS scrambled fields, SC post-hoc fields, SC scrambled fields and HS long scratchpad. The dataset-cluster bootstrap lower 95\% bound for the AFCI difference was 0.4519. The compressed structured variant was analysed separately as a positive sensitivity control and was not part of the negative-control set. Because the scorer tracks whether reports were random, post-hoc or scrambled, AFCI measures coupling rather than reporting-format availability alone (Figure~\ref{fig:matched_interface_afci}).

For scored generation \(i\), AFCI decomposed the matched-interface trace into four bounded coupling terms:
\begin{equation}
\mathrm{AFCI}_{i,v}=\frac{1}{4}\left(R_{i,v}+M_{i,v}+Q_{i,v}+S_{i,v}\right),
\end{equation}
where \(R_{i,v}\) denotes reason-action alignment, \(M_{i,v}\) memory-action alignment, \(Q_{i,v}\) veto-action alignment and \(S_{i,v}\) self-state/action alignment. Each term is a protocol-level score on \([0,1]\) computed from the emitted trace information, final action and predefined provenance information under the common schema. This design makes AFCI appropriate as a matched-format coupling readout, while leaving the finite-action-code benchmark to provide the primary format-independent behavioral test. For compact notation, \(\mathrm{SC\_full}\) denotes the structured matched-interface variant. For a comparator \(q\), the dataset-level AFCI contrast was
\begin{equation}
\Delta_d(q)=\frac{1}{|\mathcal{I}_{d,\mathrm{SC\_full}}|}\sum_{i\in\mathcal{I}_{d,\mathrm{SC\_full}}}\mathrm{AFCI}_{i,\mathrm{SC\_full}}
-\frac{1}{|\mathcal{I}_{d,q}|}\sum_{i\in\mathcal{I}_{d,q}}\mathrm{AFCI}_{i,q}.
\end{equation}
The reported lower 95\% bound is the 2.5th percentile of a dataset-cluster bootstrap distribution over \(\Delta_d(q)\), following the standard non-parametric bootstrap logic for clustered empirical summaries \citep{efron1993bootstrap}.

The AFCI component pattern constrains interpretation and prevents overinterpreting the composite. The structured matched-interface variant had reason-action and memory-action alignment near 0.5, veto-action alignment near 1.0 and a sparse self-action alignment term near 0.003. This pattern is compatible with the high finite-action-code behavioral self-continuity score, \(B_{\mathrm{SCI}}\), because the two quantities operationalize different levels of control: the AFCI self term measures sparse trace/action coupling, whereas the behavioral self-continuity probes test whether the final action code remains stable under the predefined self-continuity manipulation and irrelevant-cue controls. The composite AFCI is therefore interpreted together with its component terms. Extended Data Figure 2 gives the full dataset- and replicate-resolved AFCI profile.
\FloatBarrier

\subsection*{External API, perturbation and audit robustness}

After the open-weight state-lesion tests, additional validation blocks tested whether the binding signature was brittle to prompt wording, perturbation design, provider interface or scorer awareness. Prompt paraphrases, expanded counterfactual probes and placebo-component manipulations preserved the direction of the result within the tested scope. The perturbation set covered reason flips, reason-strength gradients, memory conflict, irrelevant memory, veto cues, late veto cues, self-continuity probes, placebo cues and adversarial-randomness cues. In the formal holdout, prompt paraphrasing remained within the registered tolerance for structured accuracy. The deterministic counterfactual-transform audit used a text--label consistency screen: candidate rows with visible label-context contradictions were retained as transform-quality boundary rows, and the consistency-passing transform regenerated the decisive context and hidden expected action together. The consistency-passing transform satisfied the predefined criterion across 120/120 scaffold--family cells, with mean structured accuracy 1.000.

The independent Mistral-7B family reproduced the three core no-context controls in 60/60 cells each, and the Qwen2.5 size series preserved the no-field, scrambled-context and distribution-matched control separations across 7B, 14B and 32B settings.

Targeted lesions were more diagnostic as boundary conditions. Ordinary target lesions were positive on average across the four-model matrix but weaker than the core no-context controls (193/240 positive cells; mean delta 0.041), and therefore are not interpreted as a universal collapse test. A stricter Qwen2.5-32B lesion slice removed the decisive intervention context more completely and produced 28,800 additional scored generations: structured accuracy exceeded the strict lesion in 60/60 cells, with structured mean 0.998, strict-lesion mean 0.037 and mean delta 0.961. Thus the ordinary 32B target-lesion weakness is treated as a lesion-design boundary condition, whereas the strict slice supports the narrower claim that removing decisive intervention context collapses the 32B action-control signature.

A formal control-audit slice added token-prompt and entropy-token control variants for the Qwen open-weight models and retained tokenizer and compute-proxy metadata. The main formal slice completed 63,360 planned rows with no duplicate analysis keys and a 0.140\% parse/error rate, and structured accuracy exceeded the no-fields, scrambled-context, distribution-matched, strict-lesion and formal control variants in the relevant scaffold--task-family cells. A 17,280-row token-prompt slice aligned the prompt-token count for the entropy-token variant, but total-token ratios and latency proxies remained outside the strict token/compute criterion. An entropy-prior no-field behavioral control used a calibration-frozen final-action prior and completed 24,480 planned rows with no duplicate keys and a 0.102\% parse/error rate. It remained behaviorally below the structured-control variant in 120/120 cells (mean delta 0.660), but the registered \(\leq 0.10\)-bit strict entropy-gap criterion was satisfied in only 1/6 model--scaffold cells. The interpretation is therefore limited to token-prompt controls with compute audit and an entropy-prior behavioral control, not strict compute or strict entropy matching.

A separate local forced-budget audit addressed the narrower question of whether the state-binding separation survives when prompt length, completion length, total-token count, latency and canonical-action entropy are simultaneously matched in an inspectable open-weight run. Using cached local Qwen2.5-1.5B-Instruct CPU inference over 20 task families and two events per family, prompts were padded to exact tokenizer equality by event and generation length was fixed. The audit passed its local gate: prompt-token match rate 1.000, completion-token match rate 1.000, total-token ratio 1.000, latency ratio 0.997 and canonical-action entropy gap 0.020 bits. Under those matched conditions, structured-state accuracy was 0.900, whereas the distribution-matched action-prior/no-state control was 0.325. This is a bounded forced-budget result, not a replacement for the failed original formal strict gates across larger open-weight and provider-served runs.

The cross-model subset used four externally served model identifiers recorded as provider-reported identifiers at run time: \texttt{gpt-5.4-mini}, \texttt{gpt-4.1-mini}, \texttt{deepseek-chat} and \texttt{gemini-2.5-flash}. Each model showed the structured-control variant exceeding the high-stochasticity comparator in 7/7 dataset units. The formal API validation slice used provider-reported run-time identifiers \texttt{gpt-5.4-mini}, \texttt{gemini-3.1-flash-lite-preview}, \texttt{deepseek-v3.2} and \texttt{qwen3-max}, completing 14,400 planned rows and 11,520 fallback-free remote calls, with zero unrecovered generations and one parse/action error record. Structured accuracy exceeded the no-fields, scrambled-context, distribution-matched and strict-lesion controls in 238/240, 240/240, 240/240 and 240/240 cells, respectively; all cluster-bootstrap lower 95\% bounds were positive. The blinded audit expanded to 1,200 full-overlap items scored by three annotators, with variant identity withheld \citep{cohen1960coefficient}. Core dimensions had 100\% valid scoring, no unscorable items and no violations of the blinding protocol; mean pairwise Cohen's kappa was 0.961, and the lowest non-constant dimension kappa was 0.860. Automated judge scores are retained only as auxiliary consistency summaries, given known position, verbosity and self-enhancement biases \citep{zheng2023judge}. Formal API, calibration and blinded-audit details are retained as Supplementary scope checks. These experiments test robustness of the separation across the tested settings; provider-reported identifiers are retained as run-time identifiers, not theoretical model categories.

\section*{Discussion}

This study reframes agent control evaluation as a causal state-binding measurement problem. The central object is whether event-specific decisive state is bound to final action: actions should follow reason, memory, veto and persistent-state interventions while remaining stable under irrelevant cues. Across the tested protocols, this binding score predicted hidden-target action behavior more directly than output entropy, action-prior matching or rationale format.

This matters because modern agents already contain reasoning, memory, planning and constraint-like modules. The question for evaluation is no longer simply whether such modules can improve behavior, but whether a benchmark can distinguish true event-specific state--action binding from surface alternatives. The hidden-target benchmark and open-weight state-lesion controls address that question by testing no-field prompts, scrambled decisive context, distribution-matched no-context action priors and strict target lesions under a common finite-action ontology.

The evidence is hierarchical. The finite-action benchmark defines the measurement paradigm because it removes free-form rationale wording and compares final action codes with hidden scorer-side targets. The open-weight matrix provides the main model-family validation evidence, with four local model families/sizes tested across scaffolds and task families. The strict-lesion slice supplies the clearest information-boundary result: removing the decisive intervention-carrying field collapses the action-control signature in the Qwen2.5-32B setting. Matched-interface AFCI analyses then test trace faithfulness under common schemas, and API validation provides external robustness rather than the primary model-family claim.

The sufficiency experiment adds the positive side of the information-boundary result within a finite-action setting. Removing the decisive state collapses action control, whereas exposing the minimal current-event decisive field recovers the prescribed action pattern in the tested API models. The local wrapper result is correspondingly practical: a finite-action guard that checks current-event decisive-field consistency can prevent binding to a scrambled decisive field without adding another model call. This remains a finite-action result, but it turns the metric into an auditable control layer that can be tested in future task environments.

The broader lesson is methodological rather than metaphysical: evaluations of language agents should measure whether decisive state is bound to action, not only whether the system produces diverse outputs, plausible rationales or high task-level performance. A system can match an action prior or emit a fluent reason without selecting actions for the event-specific reason, memory or constraint that should control the decision.

Several limitations define the scope of the result. The claim is protocol-level and concerns measured trace and action records. The finite action ontology is intentionally narrow. The validation experiments cover the tested prompts, scaffolds, model families and provider-reported identifiers, not all language agents or all architectures. The original formal entropy calibration did not meet the registered closeness criterion, and the historical token-prompt slice did not satisfy the strict token/compute criterion because total-token ratios and latency proxies remained outside the criterion. The new local forced-budget Qwen2.5-1.5B audit closes that objection only in a bounded CPU-local setting with exact prompt padding and fixed generation length. The constructive wrapper is tested only as a local finite-action guard over the present intervention records. The SWE-bench Lite validation uses real issue records and supports oracle-free implementation-file localization; repository execution, patch application and official test-suite resolution remain outside the primary endpoint. The exploratory execution slice, including a later apply-check/repair filter, is reported as a scope check rather than as evidence for SWE-bench issue resolution. The manuscript does not include full AgentBench, WebArena, OSWorld or SWE-bench task-execution outcomes. The study makes no general claim about model agency, human-like cognition or the causal status of reasoning.

In the tested language-agent protocols, causal state binding provides a measurable action-control signature: final actions track decisive event-specific state and fail when that state is removed, scrambled or replaced by an action prior.

\section*{Methods}

\subsection*{Agent family and lesions}

The agent family was designed to make component lesions explicit. Its variants selectively enable or remove branching, self-state, memory, reason-graph and veto components. The reader-facing names used in Results are aliases for exact protocol identifiers. The structured-control variant corresponds to protocol id \texttt{A4}: branching, self-state, memory, reason and veto are enabled. The high-stochasticity comparator corresponds to protocol id \texttt{A5}: branching and high-temperature sampling are enabled while self-state, memory, reason and veto are disabled. This A4--A5 contrast is an informative baseline but is not treated as the sole evidence for causal state binding because it gives the structured protocol more intervention-relevant information. The primary state-binding controls are the finite-action no-field, scrambled-context, distribution-matched no-context and strict target-lesion contrasts, which isolate whether the event-specific decisive state is available and correctly paired with the action query. The reason and veto lesions correspond to \texttt{A4\_no\_reason} and \texttt{A4\_no\_veto}, respectively. Additional stochastic controls include lower-temperature, compute-control-attempt, self-consistency and empirical entropy-attempt variants. The Supplementary Information registry retains the exact raw protocol identifiers; these identifiers are implementation labels, not established theoretical categories or latent psychological types.

Canonical protocol identifiers are retained where they are needed to reproduce the implemented variant grid. The compact registry below groups the identifiers that recur across evidence blocks; their interpretation is defined by the component settings reported here.
\begin{center}
\scriptsize
\resizebox{\textwidth}{!}{%
\begin{tabular}{lll}
\toprule
Canonical label group & Role & Naming note \\
\midrule
\texttt{A4}, \texttt{A5} & primary structured and stochastic variants & baseline lesion and behavior benchmarks \\
\texttt{A4\_no\_reason}, \texttt{A4\_no\_veto} & targeted structured-control ablations & reason and veto component tests \\
\texttt{A4\_full}, \texttt{A4\_compressed} & matched-interface positive variants & compressed variant is excluded from the control-comparator set \\
\texttt{A5\_plain}, \texttt{A5\_schema\_*}, \texttt{A4\_posthoc\_fields}, \texttt{A4\_scrambled\_fields}, \texttt{A5\_long\_scratchpad} & matched-interface control comparators & full names are enumerated in the Supplementary registry \\
\texttt{A5\_entropy\_attempt}, \texttt{A5\_empirical\_entropy\_attempt} & entropy checks & predefined entropy-closeness criterion unsatisfied \\
\bottomrule
\end{tabular}
}
\end{center}

Each variant can be represented as a protocol component vector
\begin{equation}
\mathbf{z}_v=(b_v,s_v,m_v,r_v,q_v,T_v),
\end{equation}
where \(b_v\in\{0,1\}\) indicates branching, \(s_v\in\{0,1\}\) self-state, \(m_v\in\{0,1\}\) memory, \(r_v\in\{0,1\}\) reason-graph construction, \(q_v\in\{0,1\}\) veto and \(T_v\) the sampling temperature. The core lesion contrasts hold all non-targeted components fixed as closely as the protocol permits and toggle one component or stochastic setting. Thus \(\texttt{A4\_no\_reason}\) differs from the structured-control protocol id \texttt{A4} by setting \(r_v=0\), \(\texttt{A4\_no\_veto}\) differs by setting \(q_v=0\), and the high-stochasticity protocol id \texttt{A5} differs by disabling \(s_v,m_v,r_v,q_v\) while increasing stochastic sampling.

Each generation produced both a trace record and a behavior record. For reproducibility, the JSON-oriented chat protocol requested the raw keys \texttt{branch\_set}, \texttt{self\_state}, \texttt{reason\_graph}, \texttt{first\_impulse}, \texttt{reason\_update}, \texttt{final\_action}, \texttt{memory\_write}, \texttt{veto\_state}, \texttt{delayed\_recall} and \texttt{stage\_vector}. The model-visible prompt contained the event text, dataset identifier, event granularity, enabled module flags and recent memory. Provider outputs were parsed as JSON and merged with implementation defaults when fields were absent or malformed, with provider-level interface categories retained for reproducibility reporting. The Supplementary Information gives the complete component settings for the agent family.

\texttt{A5\_entropy\_attempt} denotes a lower-temperature stochastic attempt, whereas \texttt{A5\_empirical\_entropy\_attempt} denotes the selected temperature/top-p candidate from the empirical calibration analysis; the predefined entropy-closeness criterion was not satisfied for these variants.

\subsection*{Prompt corpora and generation protocol}

The study uses dataset namespaces as prompt/event corpora for agent evaluation. The seven namespaces were \texttt{narratives\_formal}, \texttt{ds000210\_self\_loop}, \texttt{ds001618\_self\_loop}, \texttt{ds001882\_choice}, \texttt{ds000212\_choice}, \texttt{ds004917\_uncertainty} and \texttt{ds004042\_recall}. Event lists were sampled with stratification over available task, condition, run, episode and subject-like columns when such columns existed; otherwise, events were sampled and then ordered by available dataset, episode, subject, onset and event identifiers. We use ``replicate index'' for repeated generation runs; deterministic provider-side seed control was not assumed unless explicitly supported by the inference endpoint. The Supplementary Information summarizes the prompt-corpus roles and construct pressures.

The baseline lesion matrix used eight variants, three replicate indices and up to 500 sampled events per dataset. The realized analysis comprised 74,352 model queries and 74,336 model generations; 16 generations were unrecovered, for an unrecovered-generation rate of 0.000215. The provider-reported primary model was \texttt{gpt-5.4-mini}; \texttt{qwen3-30b-a3b} was available as the secondary provider setting. Primary trace metrics were raw final-action unpredictability, SCI, RSI, VEI and ACI. Unpredictability was Shannon entropy over final actions. SCI averaged identity and continuity weights in the agent self-state. RSI measured whether a non-empty reason graph was associated with a change from first impulse to final action. VEI measured whether veto was applied. ACI normalized memory depth. The structured composite averaged SCI, RSI, VEI and ACI. These metrics are operational summaries of the trace protocol.

\subsection*{Matched-interface action-field coupling controls}

Matched-interface controls required all predefined control comparator variants to use the same schema while changing how the reporting elements were generated or coupled to final action; the compressed structured protocol id \texttt{A4\_compressed} used the same schema as a positive sensitivity variant. The common schema included \texttt{first\_impulse}, \texttt{candidate\_actions}, \texttt{reason\_graph}, \texttt{memory\_trace}, \texttt{self\_state}, \texttt{veto\_state}, \texttt{final\_action}, \texttt{final\_action\_rationale} and a provenance record. The realized experiment comprised 26,946 model generations with zero unrecovered generations. The primary score was the action-field coupling index (AFCI), defined from reason-action, memory-action, veto-action and self-action alignment terms. The AFCI scorer uses emitted content, first impulse, final action and provenance information to penalize post-hoc, scrambled, random or template-generated reporting elements. AFCI is read as a matched-format coupling measure, and the finite-action-code benchmark supplies the format-independent behavioral test.

\subsection*{Finite-action-code behavioral benchmark}

The finite-action-code benchmark constrained \texttt{final\_action} to six codes: \texttt{ACTION\_A}, \texttt{ACTION\_B}, \texttt{VETO}, \texttt{DEFER}, \texttt{RECALL\_PRIOR} and \texttt{INVALID\_OR\_UNMAPPED}. The model-visible prompt contained task context, probe manipulation, module-lesion rule and allowed action ontology. Expected-action labels were scorer-side annotations and were not included in the model-visible prompt; the model was shown the task context, manipulation and allowed output ontology, but not the hidden post-manipulation target label (\texttt{expected\_action\_after}) used for scoring. Scoring compared the post-generation \texttt{canonical\_action\_code} with that hidden label. The canonicalizer maps explicit option mentions to \texttt{ACTION\_A} or \texttt{ACTION\_B}, stop/withhold/cancel language to \texttt{VETO}, delay language to \texttt{DEFER}, and recall or prior-commitment language to \texttt{RECALL\_PRIOR}; all other outputs are assigned \texttt{INVALID\_OR\_UNMAPPED}. The design specified 58,896 model queries and retained 58,835 model generations with one unrecovered generation. Predefined deduplication and inclusion rules retained one scored row per exact \texttt{dataset/base\_event/condition/variant/replicate} key, collapsing duplicate source-event identifiers and repeated records before inference. The resulting analysis used 57,816 deduplicated scored records. The parse-error rate was \(1.70\times 10^{-5}\), the unmapped-action rate was 0.001297 and the structured-control irrelevant-cue false-positive rate was 0.008302; unmapped actions are outputs that could not be mapped to the finite action alphabet. The predefined evaluation criterion required parse errors below 2\%, unmapped actions below 2\%, structured-control irrelevant-cue false positives below 15\%, structured-control behavioral-composite superiority over the high-stochasticity comparator in at least 6/7 datasets, component-specific structured-control superiority over the relevant ablations in at least 6/7 datasets, and a positive dataset-cluster bootstrap lower bound.

\subsection*{Minimal causal-state sufficiency and wrapper experiment}

The sufficiency experiment used the same 20 synthetic open-weight task families and finite four-action ontology used in the API validation controls, but changed the model-visible state available to the action query. For each family, six event indices were evaluated. The five sufficiency conditions were: full state, only the decisive causal field, surface context only, action prior only and scrambled decisive field. A sixth irrelevant-cue condition added a misleading surface cue while preserving the current event's decisive state. The two API models were provider-reported run-time identifiers \texttt{gpt-5.4-mini} and \texttt{qwen3-max}. Each model generated 120 rows per condition, for 720 rows per model and 1,440 rows total. Prompts requested a compact JSON object with \texttt{final\_action} and \texttt{rationale}. Provider metadata retained remote-call status, prompt tokens, completion tokens, total tokens, latency, temperature, top-p and parse/error status; API credentials were read from a private local file or environment variable and were not written to outputs.

The only-decisive-field condition exposed a single current-event causal field, for example a reason field selecting the action, a corrected-memory field, a persistent-state field or a veto/constraint field. The action-prior-only condition exposed only a final-action prior sampled from a distribution-matched marginal schedule and no event-specific decisive state. The scrambled-field condition exposed a decisive field sampled from a different event with a different expected action, preserving field-like content while breaking the event-specific binding. Accuracy was the match between the canonicalized final action and the current event's expected action. The sufficiency recovery fraction was \((A_{\mathrm{only}}-A_{\mathrm{control}})/(A_{\mathrm{full}}-A_{\mathrm{control}})\), where \(A_{\mathrm{control}}\) is the best non-decisive-control accuracy.

The local wrapper was evaluated post-generation on the same records. It accepted a generated action if the visible decisive field belonged to the current event and matched the action; otherwise it corrected the action to the current visible decisive field. When the visible decisive field came from a different event, the wrapper deferred instead of accepting that field. For conditions without a decisive field, the wrapper made no change. This wrapper is a deterministic finite-action guard with no additional model call; it is reported as a constructive intervention within the finite-action protocol, not as deployed-agent validation.

\subsection*{SWE-bench Lite issue-to-file validation}

The SWE-bench Lite predictive-validity validation used \texttt{princeton-nlp/SWE-bench\_Lite}, split \texttt{test}, and retained all 300 issue records. The endpoint was issue-to-file localization; repository execution was outside the primary validation. Six provider-served model identifiers were evaluated: \texttt{gpt-3.5-turbo}, \texttt{gpt-4o-mini}, \texttt{qwen-turbo}, \texttt{qwen-plus-latest}, \texttt{deepseek-chat} and \texttt{gemini-2.5-flash-lite}. For each model--issue pair, the task-only condition used one low-temperature call plus two self-consistency calls; diagnostic conditions exposed current-state, decisive-only, scrambled-state, prior-only/no-state and irrelevant-cue fields. The predictor was oracle-free: model-visible decisive state used issue text, repository-visible file trees, fixed retrieval candidates and prompt-visible localization state. Benchmark reference implementation files were withheld from prompts and used only by the scorer.

The primary outcome was constraint-clean implementation-file hit@3: at least one predicted implementation file had to match a benchmark reference implementation file, while avoiding hard violations such as nonexistent paths, wrong-repository paths, test-only targets, empty/no-op targets and scrambled-state following. The prespecified primary test compared a full non-CSB baseline against the same baseline augmented with the causal state-binding composite. The baseline included model identity, repository, retrieval rank/recall, issue length, reference-file count, action entropy, self-consistency vote margin, rationale length and verbal confidence. In the merged 300-record run, 18,000 condition/repeat rows were observed and aggregated to 1,800 model-issue records for the primary AUC analysis. The baseline AUC was 0.873 and the augmented AUC was 0.935, giving \(\Delta\)AUC 0.062 with issue-cluster bootstrap lower 95\% bound 0.025. The state-binding coefficient was positive, and leave-one-repository, leave-one-model and within-model sensitivities were positive in 12/12, 6/6 and 6/6 analyses, respectively. A binding-guard arm reduced hard-constraint violations without degrading hit@3 beyond the prespecified margin. These results support issue-to-file localization reliability; issue-resolution claims require patch generation and successful test-suite execution.

The retained SWE-bench Lite set contained 12 repository identifiers. Leave-one-repository sensitivity therefore reports 12 held-out repository folds. Within-repository sensitivity requires both positive and negative localization outcomes inside the repository-specific slice; 11 repositories satisfied that condition, whereas one retained repository had no positive constraint-clean hit@3 rows and was excluded from that within-repository denominator.

For the frozen 50-record execution slice, patch-prediction JSONL files were generated for all six models using oracle-free prompts with repository-visible source context. Two selected models were evaluated with the official SWE-bench harness, and a later patch-generation v3 pass applied a local \texttt{git apply --check} screen plus one oracle-free repair attempt before rerunning the official harness on applicable patches. These execution-slice analyses are reported in the Supplementary Information as exploratory scope checks separate from the primary localization endpoint.

\subsection*{Stochastic controls and entropy checks}

Strong stochastic controls tested whether increased stochastic opportunity, lower-temperature attempts, a compute-control attempt from the original protocol or self-consistency sampling could recover the structured-control profile. The corresponding raw protocol identifiers are retained only in the Supplementary Information registry; the full run executed 15,490 model generations with zero unrecovered generations. The compute-control-attempt label is descriptive, not evidence that the strict token/compute criterion was satisfied. Entropy checks were computed over retained behavior rows using raw final-action strings, rule-canonical actions, term frequency-inverse document frequency (TF-IDF) character n-gram semantic clusters with cosine-similarity threshold 0.90, and coarse action families. The separation between raw strings, canonical actions and semantic clusters follows from the fact that stochastic decoding alters surface-form distributions through temperature, top-\(k\), top-\(p\) and related sampling choices, whereas semantic uncertainty need not track lexical variability one-for-one \citep{fan2018hierarchical,holtzman2020degeneration,kuhn2023semantic}. These checks are reported for experiments with retained behavior rows; the baseline matrix is interpreted using its predefined raw-action entropy. The entropy-calibration analysis was interpreted under a predefined closeness criterion of 0.15 across raw/canonical/semantic/action-family entropy gaps. The selected finite-action-code calibration candidate is recorded descriptively in the reproducibility tables. Because this criterion was not satisfied, the entropy analysis is treated as a check rather than an entropy-matched stochastic-control claim.

The local strict matching audit was a separate forced-budget analysis designed for inspectability rather than scale. It used cached local Qwen2.5-1.5B-Instruct inference on CPU, a fixed tokenizer, prompt padding to exact tokenizer equality by event, and \(\texttt{min\_new\_tokens}=\texttt{max\_new\_tokens}=32\). The two conditions were \texttt{structured\_state}, which exposed the event-specific decisive field, and \texttt{entropy\_prior\_no\_state}, which exposed only a distribution-matched action prior and no event-specific decisive state. The gate required event-wise prompt-token equality, event-wise completion-token equality, total-token ratio within \([0.95,1.05]\), latency ratio within \([0.50,1.50]\) and canonical-action entropy gap \(\leq0.15\) bits. The run logged tokenizer counts, completion counts, total-token counts, latency and CPU memory traces; CUDA and GPU-memory logging were unavailable in the local environment because CUDA was not available. This audit is reported as a bounded local strict-matching result and does not retroactively convert the earlier formal entropy/token attempts into passed strict matches.
\subsection*{Open-weight and API finite-action validation controls}

Open-weight validation used the same finite-action-code scoring rule across Qwen2.5-7B-Instruct, Qwen2.5-14B-Instruct, Qwen2.5-32B-Instruct and Mistral-7B-Instruct-v0.3. Each model contributed 126,000 scored generations over three agent scaffolds, 20 task families, seven variants and three seeds. The three scaffolds were prompt-level implementations of the same finite-action ontology: a ReAct-style scaffold, a planner/executor scaffold and a memory-reflection scaffold. The 20 task families crossed reason, memory, veto, self-continuity, irrelevant-cue and adversarial-randomness probes with finite-action decision contexts. The seven variants were structured, stochastic full-context, stochastic no-fields, stochastic scrambled-context, distribution-matched offline/no-context, target lesion and strict target lesion where available.

The no-fields control removes structured state fields while preserving the visible action ontology and surface task context. The no-context control is the family of controls that withhold the decisive intervention-carrying context from the model-visible action-selection query. The scrambled-context control pairs each action-selection query with intervention context sampled from a different event, preserving surface complexity while breaking the event-specific link between decisive context and hidden expected action. The distribution-matched offline/no-context control samples final actions from the structured variant's empirical action prior within each task family but does not observe the decisive intervention context. The strict target lesion removes the decisive intervention-carrying field while preserving non-decisive context and the action ontology. The entropy-prior no-field behavioral control is a calibrated no-field variant that exposes only surface context and a final-action prior selected from a calibration split; it is not interpreted as strict entropy matching unless the registered entropy-gap criterion is met.

Prompt paraphrases tested whether equivalent prompt variants changed the direction of the structured-control result, with 11,400 model generations and zero unrecovered generations in the auxiliary perturbation block. Counterfactual probes perturbed reason, reason strength, memory, veto, self-continuity, irrelevant and adversarial-randomness cues; the expanded counterfactual-probe corpus retained 30,584 trace/behavior rows with zero unrecovered generations. Placebo components included synthetic and negative-control manipulations designed to reveal generic structured-control false positives and ran 9,960 model generations with zero unrecovered generations. The formal transform audit used deterministic prompt-paraphrase and counterfactual-flip transforms. Prompt paraphrases remained within the predefined tolerance. For counterfactual-flip transforms, text--label consistency screening retained candidate rows with visible label-context contradictions as transform-quality boundary rows; the consistency-passing transform regenerated the visible decisive context and hidden expected action together before the holdout evaluation.

Cross-model subset replication used provider-reported run-time model identifiers \texttt{gpt-5.4-mini}, \texttt{gpt-4.1-mini}, \texttt{deepseek-chat} and \texttt{gemini-2.5-flash}; each model contributed 1,995 model generations, and the Gemini subset contained two malformed-output records while preserving the same dataset-level direction. The open-weight reference model used Qwen2.5-14B-Instruct in local inference, producing 10,500 local generations, eight parse diagnostics and parse-error rate 0.000762; the structured-control AFCI \(>\) high-stochasticity AFCI criterion was met in 7/7 datasets. A stricter Qwen2.5-32B target-lesion slice added 28,800 scored generations over \texttt{structured} and \texttt{target\_lesion\_strict}. Formal Qwen control-audit slices added token-prompt and entropy-token variants, tokenizer token counts, latency metadata and GPU memory metadata. The API validation slice used the same finite-action control definitions for provider-reported run-time identifiers \texttt{gpt-5.4-mini}, \texttt{gemini-3.1-flash-lite-preview}, \texttt{deepseek-v3.2} and \texttt{qwen3-max}; the exact request metadata and manifests retain the identifiers returned or accepted by the endpoint at run time.

\subsection*{Blinded audit}

Three annotators independently scored 1,200 full-overlap items under blinded conditions with respect to variant identity; agreement was summarized with pairwise Cohen's kappa \citep{cohen1960coefficient}. The dimensions included final-action reason consistency, memory respect, veto appropriateness, first-impulse veto change, post-hoc appearance, verbose-template behavior without control and inappropriate irrelevant-cue change. Core dimensions had 100\% valid scoring, no unscorable items and no violations of the blinding protocol. Mean pairwise Cohen's kappa across dimensions was 0.961, and the lowest non-constant dimension kappa was 0.860. The annotation exercise was limited to expert scoring of model outputs and did not collect personal data or involve intervention with human participants. Automated judge scoring is reported as an auxiliary consistency check.

\subsection*{Statistical inference}

The primary inference hierarchy was dataset-level direction, dataset-cluster bootstrap and component-specific ablation contrasts. Mixed-effect fits were retained as auxiliary summaries when stable; singular, boundary or non-convergent fits were kept outside the primary inferential basis. The finite-action-code behavioral benchmark used one row per unique dataset, event, condition, variant and replicate after predefined deduplication and inclusion rules; the exact reproducibility key was \texttt{dataset/base\_event/condition/variant/replicate}. Because the main cluster count is seven dataset namespaces, the Supplementary Information also reports exact sign tests, leave-one-dataset-out means, top-effect-removed means and minimum dataset contrasts for the primary evidence blocks. For the open-weight finite-action matrix, the unit was the model--scaffold--task-family control contrast; summaries report model-, scaffold- and task-family-cluster bootstrap intervals, leave-one-cluster-out sensitivity and top-effect-removed means. Formal control-audit criteria covered completeness, parse-error, entropy-gap, token/compute, behavioral, robustness and API-readiness checks; criteria that were not satisfied are reported as boundary conditions rather than retuned passes, and the consistency-passing counterfactual transform is reported separately from the transform-quality boundary rows.

The primary statistical unit is the dataset namespace. This choice is conservative for an evaluation study in which many rows share prompt families, variant definitions and scoring rules. Directional dataset counts, dataset-cluster bootstrap intervals, sign tests and leave-one-dataset-out summaries are reported as complementary checks at the dataset-unit level, following concerns about significance testing and unit choice in natural language processing (NLP) evaluation \citep{dror2018hitchhiker}.

For a metric \(m\), comparator \(q\), and dataset \(d\), the paired dataset effect was
\begin{equation}
\delta_d(q;m)=\bar{m}_{d,\mathrm{SC}}-\bar{m}_{d,q},
\end{equation}
with \(\mathrm{SC}\) replaced by \(\mathrm{SC\_full}\) for matched-interface AFCI where appropriate. The primary reported effect was the dataset-average contrast
\begin{equation}
\bar{\delta}(q;m)=\frac{1}{|\mathcal{D}|}\sum_{d\in\mathcal{D}}\delta_d(q;m).
\end{equation}
Cluster-bootstrap intervals were computed by resampling dataset identifiers with replacement and recomputing \(\bar{\delta}^{*(b)}(q;m)\) for bootstrap draw \(b\); the lower 95\% bound is the empirical 2.5th percentile of this bootstrap distribution \citep{efron1993bootstrap}. Auxiliary mixed-effect summaries followed the usual random-effects logic for nested or repeated observations \citep{laird1982random}, but the paper's interpretation is based on the dataset-direction and cluster-bootstrap criteria described above.

\section*{Ethics and human-subjects statement}

Human annotators contributed output-level quality-control labels. No personal, behavioral, health or intervention data were collected from annotators or any other individuals. The annotation was limited to expert scoring of model outputs under a fixed rubric and blinded variant identity. The empirical units for the experimental analyses were synthetic or license-clean prompts, public prompt/event scaffolds and language-model outputs. An institutional non-human-subjects determination is available.

\section*{Data availability}

All claim-supporting derived data, source-data files, run manifests, scorer definitions, prompt templates, sampled event lists, criterion summaries, blinded-audit materials, model and hardware manifests and redistributable trace/behavior records are provided in a Zenodo archive: \reviewarchiveurl. The archive contains figure source data, table CSV exports, compact result summaries, manifests, SHA-256 checksums, environment notes, minimal rebuild commands, a clean-rebuild audit and an excluded-materials statement. Third-party source files not licensed for redistribution, private credentials and local configuration secrets are not included; derived summaries from those sources are included when permitted.

\section*{Code availability}

Analysis code, configuration templates, scorer definitions, prompt templates, run manifests, validation scripts and figure-generation code are provided in the same Zenodo archive named in Data availability. The archive includes the core run scripts, implementation modules, R figure-generation code and a clean-rebuild audit that regenerates the main validation figure and checks Table 17 and SWE-bench claim-support summaries. Private provider credentials and local configuration secrets are excluded, following transparent evaluation, model-reporting and dataset-documentation practice \citep{liang2023helm,mitchell2019modelcards,gebru2021datasheets}.

\section*{Large language model use}

The author used language-model tools during manuscript preparation for editorial assistance, consistency checks and local code navigation. The author reviewed all changes and remains responsible for the manuscript content, analyses, claims and cited sources. No large language model is listed as an author.

\section*{Acknowledgements}

We thank the maintainers of the public datasets and open methodological resources used in this study.

\section*{Author contributions}

The author is responsible for the manuscript and correspondence.

\section*{Competing interests}

The author declares no competing interests.
\clearpage
\appendix
\section*{Supplementary Information}
\section*{Supplementary Overview}
This Supplementary Information reports the dataset-resolved evidence, metric definitions and reproducibility details for the causal state-binding study. The main text introduces reader-facing names such as causal state binding, structured agent condition, decisive intervention state, trace-action coupling and hidden-target finite-action probes; the alphanumeric labels in this supplement are raw protocol identifiers retained for exact export matching. The tables cover the implemented agent family, prompt/event corpora, finite-action-code behavioral benchmark, open-weight no-context and strict-lesion validation, minimal causal-state sufficiency, local control-binding wrapper checks, local forced-budget entropy/token matching, a SWE-bench Lite issue-to-file localization validation with reliability calibration and threshold analyses, matched-interface action-field coupling controls, baseline lesion matrix, stochastic and entropy-calibration analyses, API evidence, blinded audit and interface diagnostics. Values shown in the PDF tables are rounded for readability; the CSV exports retain the full exported precision. Entropy summaries, resampling intervals and agreement statistics follow standard empirical entropy, bootstrap and nominal-agreement conventions \citep{shannon1948mathematical,efron1993bootstrap,cohen1960coefficient}.

\section*{Extended Data Figures}

\begin{figure}[p]
\centering
\includegraphics[width=\textwidth]{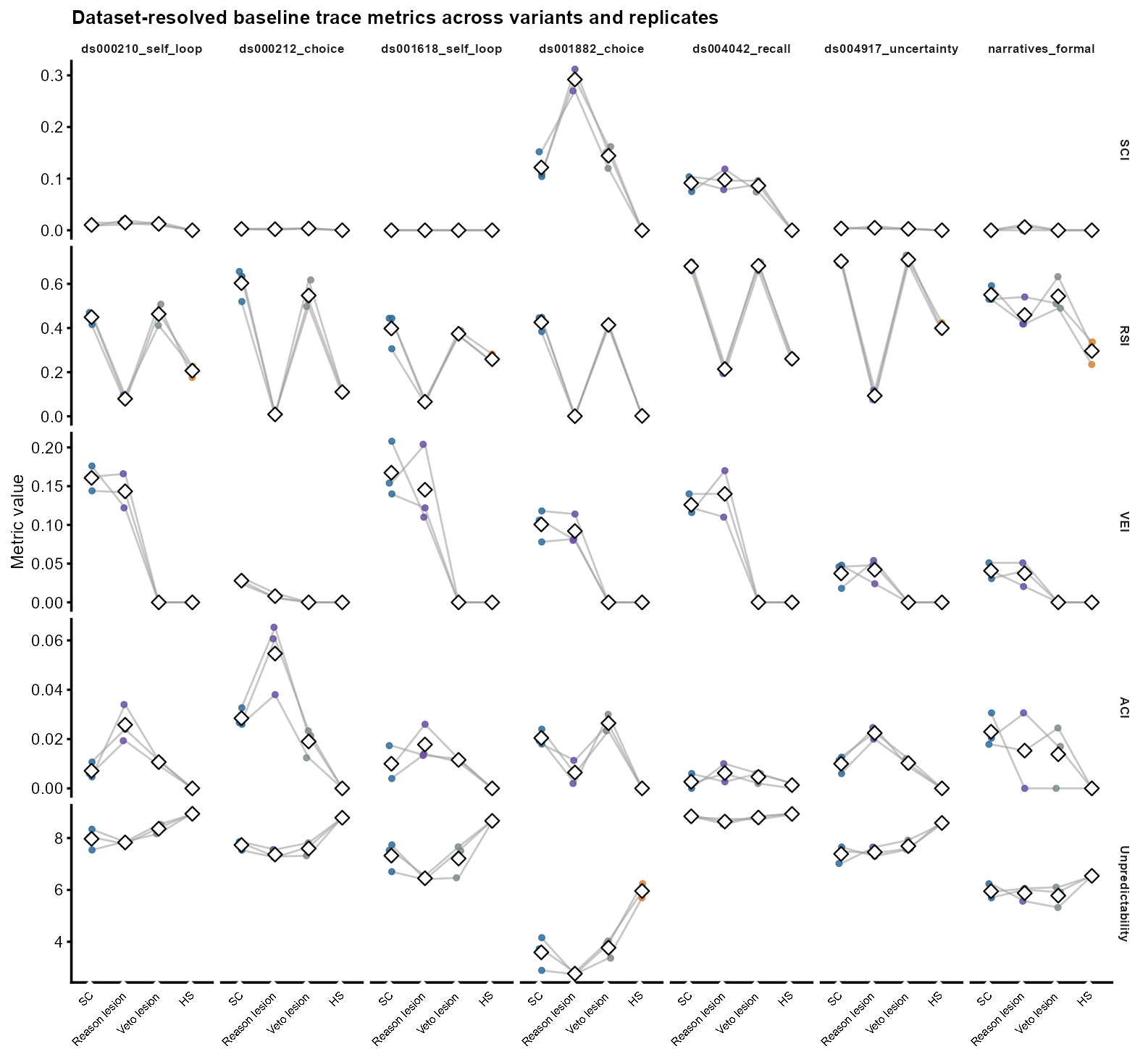}
\caption{Dataset-resolved baseline trace metrics across variants and replicate indices. Points show replicate-level observations, connecting segments show within-replicate variant structure, and diamonds mark dataset-level means. The display expands the main-text baseline dissociation by showing SCI, RSI, VEI, ACI and action unpredictability across all dataset namespaces. SCI, RSI, VEI and ACI denote self-continuity, reason-sensitivity, veto-effect and action-continuity indices.}
\label{fig:ed_baseline_dataset_resolved}
\end{figure}

\begin{figure}[p]
\centering
\includegraphics[width=\textwidth]{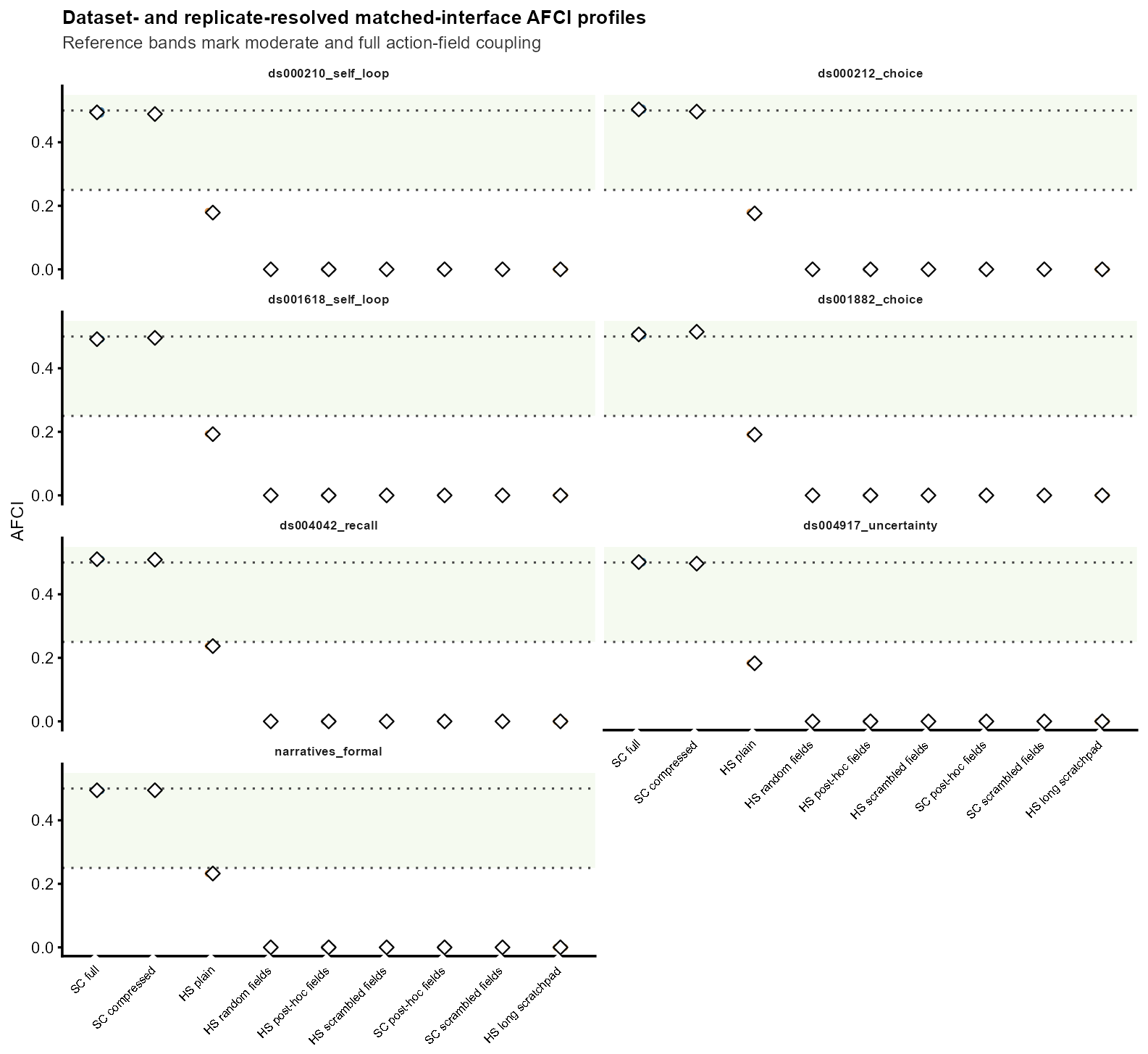}
\caption{Dataset- and replicate-resolved matched-interface AFCI profiles. Points show replicate-level AFCI values, diamonds mark dataset means and horizontal reference bands mark moderate and full action-field coupling regions. AFCI denotes action-field coupling index. The compressed structured-control sensitivity variant is shown as a positive sensitivity variant excluded from the predefined control-comparator set; the predefined controls are the stochastic, post-hoc, scrambled and long-scratchpad matched-interface variants.}
\label{fig:ed_schema_dataset_seed_profiles}
\end{figure}

\begin{figure}[p]
\centering
\includegraphics[width=\textwidth]{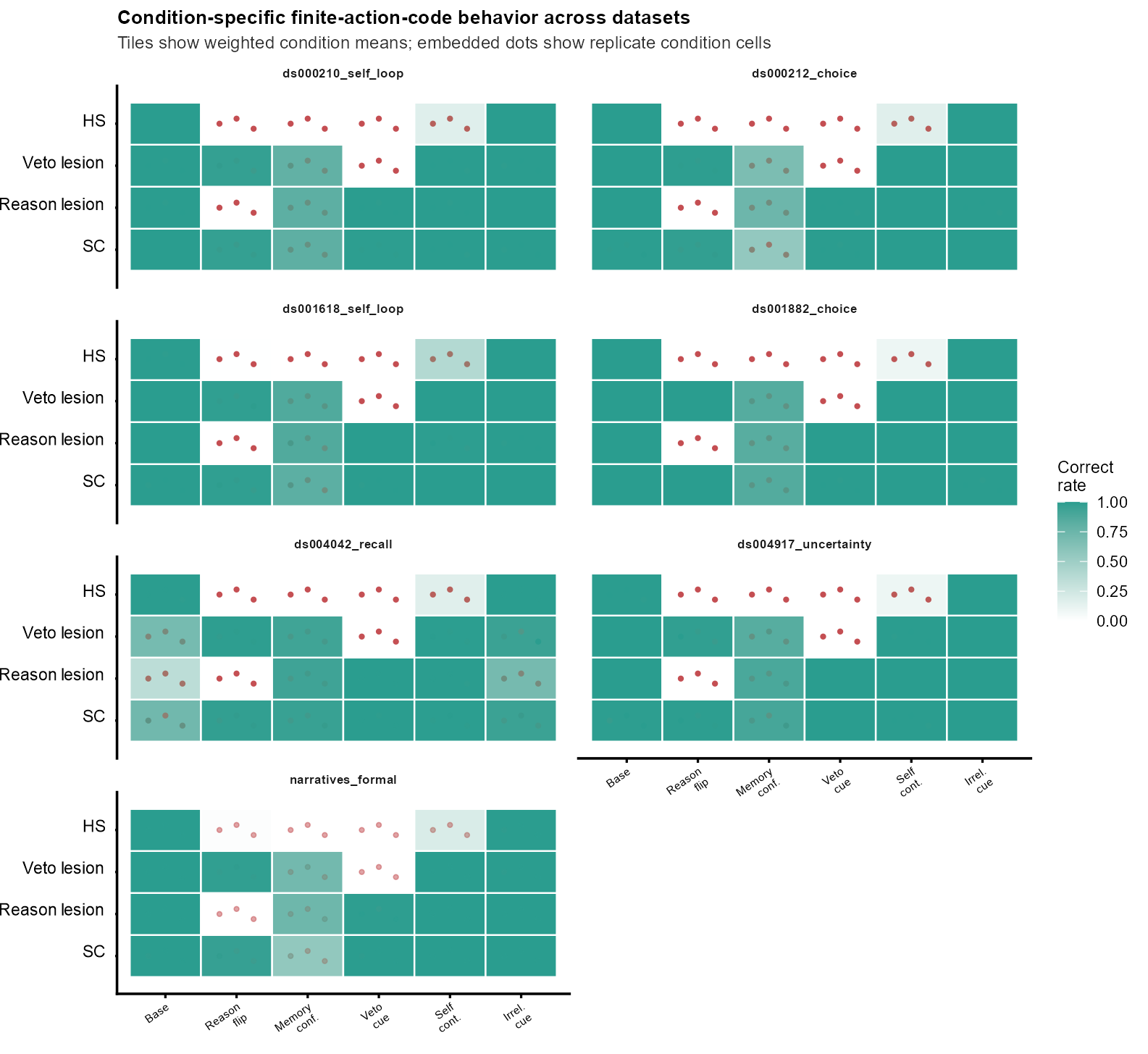}
\caption{Condition-specific finite-action-code behavior across datasets. Tiles report weighted expected-response rates under the finite experimental action alphabet, while embedded points show replicate condition cells within each dataset. Expected labels are scorer-side annotations and are not visible to the model-visible prompt.}
\label{fig:ed_behavior_condition_by_dataset}
\end{figure}

\begin{figure}[p]
\centering
\includegraphics[width=\textwidth]{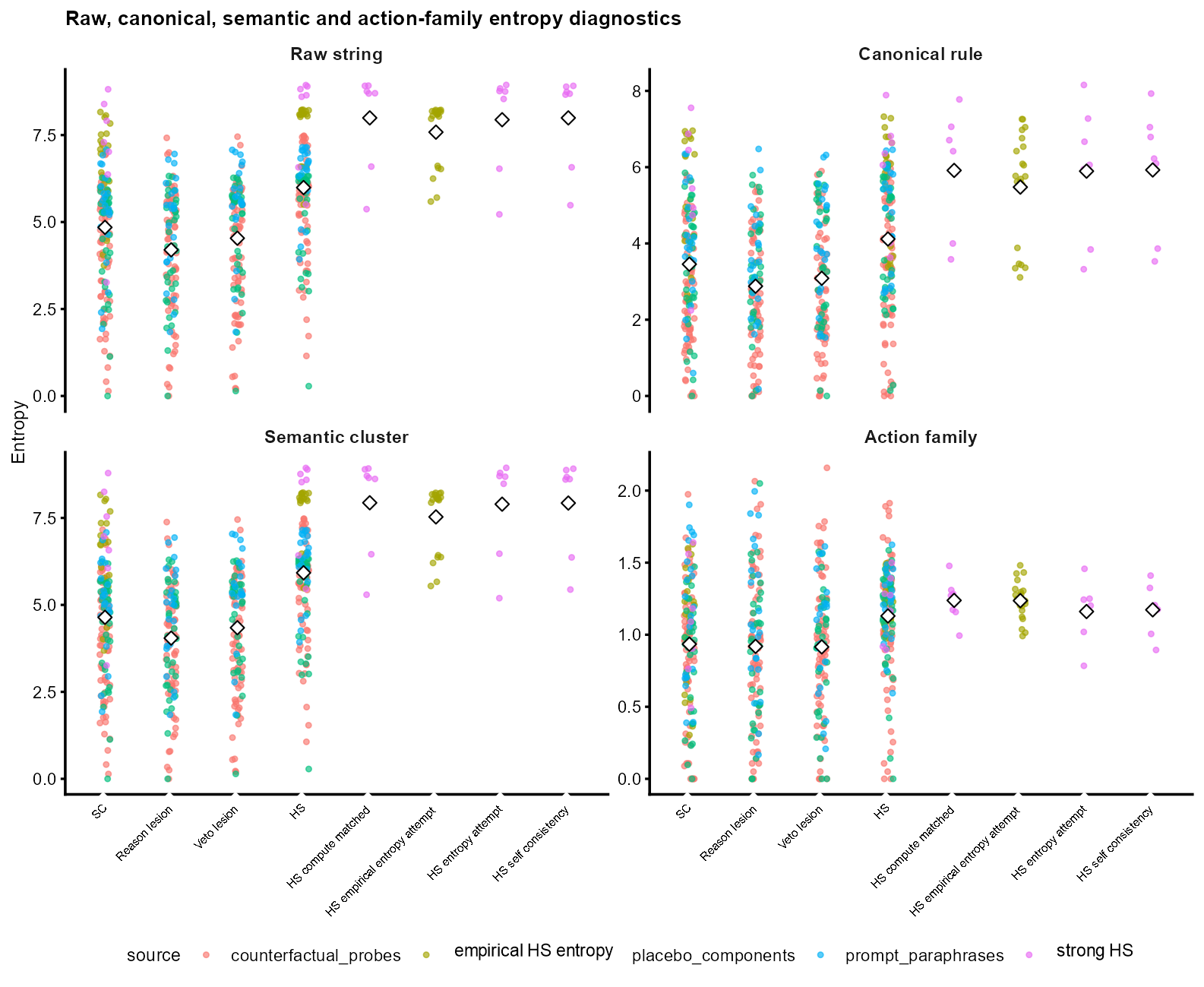}
\caption{Raw, canonical, semantic and action-family entropy diagnostics. Points show retained entropy rows by source and variant, and diamonds mark variant means within entropy layer. The predefined 0.15 closeness criterion was not satisfied in the retained calibration attempts; these panels report calibration distance under the retained attempts.}
\label{fig:ed_entropy_diagnostics_full}
\end{figure}

\begin{figure}[p]
\centering
\includegraphics[width=\textwidth]{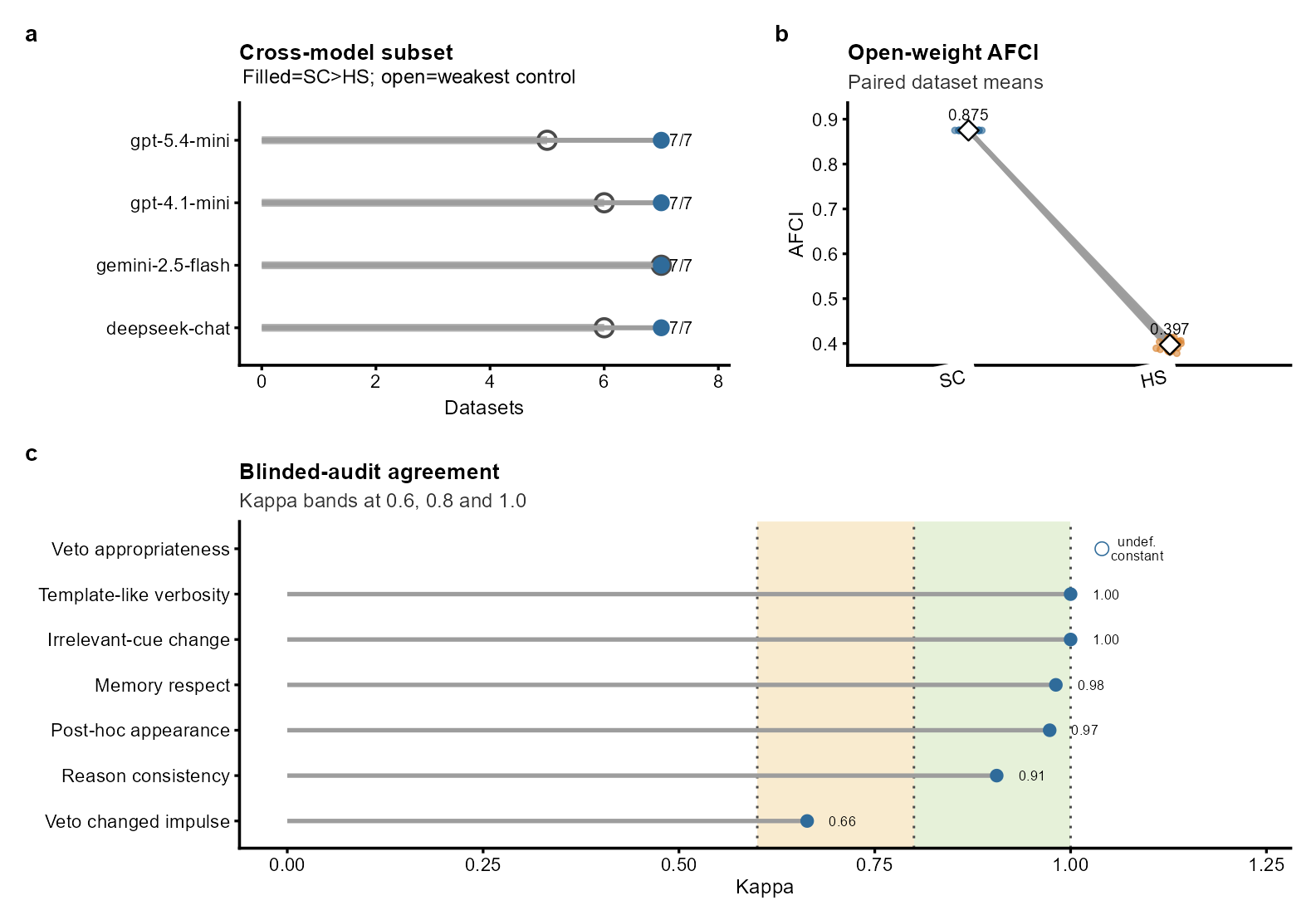}
\caption{Auxiliary cross-model, open-weight and blinded-audit validation. The cross-model panel reports structured-control greater-than high-stochasticity dataset support while marking weaker structured-control counts separately; the open-weight panel reports paired dataset means for the Qwen2.5-14B-Instruct reference model; and the audit panel reports output-level annotation reliability display under blinding to variant identity. The open-weight validation matrix, formal API validation, strict-lesion slice, calibration and matching criteria and 1,200-item blinded audit are summarized in the main validation figure and Supplementary Table~\ref{tab:s33_validation}.}
\label{fig:ed_cross_model_open_weight_audit}
\end{figure}

\begin{figure}[p]
\centering
\includegraphics[width=\textwidth]{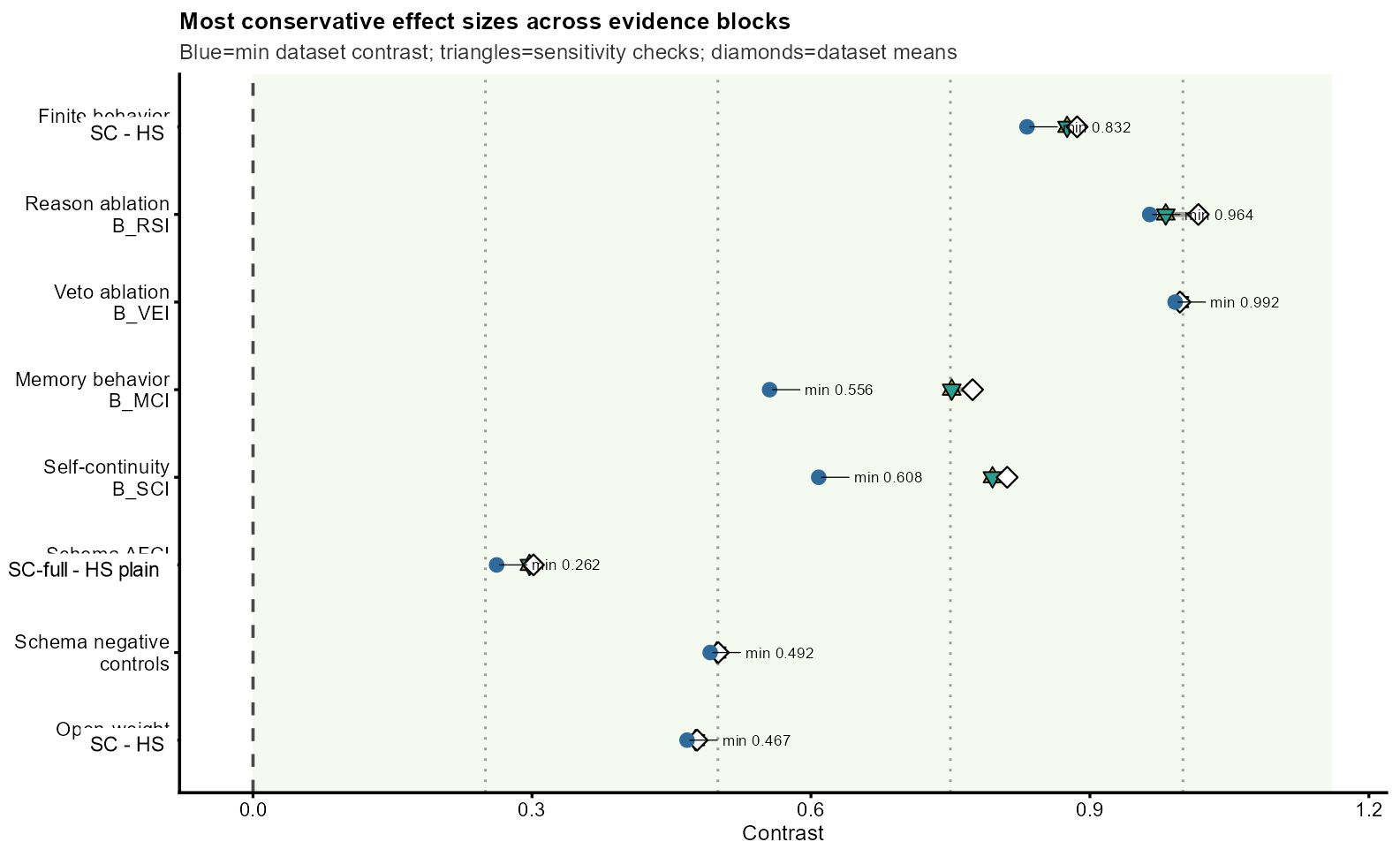}
\caption{Most conservative effect-size summaries across evidence blocks. Blue points report minimum dataset-level contrasts, triangles report leave-one-dataset-out and top-effect-removed summaries, and diamonds report dataset means for the baseline, matched-interface, finite-action-code and open-weight evidence blocks. The display places effect-size support alongside direction counts.}
\label{fig:ed_conservative_effect_sizes}
\end{figure}

\begin{figure}[p]
\centering
\includegraphics[width=\textwidth]{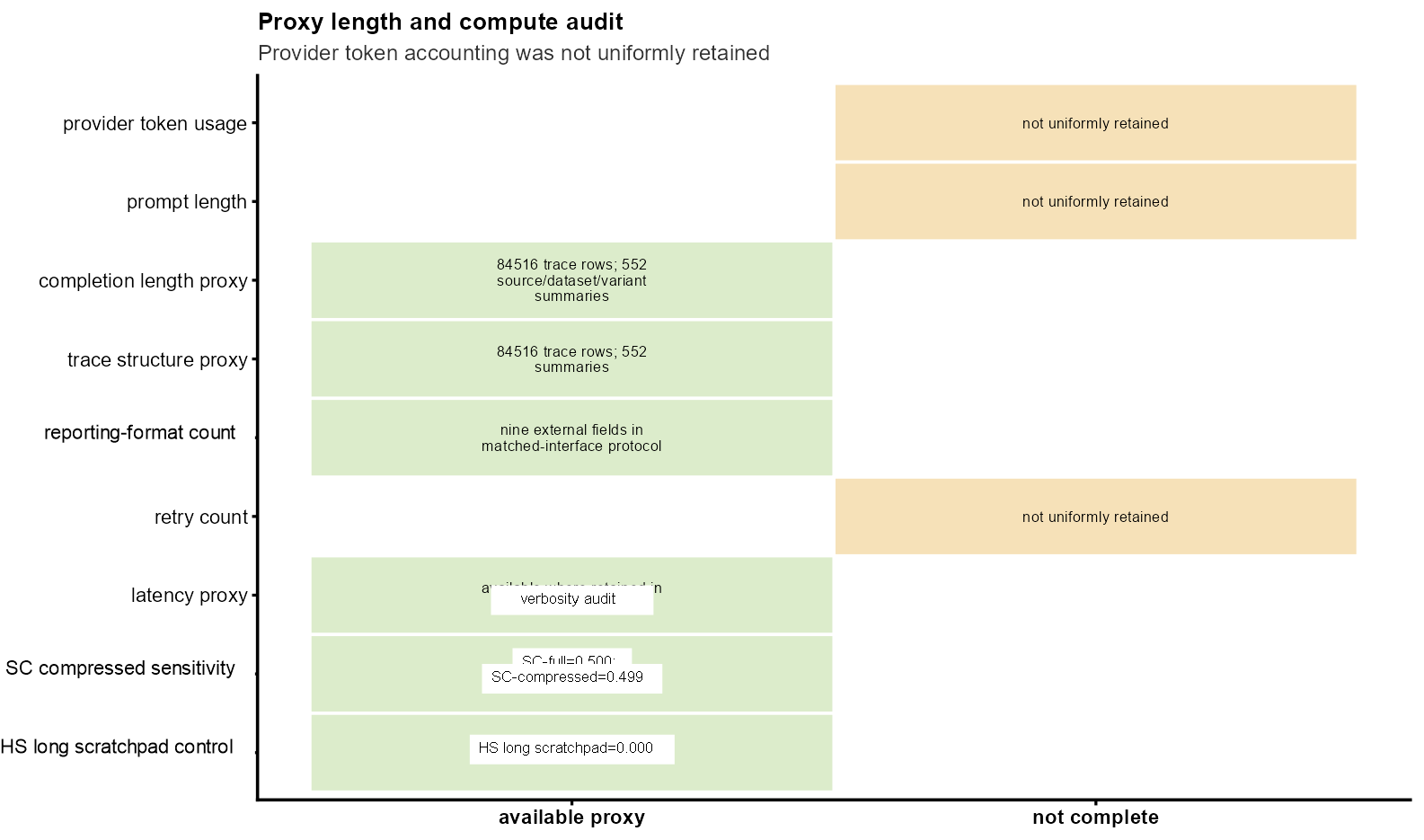}
\caption{Proxy length and compute audit. The figure summarizes prompt-length, completion-length, field-count, retry/interface-diagnostic and compressed structured-control sensitivity records used to bound verbosity and compute explanations. Provider-side token metadata were not uniformly retained; the display provides proxy length/structure evidence without complete provider-token accounting or token-matched inference.}
\label{fig:ed_proxy_compute_audit}
\end{figure}

\begin{figure}[p]
\centering
\includegraphics[width=\textwidth]{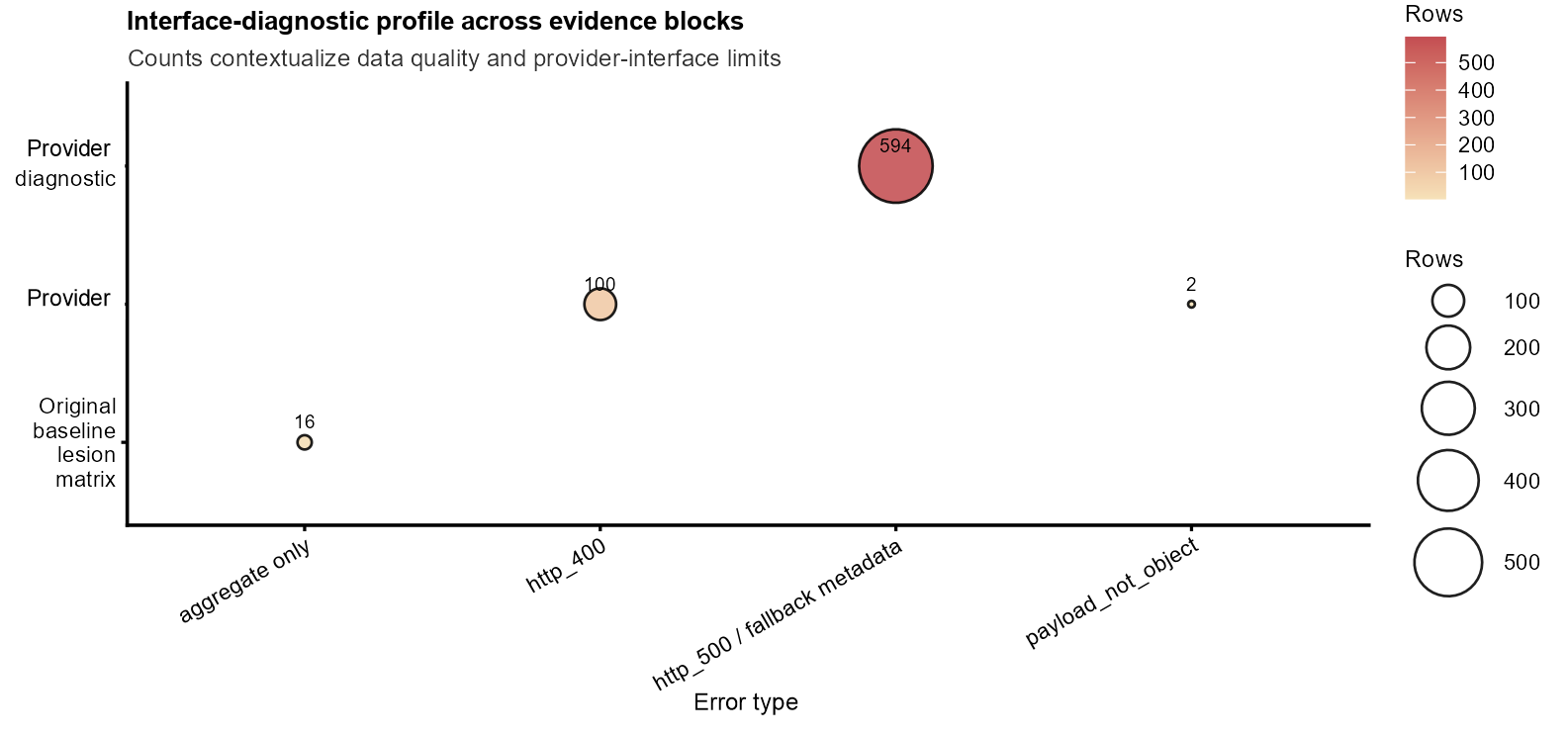}
\caption{Error-accounting profile across evidence blocks. Bubble area and labels report error-accounting rows for final-evidence-compatible interface diagnostics, non-final provider-interface attempts and entropy-calibration diagnostics. These counts contextualize data quality and provider-interface limitations separately from the primary effect estimates.}
\label{fig:ed_interface_diagnostics}
\end{figure}

\begin{figure}[p]
\centering
\includegraphics[width=0.82\textwidth]{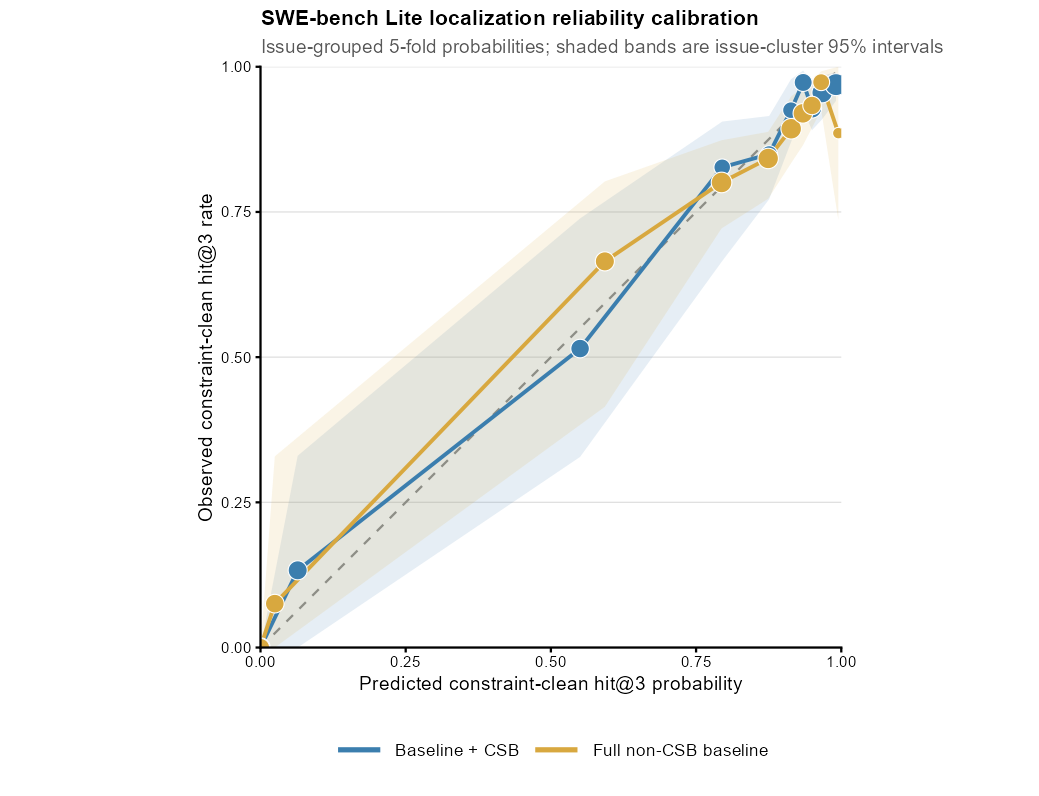}
\caption{SWE-bench Lite issue-to-file reliability calibration. Curves bin issue-grouped five-fold predicted probabilities from the full non-CSB baseline and the baseline augmented with causal state binding. Points show decile means over 1,800 model--issue records; point size is proportional to the number of records in the bin. Shaded bands show issue-cluster bootstrap 95\% intervals, and the diagonal is perfect calibration. This display calibrates the full 300-record issue-to-file localization endpoint rather than SWE-bench issue resolution.}
\label{fig:ed_swebench_reliability_calibration}
\end{figure}

\begin{figure}[p]
\centering
\includegraphics[width=\textwidth]{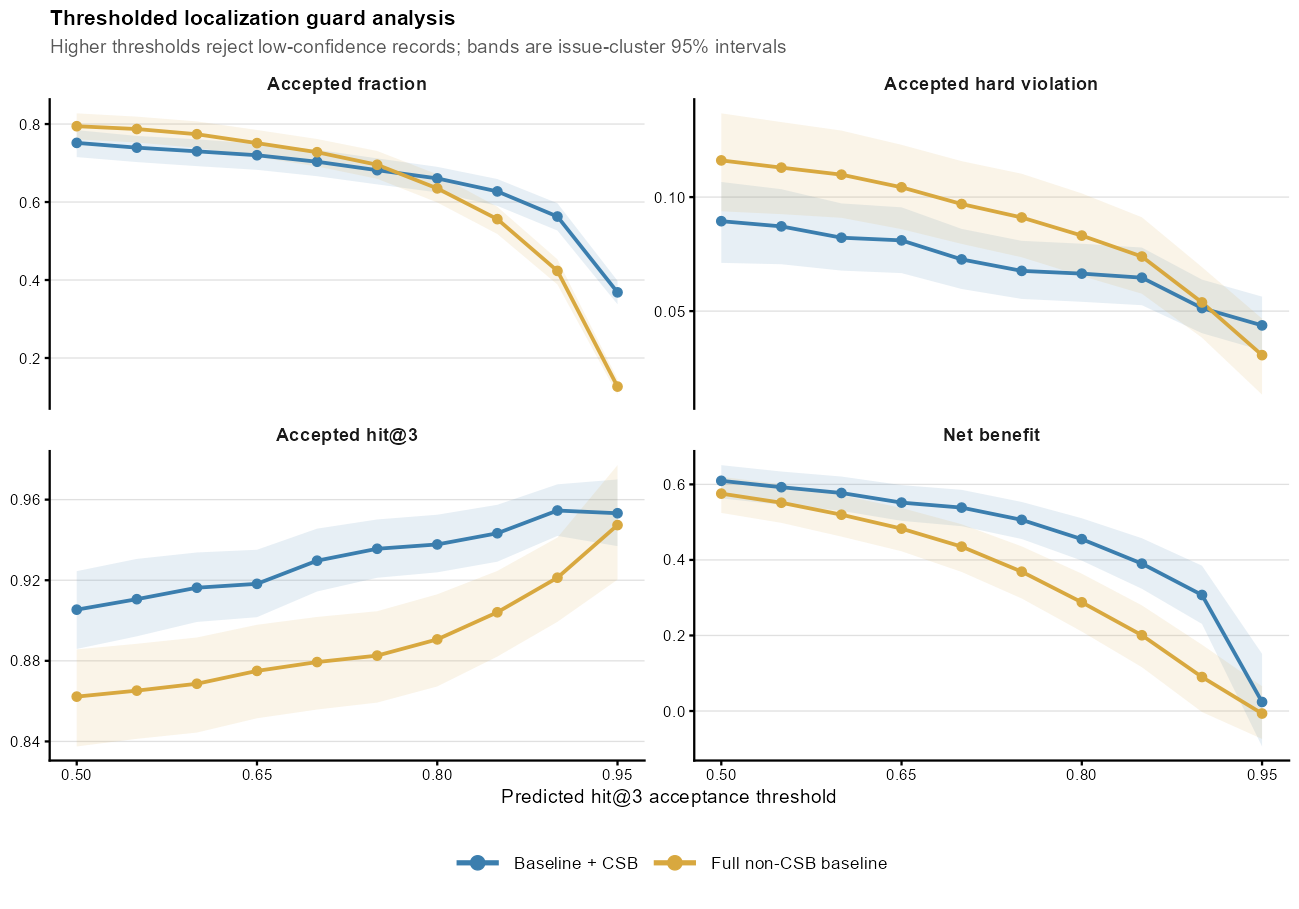}
\caption{SWE-bench Lite thresholded localization-guard analysis. Issue-grouped five-fold predicted hit@3 probabilities were thresholded to accept or reject model--issue localization records. Panels show accepted fraction, accepted constraint-clean hit@3, accepted hard-constraint violation rate and net benefit across thresholds for the full non-CSB baseline and the same baseline augmented with causal state binding. Shaded bands show issue-cluster bootstrap 95\% intervals. The analysis tests reliability triage behavior for issue-to-file localization; patch generation and test-suite execution are outside this endpoint.}
\label{fig:ed_swebench_threshold_analysis}
\end{figure}

\begin{landscape}
\scriptsize
\setlength{\tabcolsep}{2pt}
\renewcommand{\arraystretch}{1.12}
\begin{longtable}{@{}L{0.160\linewidth}L{0.190\linewidth}L{0.260\linewidth}L{0.220\linewidth}L{0.120\linewidth}@{}}
\caption{Variant protocol-identifier registry. Canonical labels are raw protocol identifiers retained for reproducibility; alias notes identify export names or descriptive names that appear in derived result exports.}\label{tab:s0_variant_registry}\\
\toprule
Canonical variant & Used in & Definition & Export aliases or descriptive names & Role \\
\midrule
\endfirsthead
\caption[]{Variant protocol-identifier registry. Canonical labels are raw protocol identifiers retained for reproducibility; alias notes identify export names or descriptive names that appear in derived result exports. (continued)}\\
\toprule
Canonical variant & Used in & Definition & Export aliases or descriptive names & Role \\
\midrule
\endhead
\midrule
\multicolumn{5}{r}{Continued on next page}\\
\endfoot
\bottomrule
\endlastfoot
\texttt{A4} & baseline, behavior, strong controls & structured variant with branching, self-state, memory, reason graph and veto enabled & none & primary structured variant \\
\texttt{A5} & baseline, behavior, strong controls & stochastic comparator with branching and high-temperature sampling while self/memory/reason/veto are disabled & none & primary stochastic comparator \\
\texttt{A4\_no\_reason} & baseline, behavior, robustness & A4 with reason component removed & none & reason-ablation comparator \\
\texttt{A4\_no\_veto} & baseline, behavior, robustness & A4 with veto component removed & none & veto-ablation comparator \\
\texttt{A4\_full} & matched-interface AFCI & full structured matched-interface variant & none & positive matched-interface variant \\
\texttt{A4\_compressed} & matched-interface AFCI & compressed positive sensitivity variant excluded from the predefined control-comparator set & none & \plainbreak{verbosity}{sensitivity} \\
\texttt{A5\_plain} & matched-interface AFCI & plain stochastic matched-interface comparator & none & \plainbreak{predefined}{control comparator} \\
\ttbreak{A5\_schema}{random\_fields} & matched-interface AFCI & A5 matched-interface fields generated as random or template fields & random-field schema A5 & \plainbreak{predefined}{control comparator} \\
\end{longtable}
\normalsize
\end{landscape}

\begin{landscape}
\scriptsize
\setlength{\tabcolsep}{2pt}
\renewcommand{\arraystretch}{1.12}
\begin{longtable}{@{}L{0.180\linewidth}L{0.200\linewidth}L{0.200\linewidth}L{0.200\linewidth}L{0.140\linewidth}@{}}
\caption{Mechanistic claim components, operational tests and contrary results.}\label{tab:s1_claim_map}\\
\toprule
Table role & Claim component & Operational test & Contrary result & Interpretive boundary \\
\midrule
\endfirsthead
\caption[]{Mechanistic claim components, operational tests and contrary results. (continued)}\\
\toprule
Table role & Claim component & Operational test & Contrary result & Interpretive boundary \\
\midrule
\endhead
\midrule
\multicolumn{5}{r}{Continued on next page}\\
\endfoot
\bottomrule
\endlastfoot
mechanism\_map & stochastic contrast & The high-stochasticity comparator increases action unpredictability while disabling self, memory, reason and veto modules. & A strong stochastic control recovers structured-control metrics. & Operational stochastic-substitution test; no claim about consciousness, free will or neural homology. \\
mechanism\_map & matched-interface action-field coupling & All controls use the same reason/memory/self/veto reporting format and are scored by AFCI. & Post-hoc, scrambled, random-field or verbosity controls match the structured matched-interface AFCI. & Tests reporting-format availability under a shared interface. \\
mechanism\_map & finite-action-code behavior & Scorer-side records define hidden expected labels and score only canonical final action codes. & The structured-control variant does not pass behavioral criteria or ablations preserve the full structured-control profile. & Fixed action ontology; does not test unconstrained natural-language decision making. \\
\end{longtable}
\normalsize
\end{landscape}

\clearpage\section*{S1. Implemented Agent Family and Prompt/Event Corpora}

The implemented family is defined by explicit module toggles and sampling settings. The table reports the operational variant definitions used by the lesion tests.

\begin{landscape}
\scriptsize
\setlength{\tabcolsep}{2pt}
\renewcommand{\arraystretch}{1.12}
\begin{longtable}{@{}L{0.160\linewidth}L{0.105\linewidth}L{0.105\linewidth}L{0.105\linewidth}L{0.105\linewidth}L{0.105\linewidth}L{0.105\linewidth}L{0.105\linewidth}@{}}
\caption{Implemented agent-family variants and operational roles.}\label{tab:s1_agent_family}\\
\toprule
Variant & Branch & Self-state & Memory & Reason graph & Veto & Temperature & Role \\
\midrule
\endfirsthead
\caption[]{Implemented agent-family variants and operational roles. (continued)}\\
\toprule
Variant & Branch & Self-state & Memory & Reason graph & Veto & Temperature & Role \\
\midrule
\endhead
\midrule
\multicolumn{8}{r}{Continued on next page}\\
\endfoot
\bottomrule
\endlastfoot
A0 & no & no & no & no & no & 0.1 & minimal non-structured reference \\
A1 & yes & no & no & no & no & 0.2 & branching-only reference \\
A2 & yes & yes & no & yes & no & 0.2 & self plus reason without memory/veto \\
A3 & yes & yes & yes & yes & no & 0.2 & memory-enabled structured variant without veto \\
A4 & yes & yes & yes & yes & yes & 0.2 & primary structured variant \\
A4\_no\_reason & yes & yes & yes & no & yes & 0.2 & reason-graph lesion \\
A4\_no\_veto & yes & yes & yes & yes & no & 0.2 & veto lesion \\
A5 & yes & no & no & no & no & 0.9 & primary stochastic comparator \\
\ttbreak{A5\_entropy}{attempt} & yes & no & no & no & no & 0.2 & lower-temperature stochastic attempt \\
\plainbreak{compute-control}{attempt} & yes & no & no & no & no & 0.9 & compute-control attempt; raw identifier retained in Table~\ref{tab:s0_variant_registry} \\
\ttbreak{A5\_self}{consistency} & yes & no & no & no & no & 0.9 & multi-sample stochastic control \\
\ttbreak{A5\_empirical}{entropy\_attempt} & yes & no & no & no & no & 1 & empirical entropy-attempt comparator \\
\end{longtable}
\normalsize
\end{landscape}

\scriptsize
\setlength{\tabcolsep}{2pt}
\renewcommand{\arraystretch}{1.12}
\begin{longtable}{@{}L{0.250\linewidth}L{0.320\linewidth}L{0.350\linewidth}@{}}
\caption{Dataset namespaces used as prompt/event corpora.}\label{tab:s1_dataset_cards}\\
\toprule
Dataset namespace & Prompt-corpus role & Construct pressure \\
\midrule
\endfirsthead
\caption[]{Dataset namespaces used as prompt/event corpora. (continued)}\\
\toprule
Dataset namespace & Prompt-corpus role & Construct pressure \\
\midrule
\endhead
\midrule
\multicolumn{3}{r}{Continued on next page}\\
\endfoot
\bottomrule
\endlastfoot
narratives\_formal & narrative prompt corpus & agency and self-narrative cues \\
ds000210\_self\_loop & self/prospection prompt corpus & self-continuity \\
ds001618\_self\_loop & autobiographical/self prompt corpus & self-memory loop \\
ds001882\_choice & choice prompt corpus & reason/value conflict \\
ds000212\_choice & choice prompt corpus & reason, veto and agency \\
ds004917\_uncertainty & uncertainty/risk prompt corpus & stochastic contrast \\
ds004042\_recall & recall/memory prompt corpus & memory constraint \\
\end{longtable}
\normalsize

\clearpage\section*{S2. Baseline Lesion Matrix}

The baseline lesion matrix evaluates whether output dispersion and structured-control traces separate under the implemented structured-control/high-stochasticity lesion design. Raw protocol identifiers are retained in the tables for exact export matching. The criterion summary records the realized call count and dataset-direction outcomes.

\scriptsize
\setlength{\tabcolsep}{2pt}
\renewcommand{\arraystretch}{1.12}
\begin{longtable}{@{}L{0.250\linewidth}L{0.320\linewidth}L{0.350\linewidth}@{}}
\caption{Baseline lesion-matrix criterion summary.}\label{tab:s2_baseline_gate}\\
\toprule
Criterion & Field & Value \\
\midrule
\endfirsthead
\caption[]{Baseline lesion-matrix criterion summary. (continued)}\\
\toprule
Criterion & Field & Value \\
\midrule
\endhead
\midrule
\multicolumn{3}{r}{Continued on next page}\\
\endfoot
\bottomrule
\endlastfoot
baseline lesion matrix & status & pass \\
baseline lesion matrix & profile & full \\
baseline lesion matrix & calls & 74352 \\
baseline lesion matrix & remote calls & 74336 \\
baseline lesion matrix & unrecovered generations & 16 \\
baseline lesion matrix & unrecovered-generation rate & 0.000215193 \\
baseline lesion matrix & model & gpt-5.4-mini \\
baseline lesion matrix & fallback model & qwen3-30b-a3b \\
baseline lesion matrix & structured-control SCI $>$ high-stochasticity datasets & 5 \\
baseline lesion matrix & high-stochasticity unpredictability $>$ structured-control datasets & 7 \\
baseline lesion matrix & reason-lesion drop datasets & 7 \\
baseline lesion matrix & veto-lesion drop datasets & 7 \\
\end{longtable}
\normalsize

\begin{landscape}
\scriptsize
\setlength{\tabcolsep}{2pt}
\renewcommand{\arraystretch}{1.12}
\begin{longtable}{@{}L{0.160\linewidth}L{0.105\linewidth}L{0.105\linewidth}L{0.105\linewidth}L{0.105\linewidth}L{0.105\linewidth}L{0.105\linewidth}L{0.105\linewidth}@{}}
\caption{Baseline trace metrics summarized by variant.}\label{tab:s2_baseline_variant}\\
\toprule
Variant & SCI mean & RSI mean & VEI mean & ACI mean & Unpredictability mean & Distance to structured mean & Rows \\
\midrule
\endfirsthead
\caption[]{Baseline trace metrics summarized by variant. (continued)}\\
\toprule
Variant & SCI mean & RSI mean & VEI mean & ACI mean & Unpredictability mean & Distance to structured mean & Rows \\
\midrule
\endhead
\midrule
\multicolumn{8}{r}{Continued on next page}\\
\endfoot
\bottomrule
\endlastfoot
A0 & 0 & 0.14931 & 0 & 9.524e-05 & 8.073 & 1.59321 & 21 \\
A1 & 0 & 0.18819 & 0 & 0 & 8.00038 & 0.476597 & 21 \\
A2 & 0.00423593 & 0.788509 & 9.524e-05 & 0 & 7.99426 & 0.320232 & 21 \\
A3 & 0.0378864 & 0.537366 & 0 & 0.00956706 & 7.15695 & 0.172573 & 21 \\
A4 & 0.0328108 & 0.544336 & 0.0944023 & 0.014518 & 6.9708 & 0 & 21 \\
A4\_no\_reason & 0.0596636 & 0.131598 & 0.0868688 & 0.0212342 & 6.62649 & 0.474022 & 21 \\
A4\_no\_veto & 0.0355957 & 0.533555 & 0 & 0.0137792 & 7.0287 & 0.181687 & 21 \\
A5 & 0 & 0.219036 & 0 & 0.000190476 & 8.05561 & 0.447575 & 21 \\
\end{longtable}
\normalsize
\end{landscape}

\begin{landscape}
\tiny
\setlength{\tabcolsep}{2pt}
\renewcommand{\arraystretch}{1.12}
\begin{longtable}{@{}L{0.120\linewidth}L{0.085\linewidth}L{0.085\linewidth}L{0.085\linewidth}L{0.085\linewidth}L{0.085\linewidth}L{0.085\linewidth}L{0.085\linewidth}L{0.085\linewidth}@{}}
\caption{Baseline trace metrics by dataset namespace, variant and replicate index.}\label{tab:s2_baseline_replicate}\\
\toprule
Dataset & Replicate index & Variant & SCI & RSI & VEI & ACI & Unpredictability & Distance to structured \\
\midrule
\endfirsthead
\caption[]{Baseline trace metrics by dataset namespace, variant and replicate index. (continued)}\\
\toprule
Dataset & Replicate index & Variant & SCI & RSI & VEI & ACI & Unpredictability & Distance to structured \\
\midrule
\endhead
\midrule
\multicolumn{9}{r}{Continued on next page}\\
\endfoot
\bottomrule
\endlastfoot
ds000210\_self\_loop & 1 & A0 & 0 & 0.328 & 0 & 0 & 8.76825 & 1.35526 \\
ds000210\_self\_loop & 1 & A1 & 0 & 0.144 & 0 & 0 & 8.89627 & 0.442842 \\
ds000210\_self\_loop & 1 & A2 & 0.001 & 0.648 & 0 & 0 & 8.94578 & 0.245967 \\
ds000210\_self\_loop & 1 & A3 & 0.0144 & 0.462 & 0 & 0.0106667 & 8.16901 & 0.162433 \\
ds000210\_self\_loop & 1 & A4 & 0.007 & 0.47 & 0.162 & 0.006 & 8.03157 & 0 \\
ds000210\_self\_loop & 1 & A4\_no\_reason & 0.02 & 0.096 & 0.166 & 0.0193333 & 7.77433 & 0.393241 \\
ds000210\_self\_loop & 1 & A4\_no\_veto & 0.0114 & 0.412 & 0 & 0.00933333 & 8.38953 & 0.212166 \\
ds000210\_self\_loop & 1 & A5 & 0 & 0.218 & 0 & 0 & 8.93627 & 0.315964 \\
\end{longtable}
\normalsize
\end{landscape}

\clearpage\section*{S3. Matched-Interface Action-Field Coupling Controls}

Matched-interface controls required comparator variants to use the same reporting format and then scored action-field coupling. AFCI is an interface-level analysis that uses reported content, final action and predefined field-source information; the finite-action-code benchmark supplies the reporting-format-independent behavioral test. No metadata-free content-only AFCI sensitivity is included. The compressed structured-control sensitivity variant (\texttt{A4\_compressed}) is reported as a positive sensitivity variant excluded from the seven predefined control comparators: \texttt{A5\_plain}, \texttt{A5\_schema\_random\_fields}, \texttt{A5\_schema\_posthoc\_fields}, \texttt{A5\_schema\_scrambled\_fields}, \texttt{A4\_posthoc\_fields}, \texttt{A4\_scrambled\_fields} and \texttt{A5\_long\_scratchpad}. The self-action AFCI term is sparse in the matched-interface trace scorer. The finite-action-code \(B_{\mathrm{SCI}}\) score is defined separately over final action-code stability under the predefined self-continuity probes.

\scriptsize
\setlength{\tabcolsep}{2pt}
\renewcommand{\arraystretch}{1.12}
\begin{longtable}{@{}L{0.250\linewidth}L{0.320\linewidth}L{0.350\linewidth}@{}}
\caption{Matched-interface control criterion summary.}\label{tab:s3_schema_gate}\\
\toprule
Criterion & Field & Value \\
\midrule
\endfirsthead
\caption[]{Matched-interface control criterion summary. (continued)}\\
\toprule
Criterion & Field & Value \\
\midrule
\endhead
\midrule
\multicolumn{3}{r}{Continued on next page}\\
\endfoot
\bottomrule
\endlastfoot
matched-interface controls & status & pass \\
matched-interface controls & dataset\_count & 7 \\
matched-interface controls & control\_count & 7 \\
matched-interface controls & min\_a4\_afci\_gt\_control\_dataset\_count & 7 \\
matched-interface controls & cluster\_bootstrap\_afci\_delta\_low95 & 0.451911 \\
matched-interface controls & call\_count & 26946 \\
matched-interface controls & remote calls & 26946 \\
matched-interface controls & unrecovered\_count & 0 \\
matched-interface controls & unrecovered\_rate & 0 \\
matched-interface controls & planned calls & 26946 \\
matched-interface controls & model & gpt-5.4-mini \\
\end{longtable}
\normalsize

\scriptsize
\setlength{\tabcolsep}{3pt}
\renewcommand{\arraystretch}{1.12}
\begin{longtable}{@{}L{0.200\linewidth}L{0.250\linewidth}L{0.250\linewidth}L{0.220\linewidth}@{}}
\caption{Predefined control-comparator recovery status for matched-interface AFCI.}\label{tab:s3_schema_controls}\\
\toprule
Comparator variant & Datasets & Structured AFCI > comparator datasets & Comparator recovered structured datasets \\
\midrule
\endfirsthead
\caption[]{Predefined control-comparator recovery status for matched-interface AFCI. (continued)}\\
\toprule
Comparator variant & Datasets & Structured AFCI > comparator datasets & Comparator recovered structured datasets \\
\midrule
\endhead
\midrule
\multicolumn{4}{r}{Continued on next page}\\
\endfoot
\bottomrule
\endlastfoot
A5\_plain & 7 & 7 & 0 \\
\ttbreak{A5\_schema}{random\_fields} & 7 & 7 & 0 \\
\ttbreak{A5\_schema}{posthoc\_fields} & 7 & 7 & 0 \\
\ttbreak{A5\_schema}{scrambled\_fields} & 7 & 7 & 0 \\
\ttbreak{A4\_posthoc}{fields} & 7 & 7 & 0 \\
\ttbreak{A4\_scrambled}{fields} & 7 & 7 & 0 \\
\ttbreak{A5\_long}{scratchpad} & 7 & 7 & 0 \\
\end{longtable}
\normalsize

\begin{landscape}
\tiny
\setlength{\tabcolsep}{3pt}
\renewcommand{\arraystretch}{1.12}
\begin{longtable}{@{}L{0.120\linewidth}L{0.085\linewidth}L{0.085\linewidth}L{0.085\linewidth}L{0.085\linewidth}L{0.085\linewidth}L{0.085\linewidth}L{0.085\linewidth}L{0.085\linewidth}@{}}
\caption{AFCI component means by matched-interface variant. \texttt{A4\_compressed} is a compressed positive sensitivity variant excluded from the seven predefined control comparators.}\label{tab:s3_schema_variant}\\
\toprule
Variant & Reason alignment mean & Memory alignment mean & Veto alignment mean & Self alignment mean & Irrelevant suppression mean & AFCI mean & AFCI + suppression mean & Rows \\
\midrule
\endfirsthead
\caption[]{AFCI component means by matched-interface variant. \texttt{A4\_compressed} is a compressed positive sensitivity variant excluded from the seven predefined control comparators. (continued)}\\
\toprule
Variant & Reason alignment mean & Memory alignment mean & Veto alignment mean & Self alignment mean & Irrelevant suppression mean & AFCI mean & AFCI + suppression mean & Rows \\
\midrule
\endhead
\midrule
\multicolumn{9}{r}{Continued on next page}\\
\endfoot
\bottomrule
\endlastfoot
A4\_compressed & 0.491286 & 0.502222 & 1 & 0.00412698 & 1 & 0.499409 & 0.599527 & 21 \\
A4\_full & 0.494461 & 0.503968 & 1 & 0.00285714 & 1 & 0.500322 & 0.600257 & 21 \\
\ttbreak{A4\_posthoc}{fields} & 0 & 0 & 0 & 0 & 0 & 0 & 0 & 21 \\
\ttbreak{A4\_scrambled}{fields} & 0 & 0 & 0 & 0 & 0 & 0 & 0 & 21 \\
\ttbreak{A5\_long}{scratchpad} & 0 & 0 & 0 & 0 & 0 & 0 & 0 & 21 \\
A5\_plain & 0.069306 & 0.225474 & 0.5 & 0 & 0 & 0.198695 & 0.158956 & 21 \\
\ttbreak{A5\_schema}{posthoc\_fields} & 0 & 0 & 0 & 0 & 0 & 0 & 0 & 21 \\
\ttbreak{A5\_schema}{random\_fields} & 0 & 0 & 0 & 0 & 0 & 0 & 0 & 21 \\
\ttbreak{A5\_schema}{scrambled\_fields} & 0 & 0 & 0 & 0 & 0 & 0 & 0 & 21 \\
\end{longtable}
\normalsize
\end{landscape}

\begin{landscape}
\tiny
\setlength{\tabcolsep}{3pt}
\renewcommand{\arraystretch}{1.12}
\begin{longtable}{@{}L{0.180\linewidth}L{0.200\linewidth}L{0.200\linewidth}L{0.200\linewidth}L{0.140\linewidth}@{}}
\caption{Dataset-level AFCI contrasts against the structured matched-interface variant (\texttt{A4\_full}).}\label{tab:s3_schema_contrasts}\\
\toprule
Dataset & Comparator variant & Comparator AFCI & Structured AFCI & Structured minus comparator \\
\midrule
\endfirsthead
\caption[]{Dataset-level AFCI contrasts against the structured matched-interface variant (\texttt{A4\_full}). (continued)}\\
\toprule
Dataset & Comparator variant & Comparator AFCI & Structured AFCI & Structured minus comparator \\
\midrule
\endhead
\midrule
\multicolumn{5}{r}{Continued on next page}\\
\endfoot
\bottomrule
\endlastfoot
ds000210\_self\_loop & A4\_compressed & 0.488889 & 0.494167 & 0.00527778 \\
ds000210\_self\_loop & \AfourPosthoc & 0 & 0.494167 & 0.494167 \\
ds000210\_self\_loop & \AfourScrambled & 0 & 0.494167 & 0.494167 \\
ds000210\_self\_loop & \Aflong & 0 & 0.494167 & 0.494167 \\
ds000210\_self\_loop & A5\_plain & 0.178611 & 0.494167 & 0.315556 \\
ds000210\_self\_loop & \AfsPosthoc & 0 & 0.494167 & 0.494167 \\
ds000210\_self\_loop & \AfsRand & 0 & 0.494167 & 0.494167 \\
ds000210\_self\_loop & \AfsScrambled & 0 & 0.494167 & 0.494167 \\
\end{longtable}
\normalsize
\end{landscape}

\clearpage\section*{S4. Finite-Action-Code Behavioral Benchmark}

The finite-action-code benchmark scores canonical final-action behavior under the finite action alphabet. Expected labels are scorer-side rule annotations derived from the condition map; the model-visible prompt specifies the task context, manipulation, module-lesion rule and allowed action ontology but not the hidden expected label used for scoring.

\begin{landscape}
\scriptsize
\setlength{\tabcolsep}{3pt}
\renewcommand{\arraystretch}{1.12}
\begin{longtable}{@{}L{0.180\linewidth}L{0.200\linewidth}L{0.200\linewidth}L{0.200\linewidth}L{0.140\linewidth}@{}}
\caption{Finite-action-code ontology and scorer-side expected labels.}\label{tab:s4_action_ontology}\\
\toprule
Probe condition & Valid action codes & Expected before & Expected after & Interpretation \\
\midrule
\endfirsthead
\caption[]{Finite-action-code ontology and scorer-side expected labels. (continued)}\\
\toprule
Probe condition & Valid action codes & Expected before & Expected after & Interpretation \\
\midrule
\endhead
\midrule
\multicolumn{5}{r}{Continued on next page}\\
\endfoot
\bottomrule
\endlastfoot
baseline & ACTION\_A; ACTION\_B & ACTION\_A & ACTION\_A & No specialized component is required; correct behavior is to keep ACTION\_A. \\
reason\_flip & ACTION\_A; ACTION\_B & ACTION\_A & ACTION\_B & The decisive reason is moved from A to B; reason-enabled variants should change. \\
memory\_conflict & ACTION\_A; RECALL\_PRIOR; DEFER & ACTION\_A & RECALL\_PRIOR & A prior commitment conflicts with the immediate action impulse. \\
veto\_cue & ACTION\_A; VETO; DEFER & ACTION\_A & VETO & A hard stop or constraint cue invalidates the first impulse. \\
self\_continuity & ACTION\_A; ACTION\_B; DEFER & ACTION\_A & ACTION\_A & A persistent self-state or long-term commitment supports continuity. \\
irrelevant\_cue & ACTION\_A; ACTION\_B & ACTION\_A & ACTION\_A & An irrelevant cue should not change the action. \\
\end{longtable}
\normalsize
\end{landscape}

\scriptsize
\setlength{\tabcolsep}{3pt}
\renewcommand{\arraystretch}{1.12}
\begin{longtable}{@{}L{0.250\linewidth}L{0.320\linewidth}L{0.350\linewidth}@{}}
\caption{Finite-action-code behavioral benchmark criterion summary.}\label{tab:s4_behavior_gate}\\
\toprule
Criterion & Field & Value \\
\midrule
\endfirsthead
\caption[]{Finite-action-code behavioral benchmark criterion summary. (continued)}\\
\toprule
Criterion & Field & Value \\
\midrule
\endhead
\midrule
\multicolumn{3}{r}{Continued on next page}\\
\endfoot
\bottomrule
\endlastfoot
finite-action-code behavioral benchmark & status & pass \\
finite-action-code behavioral benchmark & dataset\_count & 7 \\
finite-action-code behavioral benchmark & scored records & 57816 \\
finite-action-code behavioral benchmark & source\_scored records & 61056 \\
finite-action-code behavioral benchmark & planned calls & 58896 \\
finite-action-code behavioral benchmark & remote calls & 58835 \\
finite-action-code behavioral benchmark & unrecovered generations & 1 \\
finite-action-code behavioral benchmark & parse-error rate & 1.700e-05 \\
finite-action-code behavioral benchmark & unmapped action rate & 0.00129722 \\
finite-action-code behavioral benchmark & structured-control irrelevant false-positive rate & 0.0083022 \\
finite-action-code behavioral benchmark & structured-control behavioral composite $>$ high-stochasticity datasets & 7 \\
finite-action-code behavioral benchmark & structured-control B\_RSI $>$ reason-lesion datasets & 7 \\
finite-action-code behavioral benchmark & structured-control B\_VEI $>$ veto-lesion datasets & 7 \\
finite-action-code behavioral benchmark & structured-control B\_MCI $>$ high-stochasticity datasets & 7 \\
finite-action-code behavioral benchmark & structured-control B\_SCI $>$ high-stochasticity datasets & 7 \\
finite-action-code behavioral benchmark & dataset-cluster bootstrap lower 95\% bound & 0.855732 \\
\end{longtable}
\normalsize

\begin{landscape}
\tiny
\setlength{\tabcolsep}{1.5pt}
\renewcommand{\arraystretch}{1.12}
\begin{longtable}{@{}L{0.080\linewidth}L{0.046\linewidth}L{0.046\linewidth}L{0.046\linewidth}L{0.046\linewidth}L{0.046\linewidth}L{0.046\linewidth}L{0.046\linewidth}L{0.046\linewidth}L{0.046\linewidth}L{0.046\linewidth}L{0.046\linewidth}L{0.046\linewidth}L{0.046\linewidth}L{0.046\linewidth}L{0.046\linewidth}@{}}
\caption{Behavioral component scores summarized by variant.}\label{tab:s4_behavior_variant}\\
\toprule
Variant & RSI \(\mu\) & MCI \(\mu\) & VEI \(\mu\) & SCI \(\mu\) & Cue \(\mu\) & Comp. \(\mu\) & FP \(\mu\) & RSI \(\sigma\) & MCI \(\sigma\) & VEI \(\sigma\) & SCI \(\sigma\) & Cue \(\sigma\) & Comp. \(\sigma\) & FP \(\sigma\) & Sets \\
\midrule
\endfirsthead
\caption[]{Behavioral component scores summarized by variant. (continued)}\\
\toprule
Variant & RSI \(\mu\) & MCI \(\mu\) & VEI \(\mu\) & SCI \(\mu\) & Cue \(\mu\) & Comp. \(\mu\) & FP \(\mu\) & RSI \(\sigma\) & MCI \(\sigma\) & VEI \(\sigma\) & SCI \(\sigma\) & Cue \(\sigma\) & Comp. \(\sigma\) & FP \(\sigma\) & Sets \\
\midrule
\endhead
\midrule
\multicolumn{16}{r}{Continued on next page}\\
\endfoot
\bottomrule
\endlastfoot
A4 & 0.973 & 0.773 & 0.989 & 0.990 & 0.931 & 0.931 & 0.008 & 0.023 & 0.151 & 0.019 & 0.021 & 0.037 & 0.037 & 0.019 & 7 \\
\ttbreak{A4\_no}{reason} & -0.044 & 0.788 & 0.954 & 0.954 & 0.663 & 0.663 & 0.044 & 0.113 & 0.089 & 0.112 & 0.113 & 0.102 & 0.102 & 0.113 & 7 \\
\ttbreak{A4\_no}{veto} & 0.977 & 0.801 & -0.008 & 0.990 & 0.690 & 0.690 & 0.009 & 0.019 & 0.081 & 0.019 & 0.021 & 0.020 & 0.020 & 0.019 & 7 \\
A5 & 0.002 & -0.001 & -0.001 & 0.179 & 0.045 & 0.045 & 0.001 & 0.004 & 0.001 & 0.001 & 0.101 & 0.025 & 0.025 & 0.001 & 7 \\
\end{longtable}
\normalsize
\end{landscape}

\begin{landscape}
\tiny
\setlength{\tabcolsep}{2pt}
\renewcommand{\arraystretch}{1.12}
\begin{longtable}{@{}L{0.120\linewidth}L{0.085\linewidth}L{0.085\linewidth}L{0.085\linewidth}L{0.085\linewidth}L{0.085\linewidth}L{0.085\linewidth}L{0.085\linewidth}L{0.085\linewidth}@{}}
\caption{Behavioral component scores by dataset namespace and variant.}\label{tab:s4_behavior_dataset}\\
\toprule
Dataset & Variant & B\_RSI & B\_MCI & B\_VEI & B\_SCI & Cue & Composite & FP \\
\midrule
\endfirsthead
\caption[]{Behavioral component scores by dataset namespace and variant. (continued)}\\
\toprule
Dataset & Variant & B\_RSI & B\_MCI & B\_VEI & B\_SCI & Cue & Composite & FP \\
\midrule
\endhead
\midrule
\multicolumn{9}{r}{Continued on next page}\\
\endfoot
\bottomrule
\endlastfoot
ds000210\_self\_loop & A4 & 0.969444 & 0.808333 & 0.988889 & 0.994444 & 0.940278 & 0.940278 & 0.00277778 \\
ds000210\_self\_loop & A4\_no\_reason & 0 & 0.813889 & 0.997222 & 0.997222 & 0.702083 & 0.702083 & 0 \\
ds000210\_self\_loop & A4\_no\_veto & 0.969444 & 0.788889 & -0.00277778 & 0.994444 & 0.6875 & 0.6875 & 0.00277778 \\
ds000210\_self\_loop & A5 & 0 & 0 & 0 & 0.166667 & 0.0416667 & 0.0416667 & 0 \\
ds000212\_choice & A4 & 0.977778 & 0.555556 & 0.997222 & 1 & 0.882639 & 0.882639 & 0 \\
ds000212\_choice & A4\_no\_reason & -0.00277778 & 0.738889 & 0.994444 & 0.997222 & 0.681944 & 0.681944 & 0.00277778 \\
ds000212\_choice & A4\_no\_veto & 0.988889 & 0.672222 & 0 & 1 & 0.665278 & 0.665278 & 0 \\
ds000212\_choice & A5 & 0 & 0 & 0 & 0.161111 & 0.0402778 & 0.0402778 & 0 \\
\end{longtable}
\normalsize
\end{landscape}

\begin{landscape}
\scriptsize
\setlength{\tabcolsep}{3pt}
\renewcommand{\arraystretch}{1.12}
\begin{longtable}{@{}L{0.110\linewidth}L{0.077\linewidth}L{0.077\linewidth}L{0.077\linewidth}L{0.077\linewidth}L{0.077\linewidth}L{0.077\linewidth}L{0.077\linewidth}L{0.077\linewidth}L{0.077\linewidth}@{}}
\caption{Predefined dataset-direction behavioral criterion contrasts.}\label{tab:s4_behavior_contrasts}\\
\toprule
Dataset & structured comp. & reason-lesion comp. & veto-lesion comp. & HS comp. & structured-HS comp. & reason contrast & veto contrast & memory contrast & self contrast \\
\midrule
\endfirsthead
\caption[]{Predefined dataset-direction behavioral criterion contrasts. (continued)}\\
\toprule
Dataset & structured comp. & reason-lesion comp. & veto-lesion comp. & HS comp. & structured-HS comp. & reason contrast & veto contrast & memory contrast & self contrast \\
\midrule
\endhead
\midrule
\multicolumn{10}{r}{Continued on next page}\\
\endfoot
\bottomrule
\endlastfoot
ds000210\_self\_loop & 0.940278 & 0.702083 & 0.6875 & 0.0416667 & 0.898611 & 0.969444 & 0.991667 & 0.808333 & 0.827778 \\
ds000212\_choice & 0.882639 & 0.681944 & 0.665278 & 0.0402778 & 0.842361 & 0.980556 & 0.997222 & 0.555556 & 0.838889 \\
ds001618\_self\_loop & 0.949306 & 0.711806 & 0.713194 & 0.0986111 & 0.850694 & 0.986111 & 0.997222 & 0.816667 & 0.608333 \\
ds001882\_choice & 0.960417 & 0.709028 & 0.711806 & 0.0229167 & 0.9375 & 0.997222 & 0.994444 & 0.852778 & 0.905556 \\
ds004042\_recall & 0.931373 & 0.433473 & 0.677871 & 0.039916 & 0.891457 & 1.22689 & 0.997199 & 0.907563 & 0.784314 \\
\ttbreak{ds004917}{uncertainty} & 0.974 & 0.723 & 0.704 & 0.024 & 0.951 & 0.992 & 1 & 0.908 & 0.903 \\
narratives\_formal & 0.880952 & 0.681548 & 0.669643 & 0.0486111 & 0.832341 & 0.964286 & 1 & 0.56746 & 0.809524 \\
\end{longtable}
\normalsize
\end{landscape}

\clearpage\section*{S5. Strong Stochastic Controls and Entropy Diagnostics}

Strong stochastic controls test whether alternative stochastic implementations recover the structured-control profile. Entropy diagnostics summarize action variability under raw strings, rule-canonical actions, semantic clusters and action-family mappings. The calibration tables document that the predefined entropy-closeness criterion was not achieved.

\scriptsize
\setlength{\tabcolsep}{2pt}
\renewcommand{\arraystretch}{1.12}
\begin{longtable}{@{}L{0.160\linewidth}L{0.210\linewidth}L{0.210\linewidth}L{0.250\linewidth}L{0.130\linewidth}@{}}
\caption{Proxy token/compute accounting summary. Provider-side prompt-token, completion-token and total-token usage were not uniformly retained; this table reports availability and proxy measurements for length and structure.}\label{tab:s5_proxy_compute}\\
\toprule
Audit item & Measurement or proxy & Availability or value & Evidence source & Interpretation \\
\midrule
\endfirsthead
\caption[]{Proxy token/compute accounting summary. Provider-side prompt-token, completion-token and total-token usage were not uniformly retained; this table reports availability and proxy measurements for length and structure. (continued)}\\
\toprule
Audit item & Measurement or proxy & Availability or value & Evidence source & Interpretation \\
\midrule
\endhead
\midrule
\multicolumn{5}{r}{Continued on next page}\\
\endfoot
\bottomrule
\endlastfoot
provider token usage & prompt\_tokens, completion\_tokens, total\_tokens & not uniformly retained & token\_verbosity\_audit criterion summary reports provider token usage unavailable & provider-token accounting incomplete \\
prompt length & model-visible prompt character or token count & not uniformly retained & no complete prompt-length column in retained token audit summary & prompt-token matching unavailable \\
completion length proxy & json\_trace\_chars, final\_action\_chars, first\_impulse\_chars & 84,516 trace rows; 552 source/dataset/variant summaries & retained token-verbosity audit table & field-length/verbosity proxy only \\
trace structure proxy & reason\_node\_count, branch\_count, memory\_item\_count & 84,516 trace rows; 552 summaries & retained token-verbosity audit table & structure proxy for compute-risk assessment \\
schema field count & matched-interface emitted JSON schema & nine external fields in matched-interface protocol & matched-interface protocol schema & controls schema availability under the matched interface \\
retry count & provider retry attempts & not uniformly retained & unrecovered generations are reported separately by experiment & retry-cost inference unavailable \\
latency proxy & latency\_ms & available where retained in token\_verbosity\_audit & wall-clock proxy affected by provider and network conditions & indirect model-compute/cost proxy \\
compressed structured-control sensitivity & AFCI\_mean & A4\_full=0.500; A4\_compressed=0.499 & matched-interface AFCI table & coupling persisted under compressed trace expression \\
\end{longtable}
\normalsize

\begin{landscape}
\tiny
\setlength{\tabcolsep}{2pt}
\renewcommand{\arraystretch}{1.12}
\begin{longtable}{@{}L{0.095\linewidth}L{0.058\linewidth}L{0.058\linewidth}L{0.058\linewidth}L{0.058\linewidth}L{0.058\linewidth}L{0.058\linewidth}L{0.058\linewidth}L{0.058\linewidth}L{0.058\linewidth}L{0.058\linewidth}L{0.058\linewidth}L{0.058\linewidth}@{}}
\caption{Robustness and stochastic-control criterion summary. Prompt-paraphrase, expanded counterfactual and placebo rows are auxiliary robustness blocks. They are distinct from the formal counterfactual-transform quality-control screen and the consistency-passing transform reported in Supplementary Table~\ref{tab:s33_validation}.}\label{tab:s5_robust_gate}\\
\toprule
Experiment & Status & Calls & Remote & Unrec. & Unrec. rate & Min ds & HS rec. & AFCI ds & Parse n & Parse r & Kappa & Leak \\
\midrule
\endfirsthead
\caption[]{Robustness and stochastic-control criterion summary. Auxiliary robustness rows are distinct from the formal counterfactual-transform quality-control screen and consistency-passing transform. (continued)}\\
\toprule
Experiment & Status & Calls & Remote & Unrec. & Unrec. rate & Min ds & HS rec. & AFCI ds & Parse n & Parse r & Kappa & Leak \\
\midrule
\endhead
\midrule
\multicolumn{13}{r}{Continued on next page}\\
\endfoot
\bottomrule
\endlastfoot
strong stochastic controls & pass & 15490 & 15490 & 0 & 0 & 7 & 0 & -- & -- & -- & -- & -- \\
\ttbreak{A5\_empirical}{entropy\_attempt} & predefined closeness not satisfied & 17082 & 17082 & 0 & 0 & 7 & 0 & -- & -- & -- & -- & -- \\
prompt paraphrases & pass & 11400 & 11400 & 0 & 0 & 17 & 0 & -- & -- & -- & -- & -- \\
expanded counterfactual probes & auxiliary pass & 30584 & 30584 & 0 & 0 & 61 & 0 & -- & -- & -- & -- & -- \\
placebo components & pass & 9960 & 9960 & 0 & 0 & 24 & 0 & -- & -- & -- & -- & -- \\
cross-model subset & pass & -- & -- & -- & -- & -- & -- & -- & -- & -- & -- & -- \\
open-weight reference model & pass & -- & -- & -- & -- & -- & -- & 7 & 8 & 0.000762 & -- & -- \\
expanded blinded audit & pass & -- & -- & -- & -- & -- & -- & -- & -- & -- & 0.961 & 0 \\
\end{longtable}
\normalsize
\end{landscape}

\begin{landscape}
\tiny
\setlength{\tabcolsep}{2pt}
\renewcommand{\arraystretch}{1.12}
\begin{longtable}{@{}L{0.105\linewidth}L{0.070\linewidth}L{0.070\linewidth}L{0.070\linewidth}L{0.070\linewidth}L{0.070\linewidth}L{0.070\linewidth}L{0.070\linewidth}L{0.070\linewidth}L{0.070\linewidth}L{0.070\linewidth}@{}}
\caption{Entropy diagnostics summarized by experiment source and variant.}\label{tab:s5_entropy_source}\\
\toprule
Source & Variant & \plainbreak{Raw H}{mean} & \plainbreak{Rule H}{mean} & \plainbreak{Semantic H}{mean} & \plainbreak{Family H}{mean} & \plainbreak{Raw H}{SD} & \plainbreak{Rule H}{SD} & \plainbreak{Semantic H}{SD} & \plainbreak{Family H}{SD} & \plainbreak{Aggregate}{rows} \\
\midrule
\endfirsthead
\caption[]{Entropy diagnostics summarized by experiment source and variant. (continued)}\\
\toprule
Source & Variant & \plainbreak{Raw H}{mean} & \plainbreak{Rule H}{mean} & \plainbreak{Semantic H}{mean} & \plainbreak{Family H}{mean} & \plainbreak{Raw H}{SD} & \plainbreak{Rule H}{SD} & \plainbreak{Semantic H}{SD} & \plainbreak{Family H}{SD} & \plainbreak{Aggregate}{rows} \\
\midrule
\endhead
\midrule
\multicolumn{11}{r}{Continued on next page}\\
\endfoot
\bottomrule
\endlastfoot
counterfactual probes & A4 & 4.180 & 2.647 & 3.962 & 0.863 & 1.703 & 1.490 & 1.625 & 0.461 & 70 \\
counterfactual probes & \ttbreak{A4\_no}{reason} & 4.013 & 2.549 & 3.851 & 0.883 & 1.753 & 1.543 & 1.683 & 0.512 & 70 \\
counterfactual probes & \ttbreak{A4\_no}{veto} & 4.179 & 2.591 & 3.969 & 0.913 & 1.734 & 1.508 & 1.709 & 0.490 & 70 \\
counterfactual probes & A5 & 5.519 & 3.362 & 5.445 & 1.086 & 1.410 & 1.738 & 1.432 & 0.452 & 70 \\
\ttbreak{A5\_emp.}{entropy} & A4 & 6.596 & 5.106 & 6.360 & 1.070 & 1.292 & 1.541 & 1.301 & 0.411 & 21 \\
\ttbreak{A5\_emp.}{entropy} & A5 & 7.569 & 5.600 & 7.514 & 1.241 & 0.956 & 1.390 & 0.979 & 0.131 & 21 \\
\ttbreak{A5\_emp.}{entropy} & \ttbreak{A5\_emp.}{entropy} & 7.584 & 5.477 & 7.529 & 1.236 & 0.924 & 1.432 & 0.951 & 0.134 & 21 \\
placebo\_components & A4 & 4.445 & 3.443 & 4.312 & 0.822 & 1.594 & 1.628 & 1.546 & 0.393 & 33 \\
\end{longtable}
\normalsize
\end{landscape}

\scriptsize
\setlength{\tabcolsep}{3pt}
\renewcommand{\arraystretch}{1.12}
\begin{longtable}{@{}L{0.250\linewidth}L{0.320\linewidth}L{0.350\linewidth}@{}}
\caption{Predefined empirical entropy-attempt calibration criterion summary.}\label{tab:s5_strict_gate}\\
\toprule
Criterion & Field & Value \\
\midrule
\endfirsthead
\caption[]{Predefined empirical entropy-attempt calibration criterion summary. (continued)}\\
\toprule
Criterion & Field & Value \\
\midrule
\endhead
\midrule
\multicolumn{3}{r}{Continued on next page}\\
\endfoot
\bottomrule
\endlastfoot
strict\_a5\_em\_calibration & calibration\_events\_per\_dataset & 50 \\
strict\_a5\_em\_calibration & closeness\_threshold & 0.15 \\
strict\_a5\_em\_calibration & grid\_coverage\_status & complete \\
strict\_a5\_em\_calibration & max\_selected\_strict\_entropy\_gap & 1.17146 \\
strict\_a5\_em\_calibration & note & raw/canonical/semantic/action-family closeness did not meet threshold \\
strict\_a5\_em\_calibration & observed\_required\_setting\_count & 28 \\
strict\_a5\_em\_calibration & required\_setting\_count & 28 \\
strict\_a5\_em\_calibration & selected\_temperature & 0.2 \\
strict\_a5\_em\_calibration & selected\_top\_p & 0.75 \\
strict\_a5\_em\_calibration & selected\_variant & A5\_attempt\_t020\_p075 \\
strict\_a5\_em\_calibration & status & predefined closeness not satisfied \\
strict\_a5\_em\_calibration & strict\_entropy\_close\_enough & False \\
strict\_a5\_em\_calibration & wording\_rule & Entropy matching is reported only when the predefined closeness criterion is satisfied; here the empirical attempt did not satisfy that criterion. \\
\end{longtable}
\normalsize

\begin{landscape}
\tiny
\setlength{\tabcolsep}{2pt}
\renewcommand{\arraystretch}{1.12}
\begin{longtable}{@{}L{0.120\linewidth}L{0.085\linewidth}L{0.085\linewidth}L{0.085\linewidth}L{0.085\linewidth}L{0.085\linewidth}L{0.085\linewidth}L{0.085\linewidth}L{0.085\linewidth}@{}}
\caption{Predefined empirical entropy-attempt calibration grid.}\label{tab:s5_strict_grid}\\
\toprule
Variant & Temp. & Top \(p\) & \plainbreak{Raw H}{gap} & \plainbreak{Rule H}{gap} & \plainbreak{Semantic H}{gap} & \plainbreak{Family H}{gap} & \plainbreak{Strict}{score} & \plainbreak{Max strict}{gap} \\
\midrule
\endfirsthead
\caption[]{Predefined empirical entropy-attempt calibration grid. (continued)}\\
\toprule
Variant & Temp. & Top \(p\) & \plainbreak{Raw H}{gap} & \plainbreak{Rule H}{gap} & \plainbreak{Semantic H}{gap} & \plainbreak{Family H}{gap} & \plainbreak{Strict}{score} & \plainbreak{Max strict}{gap} \\
\midrule
\endhead
\midrule
\multicolumn{9}{r}{Continued on next page}\\
\endfoot
\bottomrule
\endlastfoot
\ttbreak{A5\_attempt}{t020\_p060} & 0.2 & 0.6 & 0.944 & 1.300 & 1.096 & 0.302 & 0.910 & 1.300 \\
\ttbreak{A5\_attempt}{t020\_p075} & 0.2 & 0.75 & 0.973 & 1.171 & 1.110 & 0.353 & 0.902 & 1.171 \\
\ttbreak{A5\_attempt}{t020\_p090} & 0.2 & 0.9 & 1.102 & 1.420 & 1.224 & 0.404 & 1.037 & 1.420 \\
\ttbreak{A5\_attempt}{t020\_p100} & 0.2 & 1 & 0.955 & 1.331 & 1.094 & 0.365 & 0.936 & 1.331 \\
\ttbreak{A5\_attempt}{t040\_p060} & 0.4 & 0.6 & 1.125 & 1.545 & 1.271 & 0.388 & 1.082 & 1.545 \\
\ttbreak{A5\_attempt}{t040\_p075} & 0.4 & 0.75 & 1.154 & 1.462 & 1.295 & 0.249 & 1.040 & 1.462 \\
\ttbreak{A5\_attempt}{t040\_p090} & 0.4 & 0.9 & 1.070 & 1.328 & 1.217 & 0.362 & 0.994 & 1.328 \\
\ttbreak{A5\_attempt}{t040\_p100} & 0.4 & 1 & 1.186 & 1.334 & 1.314 & 0.317 & 1.038 & 1.334 \\
\end{longtable}
\normalsize
\end{landscape}

\scriptsize
\setlength{\tabcolsep}{3pt}
\renewcommand{\arraystretch}{1.12}
\begin{longtable}{@{}L{0.250\linewidth}L{0.320\linewidth}L{0.350\linewidth}@{}}
\caption{Finite-action-code entropy-calibration criterion summary.}\label{tab:s5_calib_gate}\\
\toprule
Criterion & Field & Value \\
\midrule
\endfirsthead
\caption[]{Finite-action-code entropy-calibration criterion summary. (continued)}\\
\toprule
Criterion & Field & Value \\
\midrule
\endhead
\midrule
\multicolumn{3}{r}{Continued on next page}\\
\endfoot
\bottomrule
\endlastfoot
finite-action entropy calibration & structured-control behavioral composite $>$ selected high-stochasticity datasets & 7 \\
finite-action entropy calibration & behavior pass threshold & 6 \\
finite-action entropy calibration & calibration events per dataset & 14 \\
finite-action entropy calibration & calibration maximum predefined entropy gap & 1.22474 \\
finite-action entropy calibration & calibration mean predefined entropy gap & 1.13035 \\
finite-action entropy calibration & calls & 11088 \\
finite-action entropy calibration & closeness threshold & 0.15 \\
finite-action entropy calibration & dataset count & 7 \\
\end{longtable}
\normalsize

\begin{landscape}
\tiny
\setlength{\tabcolsep}{2pt}
\renewcommand{\arraystretch}{1.12}
\begin{longtable}{@{}L{0.105\linewidth}L{0.070\linewidth}L{0.070\linewidth}L{0.070\linewidth}L{0.070\linewidth}L{0.070\linewidth}L{0.070\linewidth}L{0.070\linewidth}L{0.070\linewidth}L{0.070\linewidth}L{0.070\linewidth}@{}}
\caption{Finite-action-code entropy-calibration grid.}\label{tab:s5_calib_grid}\\
\toprule
Variant & Sets & Rows & Mean gap & Max gap & Unmap & Raw H & Code H & Family H & Temp. & Top \(p\) \\
\midrule
\endfirsthead
\caption[]{Finite-action-code entropy-calibration grid. (continued)}\\
\toprule
Variant & Sets & Rows & Mean gap & Max gap & Unmap & Raw H & Code H & Family H & Temp. & Top \(p\) \\
\midrule
\endhead
\midrule
\multicolumn{11}{r}{Continued on next page}\\
\endfoot
\bottomrule
\endlastfoot
\ttbreak{A5\_attempt}{t01\_p07} & 7 & 588 & 1.138 & 1.189 & 0 & 1.138 & 1.138 & 1.138 & 0.1 & 0.7 \\
\ttbreak{A5\_attempt}{t01\_p10} & 7 & 588 & 1.171 & 1.249 & 0.002 & 1.171 & 1.171 & 1.171 & 0.1 & 1 \\
\ttbreak{A5\_attempt}{t02\_p07} & 7 & 588 & 1.130 & 1.225 & 0.002 & 1.130 & 1.130 & 1.130 & 0.2 & 0.7 \\
\ttbreak{A5\_attempt}{t02\_p10} & 7 & 588 & 1.142 & 1.250 & 0 & 1.142 & 1.142 & 1.142 & 0.2 & 1 \\
\ttbreak{A5\_attempt}{t04\_p07} & 7 & 588 & 1.139 & 1.189 & 0 & 1.139 & 1.139 & 1.139 & 0.4 & 0.7 \\
\ttbreak{A5\_attempt}{t04\_p10} & 7 & 588 & 1.164 & 1.219 & 0 & 1.164 & 1.164 & 1.164 & 0.4 & 1 \\
\ttbreak{A5\_attempt}{t06\_p07} & 7 & 588 & 1.162 & 1.250 & 0.002 & 1.162 & 1.162 & 1.162 & 0.6 & 0.7 \\
\ttbreak{A5\_attempt}{t06\_p10} & 7 & 588 & 1.161 & 1.223 & 0 & 1.161 & 1.161 & 1.161 & 0.6 & 1 \\
\end{longtable}
\normalsize
\end{landscape}

\begin{landscape}
\tiny
\setlength{\tabcolsep}{2pt}
\renewcommand{\arraystretch}{1.12}
\begin{longtable}{@{}L{0.105\linewidth}L{0.070\linewidth}L{0.070\linewidth}L{0.070\linewidth}L{0.070\linewidth}L{0.070\linewidth}L{0.070\linewidth}L{0.070\linewidth}L{0.070\linewidth}L{0.070\linewidth}L{0.070\linewidth}@{}}
\caption{Finite-action-code entropy-calibration holdout grid.}\label{tab:s5_holdout_grid}\\
\toprule
Variant & Sets & Rows & Mean gap & Max gap & Unmap & Raw H & Code H & Family H & Temp. & Top \(p\) \\
\midrule
\endfirsthead
\caption[]{Finite-action-code entropy-calibration holdout grid. (continued)}\\
\toprule
Variant & Sets & Rows & Mean gap & Max gap & Unmap & Raw H & Code H & Family H & Temp. & Top \(p\) \\
\midrule
\endhead
\midrule
\multicolumn{11}{r}{Continued on next page}\\
\endfoot
\bottomrule
\endlastfoot
A5 & 7 & 756 & 1.165 & 1.299 & 0 & 1.165 & 1.165 & 1.165 & -- & -- \\
\ttbreak{A5\_attempt}{t02\_p07} & 7 & 756 & 1.147 & 1.348 & 0 & 1.147 & 1.147 & 1.147 & 0.2 & 0.7 \\
\end{longtable}
\normalsize
\end{landscape}

\clearpage\section*{S6. Robustness, Open-Weight Reference Model and Blinded Audit}

Robustness checks evaluate the direction of the result across tested prompts, tested model identifiers, one local open-weight reference model and blinded annotation. These checks support robustness across the tested settings; they do not imply coverage of untested models or prompts. The agreement table reports Cohen's \(\kappa\), a chance-adjusted nominal-agreement statistic \citep{cohen1960coefficient}.

\begin{landscape}
\tiny
\setlength{\tabcolsep}{2pt}
\renewcommand{\arraystretch}{1.12}
\begin{longtable}{@{}L{0.090\linewidth}L{0.054\linewidth}L{0.054\linewidth}L{0.054\linewidth}L{0.054\linewidth}L{0.054\linewidth}L{0.054\linewidth}L{0.054\linewidth}L{0.054\linewidth}L{0.054\linewidth}L{0.054\linewidth}L{0.054\linewidth}L{0.054\linewidth}L{0.054\linewidth}@{}}
\caption{Cross-model subset criterion summary.}\label{tab:s6_cross_model}\\
\toprule
Model & Status & Exp. & Profile & Calls & Remote & Unrecov. & Unrec. rate & Vars. & Ctrls. & Min ctrl ds & Strong HS ds & Fallback & SC>HS ds \\
\midrule
\endfirsthead
\caption[]{Cross-model subset criterion summary. (continued)}\\
\toprule
Model & Status & Exp. & Profile & Calls & Remote & Unrecov. & Unrec. rate & Vars. & Ctrls. & Min ctrl ds & Strong HS ds & Fallback & SC>HS ds \\
\midrule
\endhead
\midrule
\multicolumn{14}{r}{Continued on next page}\\
\endfoot
\bottomrule
\endlastfoot
\ttbreak{gpt-5.4}{mini} & pass & cross subset & full & 1995 & 1995 & 0 & 0 & 5 & 4 & 5 & 0 & \ttbreak{qwen3}{30b-a3b} & 7 \\
\ttbreak{gpt-4.1}{mini} & pass & cross subset & full & 1995 & 1995 & 0 & 0 & 5 & 4 & 6 & 0 & \ttbreak{qwen3}{30b-a3b} & 7 \\
deepseek-chat & pass & cross subset & full & 1995 & 1995 & 0 & 0 & 5 & 4 & 6 & 0 & \ttbreak{qwen3}{30b-a3b} & 7 \\
\ttbreak{gemini-2.5}{flash} & pass & cross subset & full & 1995 & 1995 & 2 & 0.001 & 5 & 4 & 7 & 0 & \ttbreak{qwen3}{30b-a3b} & 7 \\
\end{longtable}
\normalsize
\end{landscape}

\begin{landscape}
\tiny
\setlength{\tabcolsep}{2pt}
\renewcommand{\arraystretch}{1.12}
\begin{longtable}{@{}L{0.080\linewidth}L{0.046\linewidth}L{0.046\linewidth}L{0.046\linewidth}L{0.046\linewidth}L{0.046\linewidth}L{0.046\linewidth}L{0.046\linewidth}L{0.046\linewidth}L{0.046\linewidth}L{0.046\linewidth}L{0.046\linewidth}L{0.046\linewidth}L{0.046\linewidth}L{0.046\linewidth}L{0.046\linewidth}@{}}
\caption{Open-weight Qwen2.5-14B-Instruct AFCI summary by variant.}\label{tab:s6_open_weight}\\
\toprule
Variant & R \(\mu\) & M \(\mu\) & Veto \(\mu\) & Self \(\mu\) & Supp. \(\mu\) & AFCI \(\mu\) & AFCI+ \(\mu\) & R \(\sigma\) & M \(\sigma\) & Veto \(\sigma\) & Self \(\sigma\) & Supp. \(\sigma\) & AFCI \(\sigma\) & AFCI+ \(\sigma\) & Rows \\
\midrule
\endfirsthead
\caption[]{Open-weight Qwen2.5-14B-Instruct AFCI summary by variant. (continued)}\\
\toprule
Variant & R \(\mu\) & M \(\mu\) & Veto \(\mu\) & Self \(\mu\) & Supp. \(\mu\) & AFCI \(\mu\) & AFCI+ \(\mu\) & R \(\sigma\) & M \(\sigma\) & Veto \(\sigma\) & Self \(\sigma\) & Supp. \(\sigma\) & AFCI \(\sigma\) & AFCI+ \(\sigma\) & Rows \\
\midrule
\endhead
\midrule
\multicolumn{16}{r}{Continued on next page}\\
\endfoot
\bottomrule
\endlastfoot
A4 & 1 & 0.5 & 1 & 1 & 1 & 0.875 & 0.9 & 0 & 0 & 0 & 0 & 0 & 0 & 0 & 21 \\
\ttbreak{A4\_no}{reason} & 0.5 & 0.5 & 1 & 1 & 1 & 0.75 & 0.8 & 0 & 0 & 0 & 0 & 0 & 0 & 0 & 21 \\
\ttbreak{A4\_no}{veto} & 1 & 0.5 & 0.5 & 1 & 1 & 0.75 & 0.8 & 0 & 0 & 0 & 0 & 0 & 0 & 0 & 21 \\
A5 & 0.446 & 0.25 & 0.5 & 0.393 & 0 & 0.397 & 0.318 & 0.019 & 0 & 0 & 0.031 & 0 & 0.008 & 0.006 & 21 \\
\ttbreak{A5\_schema}{posthoc\_fields} & 0 & 0 & 0 & 0 & 0 & 0 & 0 & 0 & 0 & 0 & 0 & 0 & 0 & 0 & 21 \\
\end{longtable}
\normalsize
\end{landscape}

\begin{landscape}
\scriptsize
\setlength{\tabcolsep}{3pt}
\renewcommand{\arraystretch}{1.12}
\begin{longtable}{@{}L{0.160\linewidth}L{0.160\linewidth}L{0.160\linewidth}L{0.160\linewidth}L{0.160\linewidth}L{0.120\linewidth}@{}}
\caption{Expanded blinded audit reliability by annotation dimension.}\label{tab:s6_blind}\\
\toprule
Dimension & Annotator A & Annotator B & Overlap items & Agreement & Cohen kappa \\
\midrule
\endfirsthead
\caption[]{Expanded blinded audit reliability by annotation dimension. (continued)}\\
\toprule
Dimension & Annotator A & Annotator B & Overlap items & Agreement & Cohen kappa \\
\midrule
\endhead
\midrule
\multicolumn{6}{r}{Continued on next page}\\
\endfoot
\bottomrule
\endlastfoot
final action reason-consistent & annotator\_01 & annotator\_02 & 1200 & 0.958333 & 0.912127 \\
final action reason-consistent & annotator\_01 & annotator\_03 & 1200 & 0.955833 & 0.907158 \\
final action reason-consistent & annotator\_02 & annotator\_03 & 1200 & 0.990833 & 0.980522 \\
action respects relevant memory & annotator\_01 & annotator\_02 & 1200 & 0.997500 & 0.989211 \\
action respects relevant memory & annotator\_01 & annotator\_03 & 1200 & 0.997500 & 0.989211 \\
action respects relevant memory & annotator\_02 & annotator\_03 & 1200 & 1.000000 & 1.000000 \\
veto\_was\_appropriate & annotator\_01 & annotator\_02 & 1200 & 1.000000 & 1.000000 \\
veto\_was\_appropriate & annotator\_01 & annotator\_03 & 1200 & 1.000000 & 1.000000 \\
veto\_was\_appropriate & annotator\_02 & annotator\_03 & 1200 & 1.000000 & 1.000000 \\
veto changed first impulse & annotator\_01 & annotator\_02 & 1200 & 1.000000 & 1.000000 \\
veto changed first impulse & annotator\_01 & annotator\_03 & 1200 & 1.000000 & 1.000000 \\
veto changed first impulse & annotator\_02 & annotator\_03 & 1200 & 1.000000 & 1.000000 \\
reason\_looks\_posthoc & annotator\_01 & annotator\_02 & 1200 & 0.960000 & 0.917801 \\
reason\_looks\_posthoc & annotator\_01 & annotator\_03 & 1200 & 0.961667 & 0.921303 \\
reason\_looks\_posthoc & annotator\_02 & annotator\_03 & 1200 & 0.995000 & 0.989652 \\
verbose/template-like without control & annotator\_01 & annotator\_02 & 1200 & 0.998333 & 0.971364 \\
verbose/template-like without control & annotator\_01 & annotator\_03 & 1200 & 0.996667 & 0.942728 \\
verbose/template-like without control & annotator\_02 & annotator\_03 & 1200 & 0.998333 & 0.972113 \\
irrelevant cue caused inappropriate change & annotator\_01 & annotator\_02 & 1200 & 0.974167 & 0.865842 \\
irrelevant cue caused inappropriate change & annotator\_01 & annotator\_03 & 1200 & 0.972500 & 0.860015 \\
irrelevant cue caused inappropriate change & annotator\_02 & annotator\_03 & 1200 & 0.993333 & 0.965260 \\
\end{longtable}
\normalsize
\end{landscape}

Annotation ethics scope: three trained annotators contributed output-level quality-control labels under a fixed rubric and full-overlap blinded design. The audit did not collect personal, behavioral, health, intervention or neural data from annotators or any other individuals. Annotation was performed as expert quality-control scoring of model outputs for the project record rather than as human-subjects data collection.

\clearpage\section*{S7. Validation Source-Data Map, Reproducibility and Export Coverage}

The reproducibility map links each claim-supporting statement to its table, result export, criterion summary, analysis entry point and checksum. Full SHA-256 values are retained in the corresponding CSV export; the PDF table reports the first 12 hexadecimal characters for readability. The validation source-data map below gives the direct source-data and script paths for the main validation figure, the SWE-bench Lite reliability displays and the formal control definitions.

\begin{landscape}
\scriptsize
\setlength{\tabcolsep}{3pt}
\renewcommand{\arraystretch}{1.12}
\begin{longtable}{@{}L{0.180\linewidth}L{0.250\linewidth}L{0.430\linewidth}L{0.090\linewidth}@{}}
\caption{Validation source-data and control-construction map.}\label{tab:s7_validation_source_map}\\
\toprule
Item & Role & Source data, config or script path & Checksum or provenance \\
\midrule
\endfirsthead
\caption[]{Validation source-data and control-construction map. (continued)}\\
\toprule
Item & Role & Source data, config or script path & Checksum or provenance \\
\midrule
\endhead
\midrule
\multicolumn{4}{r}{Continued on next page}\\
\endfoot
\bottomrule
\endlastfoot
Main validation panel a source data & Open-weight core no-context controls & \path{manuscript/figures/source_data/figure5a_open_weight_core_controls.csv} & 12ef8d181136 \\
Main validation panel b source data & Open-weight state-boundary lesion & \path{manuscript/figures/source_data/figure5b_open_weight_state_boundary_lesion.csv} & 803140cd02ce \\
Main validation panel c source data & SWE-bench Lite full 300-record incremental AUC & \path{manuscript/figures/source_data/figure5c_swebench_incremental_auc.csv} & 0cd8461a01b1 \\
Main validation panel d source data & SWE-bench Lite single-predictor AUC comparison & \path{manuscript/figures/source_data/figure5d_swebench_single_predictor_auc.csv} & 9ab7ba88b7d0 \\
Main validation panel e source data & SWE-bench Lite sensitivity and binding-guard summary & \path{manuscript/figures/source_data/figure5e_swebench_sensitivity_guard.csv} & b76fc1220aab \\
Main validation rendering and export QA & Figure rendering and export-QA script & \path{manuscript/figures/scripts/build_validation_figure5.R}; \path{audit/figure5_primary_validation_export_qc_20260530.csv} & 61d69f9d8264 / 80f8d170bb37 \\
SWE-bench reliability calibration source data & Issue-grouped five-fold reliability calibration for the full 300-record issue-to-file endpoint & \path{manuscript/figures/source_data/extended_data_figure9_swebench_reliability_calibration.csv}; \path{manuscript/figures/source_data/swebench_cv_reliability_predictions.csv} & archived manifest \\
SWE-bench threshold analysis source data & Thresholded localization-guard analysis over predicted hit@3 probabilities & \path{manuscript/figures/source_data/extended_data_figure10_swebench_threshold_analysis.csv}; \path{manuscript/figures/scripts/build_swebench_reliability_figures.R} & archived manifest \\
Main Table 1 source data & Claim ledger with supported claims and excluded inferences & \path{manuscript/tables/table1_validation_evidence_interpretation.csv} & archived manifest \\
Supplementary Table 33 source data & Validation controls and interpretation status & \path{manuscript/tables/supplementary_table_33_validation_controls.csv} & 7a88ca507fb7 \\
Causal-state sufficiency summaries & Two-model minimal sufficiency and wrapper result tables & \path{manuscript/tables/table8_causal_state_sufficiency_condition_summary.csv}; \path{manuscript/tables/table8_control_binding_wrapper_scrambled_summary.csv} & archived manifest \\
Causal-state sufficiency leakage audit & Prompt/scorer leakage and scope checks & \path{manuscript/tables/table8_causal_state_sufficiency_leakage_audit.csv} & e1fdfc5005ce \\
Local strict matching summaries & Local Qwen2.5-1.5B forced-budget entropy/token audit & \path{manuscript/tables/table16_local_strict_matching_condition_summary.csv}; \path{manuscript/tables/table16_local_strict_matching_gate_summary.csv} & archived manifest \\
SWE-bench Lite issue-to-file summaries & Complete 300-record issue-localization summaries and execution-slice preparation status & \path{results/paper1_revision/real_task_predictive_validity/swebench_v2_full300_api_20260529_merged/gate_status.json}; \path{results/paper1_revision/real_task_predictive_validity/swebench_v2_full300_api_20260529_merged/official_execution_slice_environment_status.json}; \path{results/paper1_revision/real_task_predictive_validity/swebench_v2_full300_api_20260529_merged/official_execution_slice_analysis_v3_apply_repair/official_execution_slice_gate_status.json} & archived manifest \\
Open-weight matrix config path & Model, scaffold, task-family and variant manifest & \path{results/paper1_revision/remote_5090_run3_20260508/nmi_formal_execution_manifest/execution_manifest.json} & 11e3390109fd \\
Formal API validation config path & API validation run manifest & \path{results/paper1_revision/remote_5090_run3_20260508/formal_api_validation/formal_api_full_20260510_0030/run_manifest.json} & 8712dbea8b1b \\
Distribution-matched control construction & Offline/no-context action-prior policy and prompt construction & \path{src/freewill/structured_control_validation.py} & eba9d1039034 \\
Strict-lesion construction & Removal of decisive intervention-carrying field with non-decisive context retained & \path{src/freewill/structured_control_validation.py} & eba9d1039034 \\
Counterfactual transform QC and consistency-passing transform & Transform definitions and formal holdout launcher & \path{src/freewill/structured_control_validation.py}; \path{scripts/run/run_formal_open_weight_validation.py} & eba9d1039034 / 0f9cfd41e1af \\
Token/compute criterion script & Prompt-token, total-token and latency-proxy criterion analysis & \path{scripts/run/analyze_formal_validation_results.py}; \path{scripts/run/audit_open_weight_token_entropy.py} & 404f5edb4132 / f037f6424513 \\
Entropy-prior behavioral-control script & Calibration-split action-prior selection and entropy-prior no-field behavioral control & \path{scripts/run/select_formal_entropy_settings.py}; \path{src/freewill/structured_control_validation.py} & f29b12ebe627 / eba9d1039034 \\
Local strict matching script & Local tokenizer padding, fixed-generation and forced-budget audit implementation & \path{scripts/run/run_strict_matching_local.py}; \path{src/freewill/strict_matching_local.py} & archived manifest \\
SWE-bench Lite localization scripts & Dataset fetch, oracle-free issue-to-file prompts, shard merge, readiness audit, execution-slice patch generation and v3 apply-check/repair filtering & \path{scripts/run/run_swebench_file_localization_v2_api.py}; \path{scripts/run/merge_swebench_fileloc_v2_shards.py}; \path{scripts/run/run_swebench_execution_slice_patch_generation_api.py}; \path{scripts/run/run_swebench_patch_generation_v3_apply_repair.py}; \path{src/freewill/real_task_predictive_validity.py} & archived manifest \\
\end{longtable}
\normalsize
\end{landscape}

\begin{landscape}
\tiny
\setlength{\tabcolsep}{2pt}
\renewcommand{\arraystretch}{1.12}
\begin{longtable}{@{}L{0.150\linewidth}L{0.065\linewidth}L{0.275\linewidth}L{0.165\linewidth}L{0.190\linewidth}L{0.090\linewidth}@{}}
\caption{Claim-to-evidence reproducibility map.}\label{tab:s7_claim_artifact_map}\\
\toprule
Claim & Main table & Result export & Criterion-summary alias & Analysis entry point & Checksum prefix \\
\midrule
\endfirsthead
\caption[]{Claim-to-evidence reproducibility map. (continued)}\\
\toprule
Claim & Main table & Result export & Criterion-summary alias & Analysis entry point & Checksum prefix \\
\midrule
\endhead
\midrule
\multicolumn{6}{r}{Continued on next page}\\
\endfoot
\bottomrule
\endlastfoot
baseline lesion matrix separates unpredictability and trace structure & Table 2 & \ttthree{table2\_baseline}{lesion\_matrix}{gate\_summary.csv} & \ttbreak{gate/baseline\_lesion\_matrix}{gate\_status.json} & \ttbreak{paper1-run-baseline-lesion-matrix}{configs/ai\_matrix.yaml} & 5f2fae5d2ab5 \\
matched-interface AFCI rules out field availability alone & Table 3 & \ttthree{table3\_schema\_gate\_summary.csv}{table3\_schema\_afci\_by\_variant.csv}{table3\_schema\_comparator\_gate\_status.csv} & \ttbreak{gate/matched\_interface\_controls}{gate\_status.json} & \ttbreak{paper1-run-matched-interface-controls}{configs/schema\_equalized\_controls.yaml} & 2fdfaffdbfdb \\
finite-action-code behavior supports action-level component coupling & Table 4 & \ttbreak{table4\_behavior\_gate\_summary.csv}{table4\_behavior\_dataset\_gate\_contrasts.csv} & \ttbreak{gate/finite\_action\_behavior}{gate\_status.json} & \ttbreak{paper1-run-finite-action-behavior}{configs/behavioral\_counterfactuals.yaml} & 55820262e805 \\
strong stochastic controls do not recover structured-control profile & Table 9 & \ttbreak{table7\_robustness}{gate\_summary.csv} & \ttbreak{gate/strong\_stochastic\_controls}{gate\_status.json} & \ttbreak{paper1-run-strong-stochastic-controls}{configs/ai\_matrix.yaml} & 1f1de7d4d18d \\
predefined entropy closeness was not satisfied & Table 6 / Table 12 & \ttthree{table6\_strict\_a5\_em\_calibration\_gate\_summary.csv}{table6\_a5\_entropy\_matching\_calibration\_gate\_summary.csv}{table12\_entropy\_calibration\_threshold\_summary.csv} & \ttbreak{gate/entropy\_calibration\_diagnostic}{gate\_status.json} & \ttbreak{paper1-run-entropy-diagnostics}{paper1-run-a5-entropy-calibration} & b6150257c3ae \\
cross-model tested subset preserves structured-control greater-than high-stochasticity direction & Table 9 & \ttbreak{table7\_cross\_model}{gate\_summary.csv} & \ttbreak{gate/cross\_model\_subset}{gate\_status.json} & \ttbreak{paper1-run-cross-model-subset}{configs/ai\_matrix.yaml} & 2f8b5ff04901 \\
open-weight reference model preserves structured-control greater-than high-stochasticity AFCI direction & Table 9 / Table 10 & \ttbreak{table7\_open\_weight\_afci\_by\_variant.csv}{table7\_open\_weight\_afci\_by\_dataset\_variant\_seed.csv} & \ttbreak{gate/open\_weight\_anchor}{gate\_status.json} & \ttbreak{paper1-run-open-weight-anchor}{configs/open\_weight\_anchor.yaml} & d76affe4ffe8 \\
minimal decisive field recovers the finite-action binding readout & Table 8 & \ttbreak{table8\_causal\_state\_sufficiency}{condition\_summary.csv} & \ttbreak{causal\_state\_sufficiency}{gate\_status.json} & \ttbreak{scripts/run}{run\_causal\_state\_sufficiency\_api.py} & archived manifest \\
local wrapper reduces scrambled-field binding & Table 8 & \ttbreak{table8\_control\_binding\_wrapper}{scrambled\_summary.csv} & \ttbreak{causal\_state\_sufficiency}{gate\_status.json} & \ttbreak{scripts/run}{run\_causal\_state\_sufficiency\_api.py} & archived manifest \\
bounded local forced-budget entropy/token audit preserves structured-state advantage & Table 16 & \ttbreak{table16\_local\_strict\_matching}{condition\_summary.csv} & \ttbreak{local\_strict\_matching}{gate\_status.json} & \ttbreak{scripts/run}{run\_strict\_matching\_local.py} & 0206bfa8d6db \\
SWE-bench Lite issue-to-file validation passes oracle-free predictor gate & Table 17 & \ttbreak{swebench\_v2\_full300}{api\_20260529\_merged/gate\_status.json} & \ttbreak{real\_task\_predictive\_validity}{swebench\_v2\_readiness\_audit.json} & \ttbreak{scripts/run}{run\_swebench\_file\_localization\_v2\_api.py} & archived manifest \\
blinded audit supports scorer interpretation & Table 9 & \ttbreak{table7\_blind\_audit\_reliability}{by\_dimension.csv} & \ttbreak{gate/strict\_blind\_audit}{gate\_status.json} & \ttbreak{paper1-run-blind-audit-export}{configs/ai\_matrix.yaml} & bffe01d6d1fd \\
\end{longtable}
\normalsize
\end{landscape}

Interface-diagnostic rows are summarized separately from effect estimates. The taxonomy distinguishes final-criterion-compatible interface diagnostics, non-final provider-interface attempts and entropy-calibration diagnostics. The export-coverage table lists the table exports underlying the manuscript and this supplement.

\begin{landscape}
\scriptsize
\setlength{\tabcolsep}{3pt}
\renewcommand{\arraystretch}{1.12}
\begin{longtable}{@{}L{0.160\linewidth}L{0.160\linewidth}L{0.160\linewidth}L{0.160\linewidth}L{0.160\linewidth}L{0.120\linewidth}@{}}
\caption{Interface-diagnostic taxonomy and interpretation.}\label{tab:s7_interface_diagnostics}\\
\toprule
Source & Experiment & Model & Diagnostic type & Rows & Interpretation \\
\midrule
\endfirsthead
\caption[]{Interface-diagnostic taxonomy and interpretation. (continued)}\\
\toprule
Source & Experiment & Model & Diagnostic type & Rows & Interpretation \\
\midrule
\endhead
\midrule
\multicolumn{6}{r}{Continued on next page}\\
\endfoot
\bottomrule
\endlastfoot
Cross-model robustness & Gemini full run & gemini-2.5-flash & payload not object & 2 & Final criterion met; retained as interface diagnostic. \\
Provider-interface compatibility check & Qwen direct-provider check & qwen3-30b-a3b & HTTP 400 & 100 & Compatibility rows outside the final-evidence criteria. \\
Entropy-calibration diagnostic & entropy-attempt calibration & qwen3-30b-a3b fallback & HTTP 500 / fallback metadata & 594 & Diagnostic calibration rows; predefined closeness was not satisfied. \\
Baseline lesion matrix & aggregate baseline accounting & gpt-5.4-mini & aggregate only & 16 & Aggregate unrecovered-output count; no trace-level taxonomy. \\
\end{longtable}
\normalsize
\end{landscape}

\begin{landscape}
\tiny
\setlength{\tabcolsep}{3pt}
\renewcommand{\arraystretch}{1.12}
\begin{longtable}{@{}L{0.200\linewidth}L{0.250\linewidth}L{0.250\linewidth}L{0.220\linewidth}@{}}
\caption{Table-export coverage. Full-precision values are retained in the CSV exports. Data rows exclude the header row and any empty trailing lines.}\label{tab:s7_exports}\\
\toprule
Export label & Data rows & Columns & Leading columns \\
\midrule
\endfirsthead
\caption[]{Table-export coverage. Full-precision values are retained in the CSV exports. Data rows exclude the header row and any empty trailing lines. (continued)}\\
\toprule
Export label & Data rows & Columns & Leading columns \\
\midrule
\endhead
\midrule
\multicolumn{4}{r}{Continued on next page}\\
\endfoot
\bottomrule
\endlastfoot
Variant name registry & 17 & 5 & canonical variant, used in, definition, aliases, role \\
Mechanistic claim map & 3 & 5 & claim component, operational test, contrary result, interpretive boundary \\
Baseline lesion-matrix criterion summary & 12 & 3 & criterion, field, value \\
Baseline trace metrics by dataset, variant and replicate index & 168 & 19 & variant, VBI, SCI, VEI, RSI, ACI, CSVI, unpredictability ... \\
Baseline trace metrics by variant & 8 & 18 & variant, VBI mean, SCI mean, VEI mean, RSI mean, ACI mean, CSVI mean, unpredictability mean ... \\
Matched-interface AFCI by dataset, variant and replicate index & 189 & 10 & dataset namespace, variant, replicate index, reason alignment, memory alignment, veto alignment ... \\
Matched-interface AFCI by variant & 9 & 16 & variant, alignment means, AFCI mean, AFCI with suppression mean, SD fields ... \\
Matched-interface dataset contrasts & 56 & 5 & dataset namespace, variant, AFCI, structured full AFCI, paired contrast \\
Matched-interface comparator criterion status & 7 & 4 & control variant, dataset count, structured greater-than-control dataset count, recovery count \\
Matched-interface criterion summary & 11 & 3 & criterion, field, value \\
Finite-action condition rates & 504 & 7 & dataset namespace, variant, replicate index, condition, correct rate, unmapped rate, row count \\
Finite-action dataset criterion contrasts & 7 & 10 & dataset namespace, structured variant, ablation comparators, high-stochasticity comparator, component contrasts ... \\
Finite-action criterion summary & 16 & 3 & criterion, field, value \\
Finite-action metrics by dataset, variant and replicate index & 84 & 16 & dataset namespace, variant, replicate index, condition rates and component scores ... \\
Finite-action summary by dataset and variant & 28 & 9 & dataset namespace, variant, B scores, cue specificity, behavioral structured action control ... \\
Finite-action summary by variant & 4 & 16 & variant, component-score means, behavioral composite mean, false-positive and row-count summaries ... \\
Entropy diagnostics by source and variant & 20 & 11 & source, variant, raw, canonical, semantic and action-family entropy means and SDs ... \\
Full entropy diagnostic units & 594 & 14 & source, output namespace, variant, replicate index, row count, unique action counts ... \\
Strict entropy-attempt calibration grid & 28 & 12 & variant, temperature, top-p, paired counts, absolute entropy gaps ... \\
Strict entropy-attempt criterion summary & 13 & 3 & criterion, field, value \\
Finite-action entropy-calibration grid & 14 & 11 & variant, dataset count, row count, predefined entropy gap, unmapped action rate, entropy gaps ... \\
Finite-action entropy-calibration criterion summary & 30 & 3 & criterion, field, value \\
Finite-action entropy-calibration holdout grid & 2 & 11 & variant, dataset count, row count, predefined entropy gap, unmapped action rate, entropy gaps ... \\
Blinded-audit reliability by dimension & 7 & 6 & dimension, annotator labels, overlap items, agreement, Cohen's kappa \\
Cross-model criterion summary & 4 & 14 & model, status, experiment profile, calls, interface diagnostics, directional dataset counts ... \\
Open-weight AFCI by dataset, variant and replicate index & 105 & 10 & dataset namespace, variant, replicate index, alignment terms and AFCI \\
Open-weight AFCI by variant & 5 & 16 & variant, alignment means, AFCI means, SD fields and rows \\
Robustness criterion summary & 8 & 13 & experiment, status, calls, interface diagnostics, directional counts, parse and audit statistics \\
Causal-state sufficiency condition summary & 12 & 13 & model, condition, rows, accuracy, wrapped accuracy, wrapper change rate ... \\
Causal-state sufficiency gate summary & 2 & 10 & model, full-state accuracy, only-decisive accuracy, best control accuracy, recovery fraction ... \\
Causal-state sufficiency task cells & 240 & 4 & model, task family, full-state accuracy, only-decisive accuracy \\
Causal-state sufficiency leakage audit & 5 & 4 & audit item, evidence, status, interpretation \\
Control-binding wrapper scrambled summary & 2 & 7 & model, rows, raw scrambled following, wrapped scrambled following, raw expected accuracy ... \\
Local strict matching condition summary & 2 & 10 & condition, rows, accuracy, invalid rate, canonical-action entropy, token and latency means \\
Local strict matching gate summary & 1 & 12 & status, prompt match rate, completion match rate, total-token ratio, latency ratio ... \\
SWE-bench file-localization predictor summary & 11 & 5 & predictor, hit@3 AUC, hit@3 Pearson, no-violation AUC, sign convention \\
SWE-bench file-localization model summary & 6 & 13 & model, model-instance rows, localization hit@3, violation rates, binding score and execution-slice boundary columns \\
SWE-bench file-localization gate summary & 1 & 21 & status, rows, model-instance rows, hit@3 counts, AUC summaries, wrapper status and execution-slice boundary columns \\
Derived effect-size summary & 6 & 3 & evidence block, direction, stronger summary \\
Proxy token/compute audit summary & 9 & 5 & audit item, measurement or proxy, availability, evidence source, interpretation \\
Entropy calibration threshold summary & 44 & 12 & source, variant, entropy gaps, strict threshold, pass flag \\
Interface-diagnostic taxonomy & 4 & 6 & source, experiment, model, diagnostic type, rows, interpretation \\
Claim-to-evidence reproducibility map & 12 & 7 & claim, main table, result export, criterion summary, analysis entry point, SHA-256 \\
Statistical sensitivity summary & 8 & 10 & contrast, independent unit, direction count, minimum contrast, sign test \\
\end{longtable}
\normalsize
\end{landscape}

\clearpage
\section*{S8. Statistical and Scoring Definitions}

\begin{landscape}
\scriptsize
\setlength{\tabcolsep}{1.5pt}
\renewcommand{\arraystretch}{1.12}
\begin{longtable}{@{}L{0.245\linewidth}L{0.090\linewidth}L{0.055\linewidth}L{0.060\linewidth}L{0.080\linewidth}L{0.080\linewidth}L{0.095\linewidth}L{0.085\linewidth}L{0.060\linewidth}@{}}
\caption{Dataset-level statistical sensitivity summary for primary contrasts. Exact sign tests are one-sided under a 0.5 null probability for a positive dataset direction; two-sided values are doubled and capped at one.}\label{tab:s8_stat_sensitivity}\\
\toprule
Contrast & Unit & Units & \plainbreak{Positive}{units} & \plainbreak{Minimum}{contrast} & \plainbreak{Mean}{contrast} & \plainbreak{LODO}{minimum mean} & \plainbreak{Top-effect}{removed mean} & \plainbreak{One-sided}{sign \(p\)} \\
\midrule
\endfirsthead
\caption[]{Dataset-level statistical sensitivity summary for primary contrasts. (continued)}\\
\toprule
Contrast & Unit & Units & \plainbreak{Positive}{units} & \plainbreak{Minimum}{contrast} & \plainbreak{Mean}{contrast} & \plainbreak{LODO}{minimum mean} & \plainbreak{Top-effect}{removed mean} & \plainbreak{One-sided}{sign \(p\)} \\
\midrule
\endhead
\midrule
\multicolumn{9}{r}{Continued on next page}\\
\endfoot
\bottomrule
\endlastfoot
finite behavior: structured-control--high-stochasticity composite & dataset namespace & 7 & 7 & 0.832 & 0.886 & 0.875 & 0.875 & 0.0078125 \\
reason ablation: structured-control--reason lesion B\_RSI & dataset namespace & 7 & 7 & 0.964 & 1.017 & 0.982 & 0.982 & 0.0078125 \\
veto ablation: structured-control--veto lesion B\_VEI & dataset namespace & 7 & 7 & 0.992 & 0.997 & 0.996 & 0.996 & 0.0078125 \\
memory behavior: structured-control--high-stochasticity B\_MCI & dataset namespace & 7 & 7 & 0.556 & 0.774 & 0.751 & 0.751 & 0.0078125 \\
self-continuity behavior: structured-control--high-stochasticity B\_SCI & dataset namespace & 7 & 7 & 0.608 & 0.811 & 0.795 & 0.795 & 0.0078125 \\
schema AFCI: structured full--stochastic plain & dataset namespace & 7 & 7 & 0.262 & 0.302 & 0.297 & 0.297 & 0.0078125 \\
schema AFCI: structured full--weakest control comparator & dataset namespace & 7 & 7 & 0.492 & 0.500 & 0.499 & 0.499 & 0.0078125 \\
open-weight AFCI: structured-control--high-stochasticity & dataset namespace & 7 & 7 & 0.467 & 0.478 & 0.476 & 0.476 & 0.0078125 \\
\end{longtable}
\normalsize
\end{landscape}

Let \(\mathcal{D}\) denote the seven dataset namespaces and \(\mathcal{I}_{d,v}\) the retained generations for dataset \(d\) and variant \(v\). Raw final-action unpredictability is reported as Shannon entropy \citep{shannon1948mathematical},
\begin{equation}
H_{d,v}=-\sum_{a\in\mathcal{A}_{d,v}}\hat{p}_{d,v}(a)\log_2\hat{p}_{d,v}(a),
\end{equation}
where \(\mathcal{A}_{d,v}\) is the observed final-action support and \(\hat{p}_{d,v}\) is the empirical action frequency. The baseline structured composite is
\begin{equation}
C_{d,v}=\frac{1}{4}\left(\mathrm{SCI}_{d,v}+\mathrm{RSI}_{d,v}+\mathrm{VEI}_{d,v}+\mathrm{ACI}_{d,v}\right).
\end{equation}
For matched-interface controls, each scored generation has
\begin{equation}
\mathrm{AFCI}_{i,v}=\frac{1}{4}\left(R_{i,v}+M_{i,v}+Q_{i,v}+S_{i,v}\right),
\end{equation}
where \(R\), \(M\), \(Q\) and \(S\) denote reason-action, memory-action, veto-action and self-action alignment terms. The implementation uses emitted trace fields, first impulse, final action and predefined field-source metadata to evaluate these terms. Thus AFCI is an interface-level action-field coupling analysis.

For the finite-action-code benchmark, the canonicalizer maps the raw final action \(o_i\) to \(\hat{y}_i=g(o_i)\in\mathcal{Y}\), where \(\mathcal{Y}=\{\texttt{ACTION\_A},\texttt{ACTION\_B},\texttt{VETO},\texttt{DEFER},\texttt{RECALL\_PRIOR},\texttt{INVALID\_OR\_UNMAPPED}\}\). If \(y_i^-\) is the scorer-side expected action before the manipulation and \(y_i^+\) is the scorer-side expected action after the manipulation, then
\begin{align}
c_i &= \mathbf{1}\{\hat{y}_i=y_i^+\},\\
u_i &= \mathbf{1}\{z_i=\mathrm{irrelevant}\}\mathbf{1}\{\hat{y}_i\neq y_i^-\}.
\end{align}
The component score for component \(k\) is the target-minus-irrelevant contrast
\begin{equation}
B_k(v)=\widehat{\Pr}(c_i=1\mid z_i\in\mathcal{T}_k,v)-\widehat{\Pr}(u_i=1\mid z_i\in\mathcal{N}_k,v),
\end{equation}
where \(\mathcal{T}_k\) is the target-probe set and \(\mathcal{N}_k\) is the irrelevant or placebo-probe set.

Dataset-cluster bootstrap intervals resample dataset identifiers with replacement, following the non-parametric bootstrap principle of estimating uncertainty from resampled empirical units \citep{efron1993bootstrap}. For a metric \(m\), comparator \(q\), and dataset \(d\), the paired contrast is
\begin{equation}
\delta_d(q;m)=\bar{m}_{d,\mathrm{A4}}-\bar{m}_{d,q},
\end{equation}
with \(\mathrm{A4}\) replaced by \(\mathrm{A4\_full}\) for matched-interface AFCI. The reported lower 95\% bound is the empirical 2.5th percentile of the bootstrap distribution over \(|\mathcal{D}|\)-dataset resamples.

For entropy calibration, layer-specific entropy \(H^{(\ell)}_{d,v}\) is computed for raw, rule-canonical, semantic-cluster and action-family representations. If \(\theta=(T,p)\) denotes a candidate high-stochasticity temperature/top-p setting, the predefined calibration gap is
\begin{equation}
\Gamma(\theta)=\max_{\ell}\left|\frac{1}{|\mathcal{D}|}\sum_{d\in\mathcal{D}}H^{(\ell)}_{d,\mathrm{A5}(\theta)}-\frac{1}{|\mathcal{D}|}\sum_{d\in\mathcal{D}}H^{(\ell)}_{d,\mathrm{A4}}\right|.
\end{equation}
The predefined criterion requires \(\Gamma(\theta)\leq 0.15\). The retained calibration attempts did not satisfy this criterion; the entropy results are reported as calibration diagnostics.

\clearpage
\section*{S10. Open-Weight and Formal Validation Controls}

This section records the validation experiments that support the main-text evidence summary. The baseline, matched-interface, finite-action and entropy-diagnostic sections above provide the primary mechanistic evidence and calibration checks. The rows below summarize the open-weight finite-action matrix, strict-lesion slice, minimal causal-state sufficiency experiment, local control-binding wrapper, formal control-audit criteria, local forced-budget entropy/token matching, entropy-prior behavioral control, the SWE-bench Lite issue-to-file localization validation, counterfactual-transform audit, API validation and expanded blinded annotation.

\scriptsize
\setlength{\LTleft}{0pt}
\setlength{\LTright}{0pt}
\begin{longtable}{p{0.20\textwidth}p{0.27\textwidth}p{0.31\textwidth}p{0.16\textwidth}}
\caption{Validation controls and interpretation status.}
\label{tab:s33_validation}\\
\toprule
Evidence block & Scope & Main result & Interpretation status \\
\midrule
\endfirsthead
\toprule
Evidence block & Scope & Main result & Interpretation status \\
\midrule
\endhead
Open-weight finite-action matrix & Qwen2.5-7B-Instruct, Qwen2.5-14B-Instruct, Qwen2.5-32B-Instruct and Mistral-7B-Instruct-v0.3; 126,000 scored generations per model; three agent scaffolds; 20 task families; seven variants; three seeds & Structured accuracy exceeded distribution-matched offline/no-context controls in 236/240 model--scaffold--task cells, scrambled-context controls in 240/240 cells and no-fields controls in 238/240 cells; mean deltas were 0.593, 0.895 and 0.550, respectively & Supports open-weight size and family validation of the no-context control result \\
Independent Mistral family & Mistral-7B-Instruct-v0.3 full matrix & Core no-context controls satisfied the predefined criteria in 60/60 cells for no-fields, scrambled-context and distribution-matched controls & Supports non-Qwen model-family generalization \\
Strict Qwen32B target lesion & Qwen2.5-32B strict lesion slice; 28,800 scored generations & Structured accuracy exceeded strict target lesion in 60/60 cells; structured mean 0.998, strict-lesion mean 0.037, mean delta 0.961 & Supports the narrower strict-lesion claim; ordinary target lesions remain a boundary condition \\
Minimal causal-state sufficiency & \texttt{gpt-5.4-mini} and \texttt{qwen3-max}; 20 task families; six events per family; 1,440 fallback-free API generations & Full-state and only-decisive-field accuracy were both 1.000 in both models; the best non-decisive control was 0.208 and sufficiency recovery was 1.000 & Supports a diagnostic decisive-field readout within the finite-action protocol \\
Local control-binding wrapper & Same 1,440-row sufficiency experiment; no second model call & Irrelevant-cue accuracy remained 1.000; raw scrambled-field following was 0.958 for \texttt{gpt-5.4-mini} and 0.533 for \texttt{qwen3-max}, reduced to 0.033 in both models after wrapping & Finite-action guardrail evidence; not real-environment localization or task-execution reliability \\
Formal Qwen control-audit slice & 63,360 planned and observed rows; no duplicate analysis keys; parse/error rate 0.140\% & Structured accuracy exceeded no-fields, scrambled-context, distribution-matched, strict-lesion and formal control variants in the relevant cells & Supports formal-control behavior in the Qwen slice \\
Token-prompt audit slice & 17,280 planned and observed rows; no duplicate analysis keys; parse/error rate 0.069\% & Prompt-token counts were aligned for the entropy-token variant, but mean total-token ratios and latency proxies remained outside the strict token/compute criterion & Report as token-prompt controls with compute audit, not strict token/compute matching \\
Strict entropy criterion & Formal entropy calibration and holdout audit & Registered entropy gap criterion was not met across model--scaffold cells & Strict entropy matching is not established \\
Local forced-budget entropy/token audit & Qwen2.5-1.5B-Instruct CPU local inference; 20 task families; two events per family; 80 generations & Prompt and completion tokens matched exactly by event; total-token ratio 1.000; latency ratio 0.997; canonical-action entropy gap 0.020 bits; structured accuracy was 0.900 versus 0.325 for the action-prior no-state control & Bounded local strict-matching pass; original formal strict entropy/token gates remain boundary conditions \\
Entropy-prior no-field behavioral control & Qwen2.5-14B and Qwen2.5-32B calibration and holdout rows; 24,480 planned and observed rows; no duplicate analysis keys; parse/error rate 0.102\% & The entropy-prior no-field behavioral control remained below the structured-control variant in 120/120 cells, with mean delta 0.660; the strict entropy-gap criterion was satisfied in only 1/6 model--scaffold cells & Report as a final-action entropy-prior behavioral control, not strict entropy matching \\
SWE-bench Lite issue-to-file validation & \texttt{princeton-nlp/SWE-bench\_Lite} test split; 300 issue records; six API models; 18,000 condition/repeat rows aggregated to 1,800 model-issue records for the primary AUC analysis & Adding the oracle-free causal state-binding composite to the full non-CSB baseline increased constraint-clean hit@3 AUC from 0.873 to 0.935, with \(\Delta\)AUC 0.062 and issue-cluster bootstrap lower 95\% bound 0.025; leave-one-repository and leave-one-model sensitivities were positive in 12/12 and 6/6 analyses & Real issue-record localization predictor evidence; patch/test execution and issue resolution are outside this endpoint \\
SWE-bench Lite reliability calibration & Same 1,800 model--issue records; issue-grouped five-fold logistic probabilities for the full non-CSB baseline and baseline plus CSB & Reliability curves and threshold analyses stratify predicted hit@3 probability, observed constraint-clean hit@3 and hard-constraint violations across acceptance thresholds & Secondary issue-to-file reliability analysis of the same endpoint \\
Counterfactual transform QC & Deterministic counterfactual-flip transform-quality screen & Text--label consistency screening retained 57/240 candidate counterfactual rows as transform-quality boundary rows & Interpret only consistency-passing transforms as robustness evidence \\
Consistency-passing counterfactual transform & Frozen transform on Qwen2.5-14B and Qwen2.5-32B with jointly regenerated decisive context and expected action & The consistency-passing transform satisfied the predefined criterion in 120/120 scaffold--family cells; mean structured accuracy was 1.000, compared with 0.992 in the original structured rows & Use as deterministic-transform robustness evidence \\
Expanded blinded annotation & 1,200 full-overlap items, three annotators & Core dimensions had 100\% valid scoring, no unscorable items and no violations of the blinding protocol; mean pairwise Cohen's kappa was 0.961 and the lowest non-constant dimension kappa was 0.860 & Supports annotation reliability \\
Formal API external validation & Provider-reported run-time identifiers \texttt{gpt-5.4-mini}, \texttt{gemini-3.1-flash-lite-preview}, \texttt{deepseek-v3.2} and \texttt{qwen3-max}; 14,400 planned rows; 11,520 fallback-free remote API calls & All rows were present, duplicate analysis keys were zero, fallback calls were zero, unrecovered generations were zero and one parse/action error record remained. Structured accuracy exceeded no-fields, scrambled-context, distribution-matched and strict-lesion controls in 238/240, 240/240, 240/240 and 240/240 model--scaffold--task-family cells; all model-, scaffold- and task-family-cluster bootstrap lower 95\% bounds were positive & Supports formal external validation of the core no-context and strict-lesion controls \\
\bottomrule
\end{longtable}
\normalsize

The sufficiency experiment is a controlled finite-action test. The only-decisive condition exposes the current event's minimal action-relevant field and asks whether the model follows that field rather than surface context, a calibrated action prior or a decisive field from another event. It is not interpreted as unconstrained task solving. The examples below show the prompt-level semantics used for the minimal field; the expected action column remains scorer-side metadata used for evaluation, whereas the visible field is the intervention-carrying state supplied to the action query.

\begin{center}
\scriptsize
\begin{tabular}{p{0.18\textwidth}p{0.54\textwidth}p{0.18\textwidth}}
\toprule
Probe component & Example visible only-decisive field & Scorer-side target role \\
\midrule
Reason & Decisive reason field for the event: calibrated task evidence selects \texttt{ACT\_B}. & Target checks whether final action follows the current reason field. \\
Memory & Decisive memory field for the event: the corrected prior commitment is \texttt{ACT\_A}; stale or unrelated memory is non-decisive. & Target checks whether final action follows corrected memory rather than injected memory. \\
Veto/constraint & Decisive constraint/veto field for the event: a hard constraint invalidates ordinary actions and requires \texttt{ACT\_VETO}. & Target checks whether final action respects the current constraint field. \\
\bottomrule
\end{tabular}
\end{center}

The sufficiency leakage and scope audit separates hidden scorer metadata from intentionally visible decisive-field content. The only-decisive condition is not an answer-token-free hidden-reasoning task: it deliberately exposes the current event's minimal action-relevant field. The appropriate leakage question is therefore whether the scorer target variable or target key was serialized as metadata, and whether non-decisive controls withheld or broke the current decisive field.

\begin{center}
\scriptsize
\begin{tabular}{p{0.21\textwidth}p{0.43\textwidth}p{0.12\textwidth}p{0.18\textwidth}}
\toprule
Audit item & Evidence & Status & Interpretation \\
\midrule
\texttt{expected\_action} column not a prompt key & Prompt construction serializes condition, event id, task family, candidate actions and context; \texttt{expected\_action} is retained in traces for scorer-side evaluation after generation. & pass & The scorer target variable is not exposed as a prompt field or JSON key. \\
Scorer-side target used post-generation & Across 1,440 traces, \texttt{expected\_action\_match} is computed after generation from canonical \texttt{final\_action} versus \texttt{expected\_action}. & pass & The target column is a scorer-side annotation rather than a model-output field. \\
Non-decisive controls do not expose current decisive state & \texttt{surface\_context\_only} and \texttt{action\_prior\_only} had \texttt{field\_event\_id=none} in 240/240 rows each; \texttt{scrambled\_decisive\_field} used a non-current field id in 240/240 rows. & pass & Surface, prior and scrambled controls separate action priors or wrong-state fields from the current decisive field. \\
Only-decisive action-token status & \texttt{only\_decisive\_causal\_field} exposed a visible decisive action matching the scorer target in 240/240 rows; full-state and irrelevant-cue rows did the same by design. & scope & The sufficiency claim tests following the current minimal decisive field; it is not an answer-token-free hidden-reasoning claim. \\
Label wording scope & The prompt contains a generic instruction not to infer hidden expected-action labels, but does not serialize the \texttt{expected\_action} column or scorer target key. & pass & The generic instruction names the evaluation concept; it does not provide the hidden scorer variable. \\
\bottomrule
\end{tabular}
\end{center}

\paragraph{Scope of validation and execution slice.}
The formal criteria are interpreted conservatively. No-context and formal-control contrasts support the bounded claim that measured finite-action performance depended on structured state information in the tested agent protocol. Strict entropy and strict token/compute criteria that were not satisfied in the original formal slices are reported as limitations rather than post hoc passes. The local Qwen2.5-1.5B forced-budget audit is a separate bounded pass under exact prompt padding and fixed generation length. The SWE-bench Lite result is a full 300-record issue-to-file localization validation; patch application and test-suite execution were evaluated only in an exploratory slice. That frozen 50-record slice included official harness outcomes for two selected models, with 1/100 selected-model rows resolved and an inconclusive association gate. A later v3 apply-check/repair filter retained 8/100 locally applicable selected-model patches, eliminated harness errors on that filtered subset, and resolved 0/8. These outcomes define the scope of the localization result. The counterfactual-transform quality-control screen separates transform-quality rows from the consistency-passing deterministic transform used as robustness evidence.

\clearpage
\bibliographystyle{unsrtnat}
\bibliography{references}

@article{miller2001integrative,
  author = {Miller, Earl K. and Cohen, Jonathan D.},
  title = {An Integrative Theory of Prefrontal Cortex Function},
  journal = {Annual Review of Neuroscience},
  volume = {24},
  pages = {167--202},
  year = {2001},
  doi = {10.1146/annurev.neuro.24.1.167}
}

@article{logan1984inhibit,
  author = {Logan, Gordon D. and Cowan, William B.},
  title = {On the Ability to Inhibit Thought and Action: A Theory of an Act of Control},
  journal = {Psychological Review},
  volume = {91},
  number = {3},
  pages = {295--327},
  year = {1984},
  doi = {10.1037/0033-295X.91.3.295}
}

@article{aron2014inhibition,
  author = {Aron, Adam R. and Robbins, Trevor W. and Poldrack, Russell A.},
  title = {Inhibition and the Right Inferior Frontal Cortex: One Decade On},
  journal = {Trends in Cognitive Sciences},
  volume = {18},
  number = {4},
  pages = {177--185},
  year = {2014},
  doi = {10.1016/j.tics.2013.12.003}
}

@article{shenhav2013expected,
  author = {Shenhav, Amitai and Botvinick, Matthew M. and Cohen, Jonathan D.},
  title = {The Expected Value of Control: An Integrative Theory of Anterior Cingulate Cortex Function},
  journal = {Neuron},
  volume = {79},
  number = {2},
  pages = {217--240},
  year = {2013},
  doi = {10.1016/j.neuron.2013.07.007}
}

@book{efron1993bootstrap,
  author = {Efron, Bradley and Tibshirani, Robert J.},
  title = {An Introduction to the Bootstrap},
  publisher = {Chapman and Hall/CRC},
  year = {1993}
}

@inproceedings{wei2022cot,
  author = {Wei, Jason and Wang, Xuezhi and Schuurmans, Dale and Bosma, Maarten and Ichter, Brian and Xia, Fei and Chi, Ed and Le, Quoc and Zhou, Denny},
  title = {Chain-of-Thought Prompting Elicits Reasoning in Large Language Models},
  booktitle = {Advances in Neural Information Processing Systems},
  volume = {35},
  pages = {24824--24837},
  year = {2022}
}

@inproceedings{yao2023react,
  author = {Yao, Shunyu and Zhao, Jeffrey and Yu, Dian and Du, Nan and Shafran, Izhak and Narasimhan, Karthik and Cao, Yuan},
  title = {{ReAct}: Synergizing Reasoning and Acting in Language Models},
  booktitle = {International Conference on Learning Representations},
  year = {2023},
  url = {https://openreview.net/forum?id=WE_vluYUL-X}
}

@inproceedings{yao2023tot,
  author = {Yao, Shunyu and Yu, Dian and Zhao, Jeffrey and Shafran, Izhak and Griffiths, Thomas L. and Cao, Yuan and Narasimhan, Karthik},
  title = {Tree of Thoughts: Deliberate Problem Solving with Large Language Models},
  booktitle = {Advances in Neural Information Processing Systems},
  volume = {36},
  year = {2023},
  url = {https://openreview.net/forum?id=5Xc1ecxO1h}
}

@inproceedings{park2023generative,
  author = {Park, Joon Sung and O'Brien, Joseph C. and Cai, Carrie J. and Morris, Meredith Ringel and Liang, Percy and Bernstein, Michael S.},
  title = {Generative Agents: Interactive Simulacra of Human Behavior},
  booktitle = {Proceedings of the 36th Annual ACM Symposium on User Interface Software and Technology},
  year = {2023},
  doi = {10.1145/3586183.3606763}
}

@inproceedings{holtzman2020degeneration,
  author = {Holtzman, Ari and Buys, Jan and Du, Li and Forbes, Maxwell and Choi, Yejin},
  title = {The Curious Case of Neural Text Degeneration},
  booktitle = {International Conference on Learning Representations},
  year = {2020},
  url = {https://openreview.net/forum?id=rygGQyrFvH}
}

@inproceedings{zheng2023judge,
  author = {Zheng, Lianmin and Chiang, Wei-Lin and Sheng, Ying and Zhuang, Siyuan and Wu, Zhanghao and Zhuang, Yonghao and Lin, Zi and Li, Zhuohan and Li, Dacheng and Xing, Eric P. and Zhang, Hao and Gonzalez, Joseph E. and Stoica, Ion},
  title = {Judging {LLM}-as-a-Judge with {MT}-Bench and Chatbot Arena},
  booktitle = {Advances in Neural Information Processing Systems},
  volume = {36},
  year = {2023},
  url = {https://proceedings.neurips.cc/paper_files/paper/2023/hash/91f18a1287b398d378ef22505bf41832-Abstract-Datasets_and_Benchmarks.html}
}

@inproceedings{shinn2023reflexion,
  author = {Shinn, Noah and Cassano, Federico and Berman, Edward and Gopinath, Ashwin and Narasimhan, Karthik and Yao, Shunyu},
  title = {Reflexion: Language Agents with Verbal Reinforcement Learning},
  booktitle = {Advances in Neural Information Processing Systems},
  volume = {36},
  year = {2023},
  url = {https://papers.nips.cc/paper_files/paper/2023/hash/1b44b878bb782e6954cd888628510e90-Abstract-Conference.html}
}

@article{wang2024voyager,
  author = {Wang, Guanzhi and Xie, Yuqi and Jiang, Yunfan and Mandlekar, Ajay and Xiao, Chaowei and Zhu, Yuke and Fan, Linxi and Anandkumar, Anima},
  title = {Voyager: An Open-Ended Embodied Agent with Large Language Models},
  journal = {Transactions on Machine Learning Research},
  year = {2024},
  url = {https://voyager.minedojo.org/}
}

@inproceedings{fan2018hierarchical,
  author = {Fan, Angela and Lewis, Mike and Dauphin, Yann},
  title = {Hierarchical Neural Story Generation},
  booktitle = {Proceedings of the 56th Annual Meeting of the Association for Computational Linguistics},
  pages = {889--898},
  year = {2018},
  doi = {10.18653/v1/P18-1082}
}

@inproceedings{kuhn2023semantic,
  author = {Kuhn, Lorenz and Gal, Yarin and Farquhar, Sebastian},
  title = {Semantic Uncertainty: Linguistic Invariances for Uncertainty Estimation in Natural Language Generation},
  booktitle = {International Conference on Learning Representations},
  year = {2023},
  url = {https://openreview.net/forum?id=VD-AYtP0dve}
}

@article{shannon1948mathematical,
  author = {Shannon, Claude E.},
  title = {A Mathematical Theory of Communication},
  journal = {Bell System Technical Journal},
  volume = {27},
  number = {3},
  pages = {379--423},
  year = {1948},
  doi = {10.1002/j.1538-7305.1948.tb01338.x}
}

@article{liang2023helm,
  author = {Liang, Percy and Bommasani, Rishi and Lee, Tony and Tsipras, Dimitris and Soylu, Dilara and Yasunaga, Michihiro and Zhang, Yian and Narayanan, Deepak and Wu, Yuhuai and Kumar, Ananya and Newman, Benjamin and Yuan, Binhang and Yan, Bobby and Zhang, Ce and Cosgrove, Christian and Manning, Christopher D. and R{\'e}, Christopher and others},
  title = {Holistic Evaluation of Language Models},
  journal = {Transactions on Machine Learning Research},
  year = {2023},
  url = {https://openreview.net/forum?id=iO4LZibEqW}
}

@inproceedings{mitchell2019modelcards,
  author = {Mitchell, Margaret and Wu, Simone and Zaldivar, Andrew and Barnes, Parker and Vasserman, Lucy and Hutchinson, Ben and Spitzer, Elena and Raji, Inioluwa Deborah and Gebru, Timnit},
  title = {Model Cards for Model Reporting},
  booktitle = {Proceedings of the Conference on Fairness, Accountability, and Transparency},
  pages = {220--229},
  year = {2019},
  doi = {10.1145/3287560.3287596}
}

@article{gebru2021datasheets,
  author = {Gebru, Timnit and Morgenstern, Jamie and Vecchione, Briana and Vaughan, Jennifer Wortman and Wallach, Hanna and Daume III, Hal and Crawford, Kate},
  title = {Datasheets for Datasets},
  journal = {Communications of the ACM},
  volume = {64},
  number = {12},
  pages = {86--92},
  year = {2021},
  doi = {10.1145/3458723}
}

@inproceedings{dror2018hitchhiker,
  author = {Dror, Rotem and Baumer, Gili and Shlomov, Segev and Reichart, Roi},
  title = {The Hitchhiker's Guide to Testing Statistical Significance in Natural Language Processing},
  booktitle = {Proceedings of the 56th Annual Meeting of the Association for Computational Linguistics},
  pages = {1383--1392},
  year = {2018},
  doi = {10.18653/v1/P18-1128}
}

@article{cohen1960coefficient,
  author = {Cohen, Jacob},
  title = {A Coefficient of Agreement for Nominal Scales},
  journal = {Educational and Psychological Measurement},
  volume = {20},
  number = {1},
  pages = {37--46},
  year = {1960},
  doi = {10.1177/001316446002000104}
}

@article{laird1982random,
  author = {Laird, Nan M. and Ware, James H.},
  title = {Random-Effects Models for Longitudinal Data},
  journal = {Biometrics},
  volume = {38},
  number = {4},
  pages = {963--974},
  year = {1982},
  doi = {10.2307/2529876}
}

\end{document}